\documentclass[preprint,12pt,authoryear]{elsarticle}
\journal{Remote Sensing of Environment}

\usepackage[utf8]{inputenc}
\usepackage[T1]{fontenc}
\usepackage{graphicx}
\usepackage{url}
\usepackage{setspace}
\usepackage{amsmath,amssymb,amsfonts}
\usepackage{algorithmic}
\usepackage{textcomp}
\usepackage{xcolor}
\usepackage{booktabs}
\usepackage[normalem]{ulem}
\useunder{ine}{}{}
\usepackage{subcaption}
\usepackage{enumitem}
\usepackage{rotating}
\usepackage{multirow}{\tiny}
\usepackage{scalerel}
\usepackage{tikz}
\usepackage{lineno}

\usetikzlibrary{svg.path}
\usepackage[left=1.00in, right=1.00in, top=1.00in, bottom=1.00in]{geometry}
\usepackage{hyperref}

\title{In-Season Crop Progress in Unsurveyed Regions using Networks Trained on Synthetic Data}
\author{George Worrall}

\author{Jasmeet Judge \\ \small{Center for Remote Sensing \\ Agricultural and Biological Engineering Department \\ University of Florida}}

\begin{document}
	
	\begin{highlights}
		\item Satellite data are used in neural networks (NN) for crop progress estimation (CPE)
		\item Current CPE methods require survey data and cannot be used in unsurveyed regions (UR)
		\item Linked weather-crop-optical radiative transfer models provide synthetic data for UR
		\item NNs can be trained on synthetic data and real CP data from surveyed regions (SR)
		\item NNs trained on simultaneously on CP from SR and synthetic data improves CPE in UR
	\end{highlights}
	
	\begin{abstract}
		Many commodity crops have growth stages during which they are particularly vulnerable to stress-induced yield loss. In-season crop progress information is useful for quantifying crop risk, and satellite remote sensing (RS) can be used to track progress at regional scales. At present, all existing RS-based crop progress estimation (CPE) methods which target crop-specific stages rely on ground truth data for training/calibration. Such data are collected via field trials or surveys. This reliance on ground survey data confines CPE methods to surveyed regions, limiting their utility. In this study, a new method is developed for conducting RS-based in-season CPE in unsurveyed regions by combining data from surveyed regions with synthetic crop progress data generated for an unsurveyed region of interest. Corn-growing zones in Argentina were used as surrogate `unsurveyed' regions. These zones have climates and dual planting systems which differ the single planting system in the US Midwest -- the surveyed region in this study. Existing weather generation, crop growth, and optical radiative transfer models were linked to produce synthetic weather, crop progress, and canopy reflectance data. These data mimic weather, cultivars, and cropping practices in the unsurveyed region. A neural network (NN) method based upon bi-directional Long Short-Term Memory was trained separately on surveyed data, synthetic data, and two different combinations of surveyed and synthetic data. In the absence of real validation data in unsurveyed regions, a stopping criterion was developed which uses the weighted divergence of surveyed and synthetic data validation loss. F$_1$ score was modified to measure CPE accuracy when trained the NN was trained on each data combination, with scores based on over- and under-estimates of crop progress throughout the season. Including synthetic data during training improved performance in 9 out of 11 corn-growing zones in Argentina. Net F$_1$ scores across all crop progress stages increased by 8.7\% when trained on a combination of surveyed region and synthetic data, and overall performance was only 21\% lower than the best case scenario, when the NN was trained on surveyed data and applied in the US Midwest. Performance gain from synthetic data was greatest in zones with dual planting windows, while the inclusion of surveyed region data from the US Midwest helped mitigate NN sensitivity to noise in NDVI data. Overall results suggest in-season CPE in other unsurveyed regions may be possible with increased quantity and variety of synthetic crop progress data.
	\end{abstract}


	\doublespacing
	\maketitle
	\section{Introduction}
	Many commodity crops have critical growth stages during which they are at an increased risk of stress-induced yield loss, such as silking in corn and pod-fill in soybean \citep{claassen_water_1970, doss_effect_1974}. Extreme weather events are expected to disrupt global food supply with increased frequency \citep{ray_climate_2015, mehrabi_synchronized_2019}, and knowledge of the timing of crop growth stages in major growing regions is vital to accurately anticipate and quantify potential yield loss. Tracking crop growth stage progression, hereafter crop progress, is commonly done using ground surveyors. This task is laborious and requires widespread reporting infrastructure to produce the accurate and timely data. For example, in the US, the USDA coordinates weekly Crop Progress and Condition (CPR) reporting from over 5000 surveyors during the growing season \citep{usda_national_agricultural_statistics_service_national_nodate}. However, many major exporting countries do not have publicly available crop progress data or surveys. Growers and grain market participants are then less informed about the present vulnerability of a region's crop to adverse weather, which can lead to significant revisions to production estimates by government institutions when effects become apparent late in a growing season, (e.g. \cite{colussi_safrinha_2021}). For many regions, their geographical size and lack of survey infrastructure prevent the collection of comprehensive crop progress data.
	
	In these unsurveyed regions, remote sensing (RS) observations may be used to estimate crop progress. The global coverage and frequent revisit times of modern earth-observing satellite platforms can provide cost-effective assessments of crop progress timing and may alleviate the need for ground survey infrastructure. Existing RS-based crop progress estimation (CPE) methods use observed crop canopy reflectances from a target growing region to derive relationships between vegetation indices (VI) such as normalized difference vegetation index (NDVI) and crop progress \citep{gao_mapping_2021}. Many studies relate full-season VI curves to crop progress timing, such as \cite{diao_remote_2020, diao_hybrid_2021, seo_improving_2019}. These methods are useful for post-season analysis, but timely and frequent in-season estimates of crop progress are needed to quantify crop risk. This is because many critical growth stages in commodity crops, such as silking in corn \citep{claassen_water_1970}, last under two weeks. To meet these requirements, RS-based in-season studies have sought to produce actionable CPE through trend-based approaches, which use momentum indicators and thresholding to produce estimates for events such as emergence \citep{gao_within-season_2020} and cover crop termination \citep{gao_detecting_2020}. These approaches, however, utilize region-specific, user-defined parameters and do not target all growth stages. In addition, relationships between trend-based reference points such as `green-up' and `maximum greenness' and crop-specific growth stages is known to vary by method used \citep{liu_real-time_2018, gao_mapping_2021}.
	
	Machine learning (ML) techniques, such as Hidden Markov Models (HMM) \citep{shen_hidden_2013} and Parametric Empirical Bayes (PEB) \citep{ghamghami_parametric_2020}, have been used to develop in-season CPE methods that can derive crop progress across all stages. These methods use historical crop progress as priors and estimate stage transition probabilities directly from RS and weather observations. However, both HMM and PEB rely on expectation maximization, which is sensitive to random weight initialization, and cannot derive higher order relationships between variables that are possible with contemporary ML techniques such as neural networks (NNs). Recently, \cite{worrall_domain-guided_2021} found that Long Short-Term Memory (LSTM), an NN architecture widely used in time series analysis \citep{zeyer_comparison_2019, cheng_machine_2019, song_time-series_2020}, could be used to combine RS-based Fraction of Photosynthetically Active Radiation (FPAR) data with soil and weather conditions for in-season CPE. \cite{worrall_domain-guided_2021} found that LSTM-based in-season CPE produced 43\% higher average Nash-Sutcliffe Efficiency across all stages than the HMM method from \cite{shen_hidden_2013}. NN-based in-season CPE has shown further improvement when guided by biophysical crop models. For example, \cite{worrall_-season_2022} found that growth stage and water stress-related information from biophysical crop growth model simulations increased overall in-season CPE accuracy, particularly during seasons with abnormal temperatures and higher rainfall. Despite these recent advances in RS-based CPE, all existing methods still require ground truth crop progress data for calibration and/or training, limiting their application to regions where such data are available \citep{seo_improving_2019, diao_hybrid_2021, worrall_-season_2022}.
	
	In unsurveyed regions, studies have mapped RS-derived land surface phenology, such as `start-of-season' and `end-of-season' \citep{vintrou_comparative_2014,duncan_potential_2015}. However, crop progress in unsurveyed regions cannot be estimated from these points because ground survey data is required to derive crop-specific stage estimates from land surface phenology \citep{seo_improving_2019, diao_remote_2020}, which are known to change based on the target crop, method used, and region of application \citep{liu_real-time_2018, gao_mapping_2021}. To date, there are no RS-based methods that produce in-season CPE without ground survey data.
	
	RS-based CPE in unsurveyed regions is difficult because there are no labeled data. In agricultural RS, crop mapping and yield estimation studies have addressed labeled data sparsity using transfer learning \citep{nowakowski_crop_2021, luo_developing_2022, khaki_simultaneous_2021}, which involves pre-training an NN on a large, often unrelated dataset before fine-tuning it on limited available labeled data. Given the success of transfer learning in these studies, it may be possible to leverage information from labeled data in surveyed regions, such as the US, to train NN-based methods and apply them to unsurveyed regions for in-season CPE. However, without any fine-tuning examples of cropping practices, cultivar choices, and climate in unsurveyed regions, their transferability may be limited. 
	
	Another approach to mitigating data sparsity is to use generative methods to provide additional, synthetic training examples. Generation of synthetic examples is common in RS-based scene classification, where manual partitioning and labeling of scenes is labor intensive \citep{cheng_remote_2020}. A common generative method used in scene classification is Generative Adversarial Networks (GANs), which produce additional, synthetic examples of class-specific patterns from a limited number of hand-labeled samples \citep{qian_generating_2020,han_sample_2020}. GANs are effective for image recognition tasks, but they require some labeled data to learn by example, which are not available for unsurveyed regions. Therefore, any generative method used to produce synthetic crop progress data must contain the underlying physical relationships among weather, crop growth and development, and observed reflectances that are present in real data without the need for labeled examples. 

	Physically-based models for weather generation, crop growth, and canopy reflectances are widely used in agriculture and RS studies. Studies in agronomy often utilize stochastic weather generators to assess the impacts of different climate scenarios on crop production \citep{semenov_use_1997, apipattanavis_linking_2010, supit_assessing_2012}. Some studies calibrate biophysical crop models, driven by soil and weather data, to produce in-season yield estimates  \citep{archontoulis_predicting_2020, akhavizadegan_time-dependent_2021}. In RS, optical radiative transfer models (RTMs), which simulate canopy reflectances at visible and infra-red wavelengths from plant structure and leaf pigment concentrations, are often inverted to retrieve estimates of Leaf Area Index (LAI) and chlorophyll content \citep{duan_inversion_2014, jay_retrieving_2017}. When these weather, crop, and RS models are linked, they can produce synthetic data that emulate real spatio-temporal patterns in agricultural production and their optical signatures, as observed from RS platforms. For example, \cite{lobell_scalable_2015} used the APSIM crop model and LAI-VI relationships to calibrate an RS-based regression model for post-season yield estimation. There is precedent in the literature for using historical weather, mechanistic crop models, and RTMs to generate data for synthetic studies, such as soil moisture downscaling using microwaves \citep{chakrabarti_downscaling_2015} and forest biomass estimation \citep{fassnacht_using_2018}. However, to our knowledge, synthetic data generated from linked weather-crop-RTM models have not been leveraged to train NNs for application in areas without ground truth data.
	
	The goal of this study is to explore how well synthetic data may mitigate the need for ground survey data to produce in-season crop progress estimates in unsurveyed regions. In this study, domain-specific models from meteorology, agronomy, and optical RS are linked to generate synthetic crop progress data. An NN-based RS CPE method is trained on both this synthetic data and observed data from surveyed regions in the US, then applied to CPE in unsurveyed regions. To assess the value of including synthetic crop progress data in unsurveyed region CPE, we investigated the performance of the NN-based method when it was trained solely on surveyed region data from the US. Specifically, the objectives of this study are to (1) implement a linked weather generator-crop model-RTM framework that can produce synthetic examples of crop progress data for an unsurveyed region, (2) investigate the impact of exposure to synthetic crop progress data during training on NN-based in-season CPE accuracy in unsurveyed regions, and (3) assess how NN-based CPE performance is affected by varied cropping practices in unsurveyed regions.

	\section{Study Regions and Datasets}
\begin{figure}[htbp]
		\centering
		\includegraphics[height=0.7\textheight]{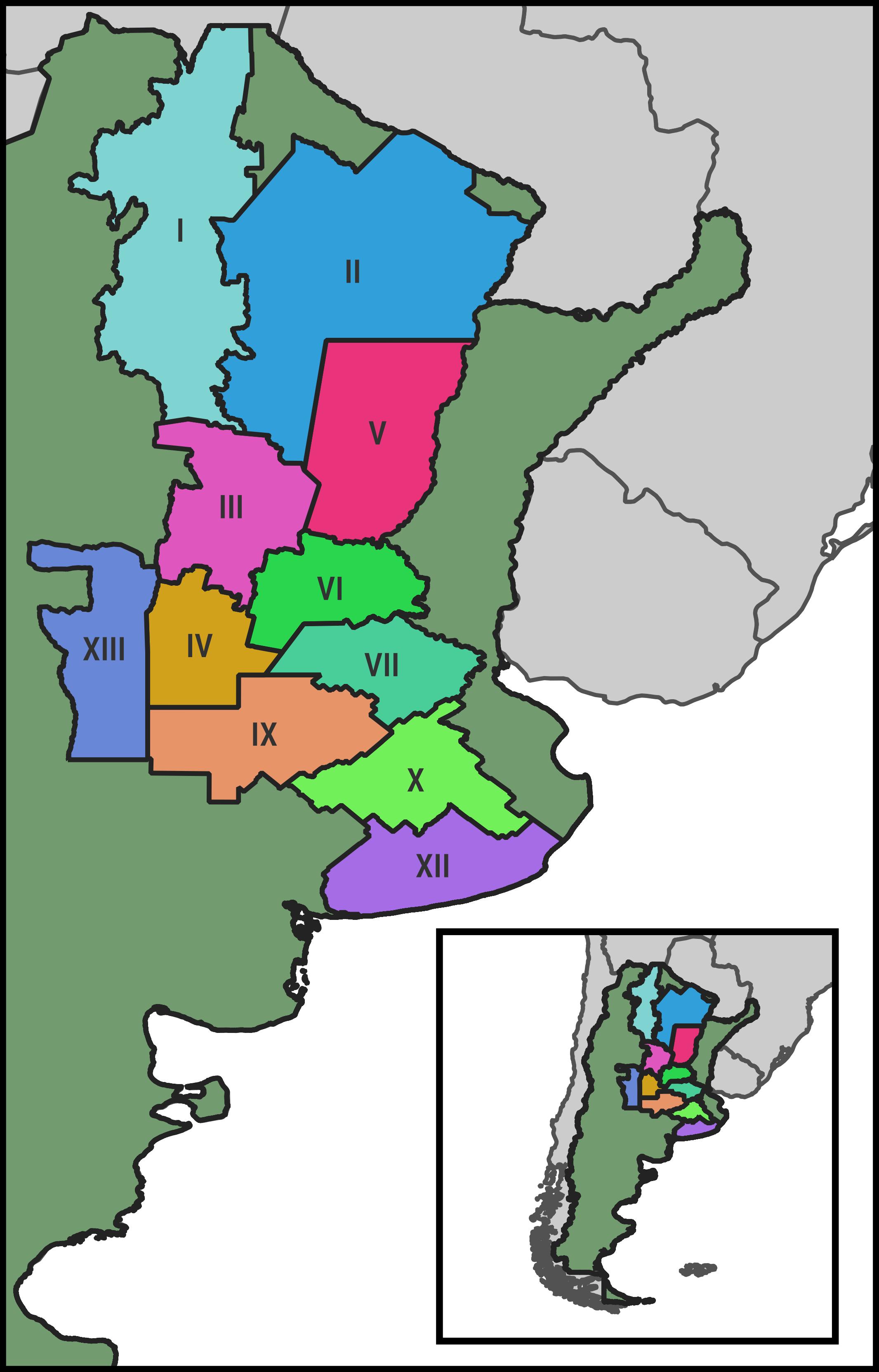}
		\caption{Bolsa de Cereales-defined corn growing zones in Argentina \citep{departamento_de_estimaciones_agricolas_guimetodologica_2020}.}
		\label{fig:arg_zones}
	\end{figure}

\begin{figure}[htbp]
		\centering
		\includegraphics[height=0.5\textheight]{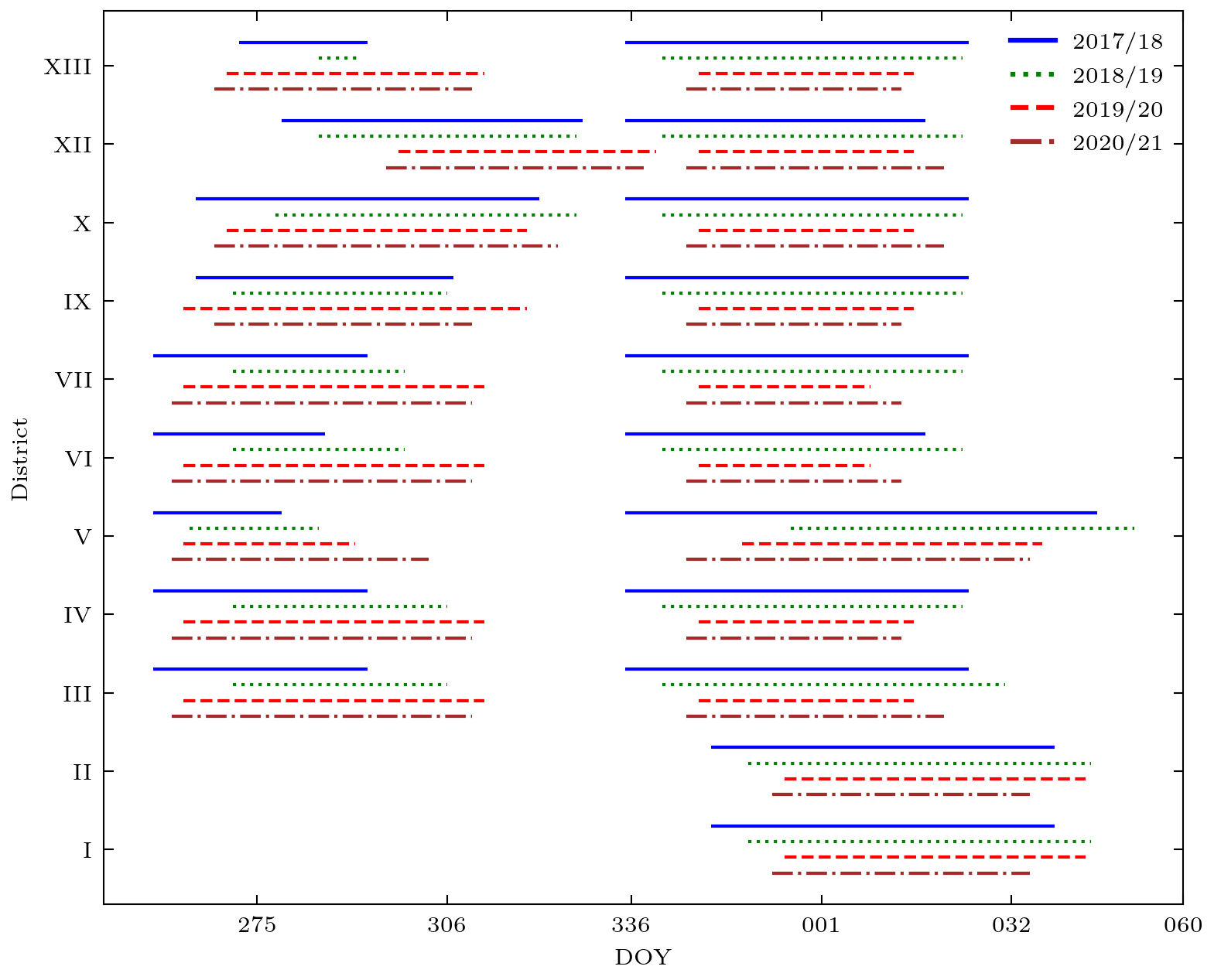}
		\caption{Early and late planting windows for corn growing zones in Argentina. Zones I and II have only one corn planting window.}
		\label{fig:arg_planting_windows}
	\end{figure}

\begin{figure}[htbp]
		\centering
		\includegraphics[width=0.7\textwidth]{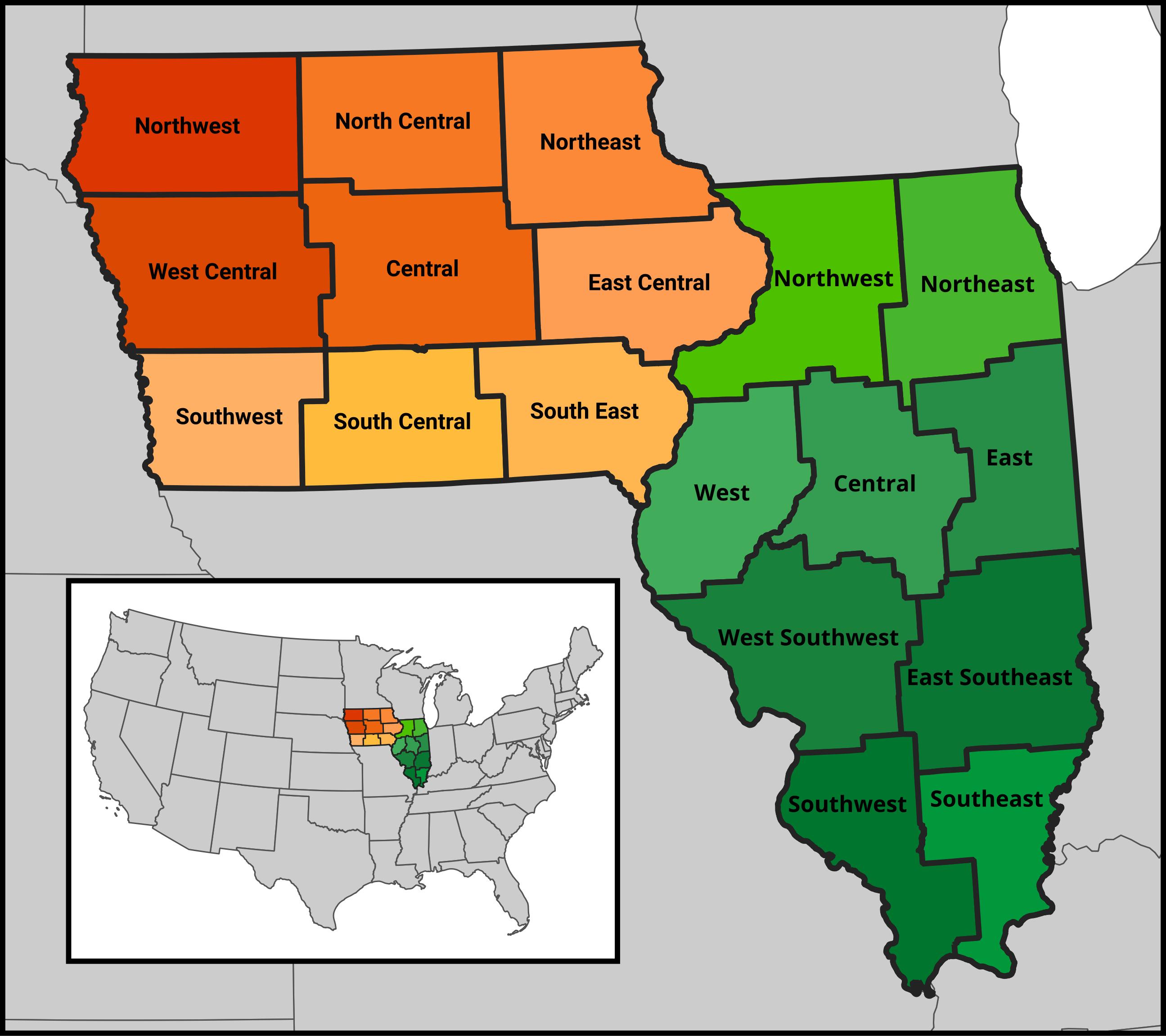}
		\caption{USDA-defined Agricultural Statistical Districts in Iowa (left) and Illinois (right) \citep{usda_national_agricultural_statistics_service_charts_nodate}.}
		\label{fig:midwest_districts}
	\end{figure}

	Argentina represented the `unsurveyed' region in this study as it is one of the few countries, other than the US, with publicly available weekly crop progress data that could be used for evaluation of the NN-based method. The states of Iowa and Illinois in the US Midwest represented the `surveyed' region. Both Argentina and the US are large exporters of staple crops such as corn and wheat. Corn, the second largest crop by global production, was used as the example crop \citep{fao_faostat_2022}.
	In Argentina, the majority of corn is grown in a two-crop system, in which an early and late-season crop are planted September-November and December-January, respectively. The Argentine Bolsa de Cereales (BdC) defines 13 growing zones, of which 11 are surveyed for corn production (see Figure \ref{fig:arg_zones}). Nine out of these 11 zones grow both an early and a late corn crop (see Table \ref{tab:early_late_split} and Figure \ref{fig:arg_planting_windows}), which creates bimodal corn crop progress across a growing season. Additionally, temperate climates in the central zones allows for near-year round production, including winter-summer double cropping and post-season cover crops. 
	In contrast, in the US Midwest, corn is grown in a single crop system from April to November, with an April-May planting window in each of the eighteen USDA-defined districts in Iowa and Illinois (see Figure \ref{fig:midwest_districts}). The differences in corn cultivation between Argentina and the US Midwest make Argentina an excellent disparate `unsurveyed' test case for assessing the impacts of varied climate and cropping practices on the performance of the in-season CPE methods developed in this study.
	
	\subsection{Datasets}
	
	This study used crop progress data from ground surveys, satellite RS and meteorological data to monitor corn crop progress, and soil datasets  to parameterize biophysical model simulations.
	
	\subsubsection{Crop Progress Data}
	In Argentina, weekly crop progress data is collected by both ground surveys and telephone surveys for each zone by the BdC for both the early and late season corn crops \citep{departamento_de_estimaciones_agricolas_guimetodologica_2020}. Growth stages recorded by the BdC include, expansión foliar (leaf expansion), panojamiento (tasseling), floración femenina (silking), grano pastoso (dough), madurez fisiológica (mature), and cosecha (harvested). Argentine growing seasons traverse the new year, and crop progress reports can span September to September for zones in which the climate is favorable for year-round production. In this study, crop progress data for the Argentine growing seasons of 2019/20 and 2020/21 were \textit{only used for evaluation of the in-season CPE method}, and the NN-based CPE method was never exposed to real Argentine crop progress data during NN development and hyperparameter selection. In the US Midwest, crop progress data for corn is collected weekly by ground surveyors from the USDA at Agricultural Statistical District, hereafter district, level throughout the growing season \citep{usda_national_agricultural_statistics_service_national_nodate}. Stages recorded include planting, emerged, silking, dough, dent, mature, and harvested. Only crop progress data from the US Midwest growing seasons for years 2003--2020 and generated synthetic data for Argentina were used to train the NN-based in-season CPE method.
	
	\subsubsection{Remote Sensing and Meteorological Data}
	\label{sec:rs_and_weather_data}
	Table \ref{tab:inputs} shows the remotely sensed data used as inputs to the NN-based CPE method, their resolutions and pre-processing.
	In this study, crop cover maps at 30 m of Argentina \citep{de_abelleyra_first_2020} for the 2019/20 and 2020/21 seasons and of the US \citep{usda_ag_data_commons_cropscape_2021} for the 2003--2020 seasons were used to identify corn pixels. Crop maps were upscaled to 250 m using pixel averaging to match the resolution of the MODIS NDVI product. All pixels that contained < 90\% corn cover were masked and the NDVI values from MODIS observations for each corn pixel were linearly interpolated to produce daily time series values for the pixel. These time series were then collated into histograms of daily NDVI values across each district/zone, with 20 histogram bins in the range 0 to 1. Negative NDVI values were set to 0. 
	
	Daily maximum and minimum air temperatures were used to form daily growing degree day (GDD) values for each district/zone, where GDD is calculated as
	\[
	GDD = 
	\begin{cases}
		\frac{T_{max} + T_{min}}{2} - T_{base} ,& \text{if } \frac{T_{max} + T_{min}}{2}\geq T_{base}\\
		0,              & \text{otherwise}
	\end{cases}
	\]
	with $T_{max}$ as the lesser of the daily maximum temperature and $34^\circ$C, $T_{base} = 8^\circ$C, and $T_{min}$ as the daily minimum temperature. GDD is a common measure used to model the effects of temperature on crop growth and development. GDD time series for each corn pixel were aggregated into daily histograms with 27 bins, ranging from 0 to 26$^\circ$C.
	
\begin{figure}[htbp]
		\centering
		\begin{subfigure}[t]{0.5\textwidth}
			\centering
			\includegraphics[width=\textwidth]{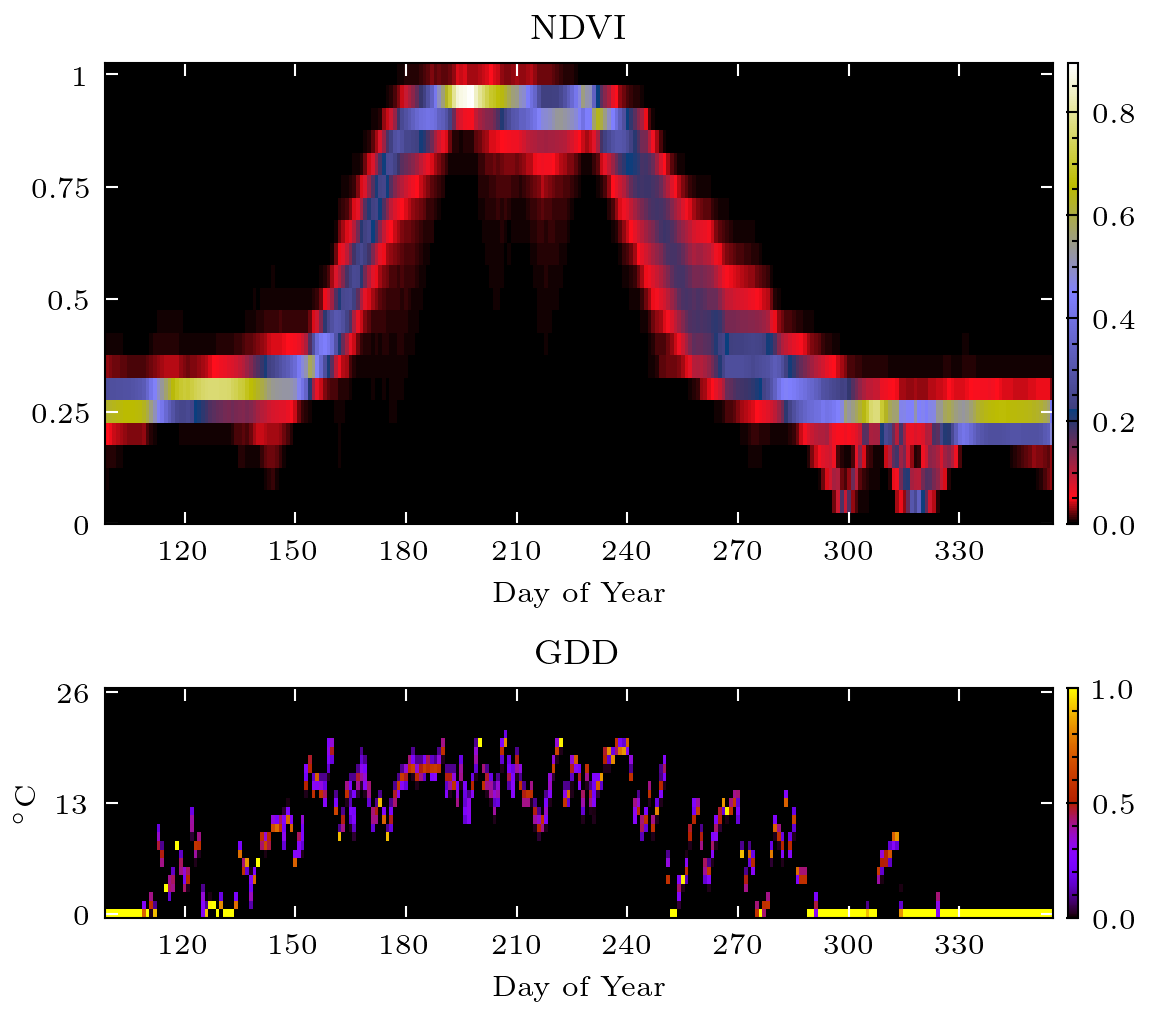}
			\caption{Northwest Iowa, US, 2020}
		\end{subfigure}%
		~ 
		\begin{subfigure}[t]{0.5\textwidth}
			\centering
			\includegraphics[width=\textwidth]{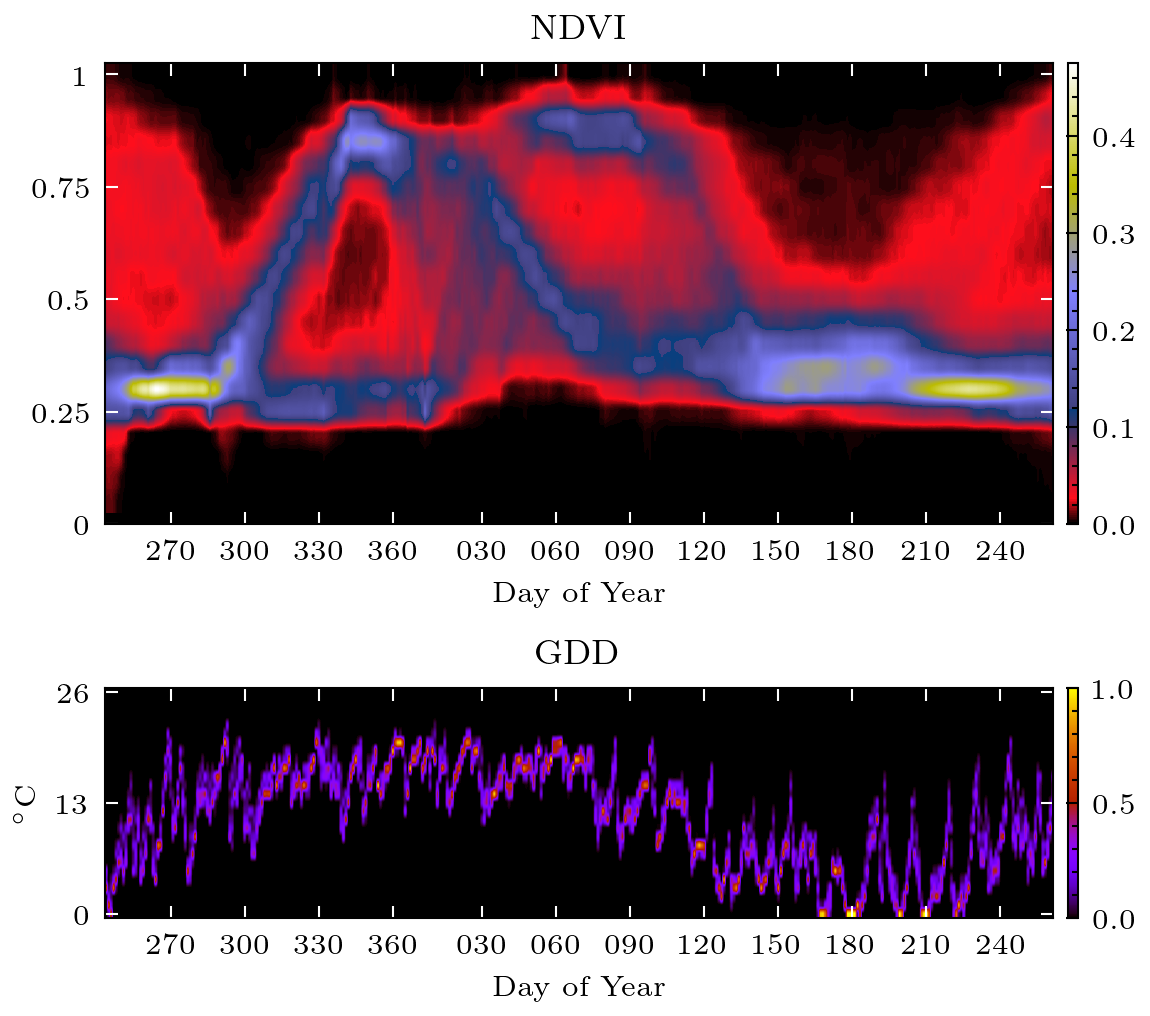}
			\caption{Zone V, Argentina, 2020/21}
		\end{subfigure}
		\caption{Example daily Normal Difference Vegetation Index (NDVI) and growing degree day (GDD) histograms inputs from (a) the US and (b) Argentina. NDVI noise related to winter-summer cropping systems and post-season cover cropping can be observed at the beginning and end of the NDVI histogram in (b).}
		\label{fig:input_examples}
	\end{figure}

	 Histograms of NDVI and GDD time series for each district/zone began two weeks before the first historical planting date for that region. Figure \ref{fig:input_examples} shows example of district/zone-level time series of NDVI and GDD for Northwest Iowa and Zone V. As shown in the figure, the NDVI values in Zone V exhibit two distinct peaks caused by the dual planting windows.
	
	\subsubsection{Soil Data}
	\label{sec:soil_data}
	Soil reflectance spectra samples and spatial data on soil physical properties were used to generate synthetic data. The soil properties dataset used to run the biophysical crop model simulations at 10 km resolution for each corn growing zone within Argentina was a global, pre-formatted soil database derived from SoilGrids \citep{hengl_soilgrids250m_2017} and HarvestChoice 27 \citep{koo_hc27_2013} produced by \cite{han_development_2019}. This dataset contained gridded texture, bulk density, drainage rate, root growth factor, soil organic carbon, and saturated hydraulic conductivity of soils to a depth 2 m at a spatial resolution of 10 km. 
	
	Soil reflectance data at wavelengths from 400 nm to 2500 nm at 1 nm spectral resolution were acquired from the NASA Ecosystem Spaceborne Thermal Radiometer Experiment on Space Station (ECOSTRESS) spectral library \citep{meerdink_ecostress_2019}. Surface soil properties from each 10 km grid within Argentina were matched to one of the 69 ECOSTRESS soil samples based on texture. Soil reflectance data from that sample were then used to simulate the soil component of synthetic optical reflectance for that grid.
		
	\section{Methodology}
	
	\subsection{Network Architecture}
\begin{figure}[htbp]
		\centering
		\includegraphics[width=0.8\textwidth]{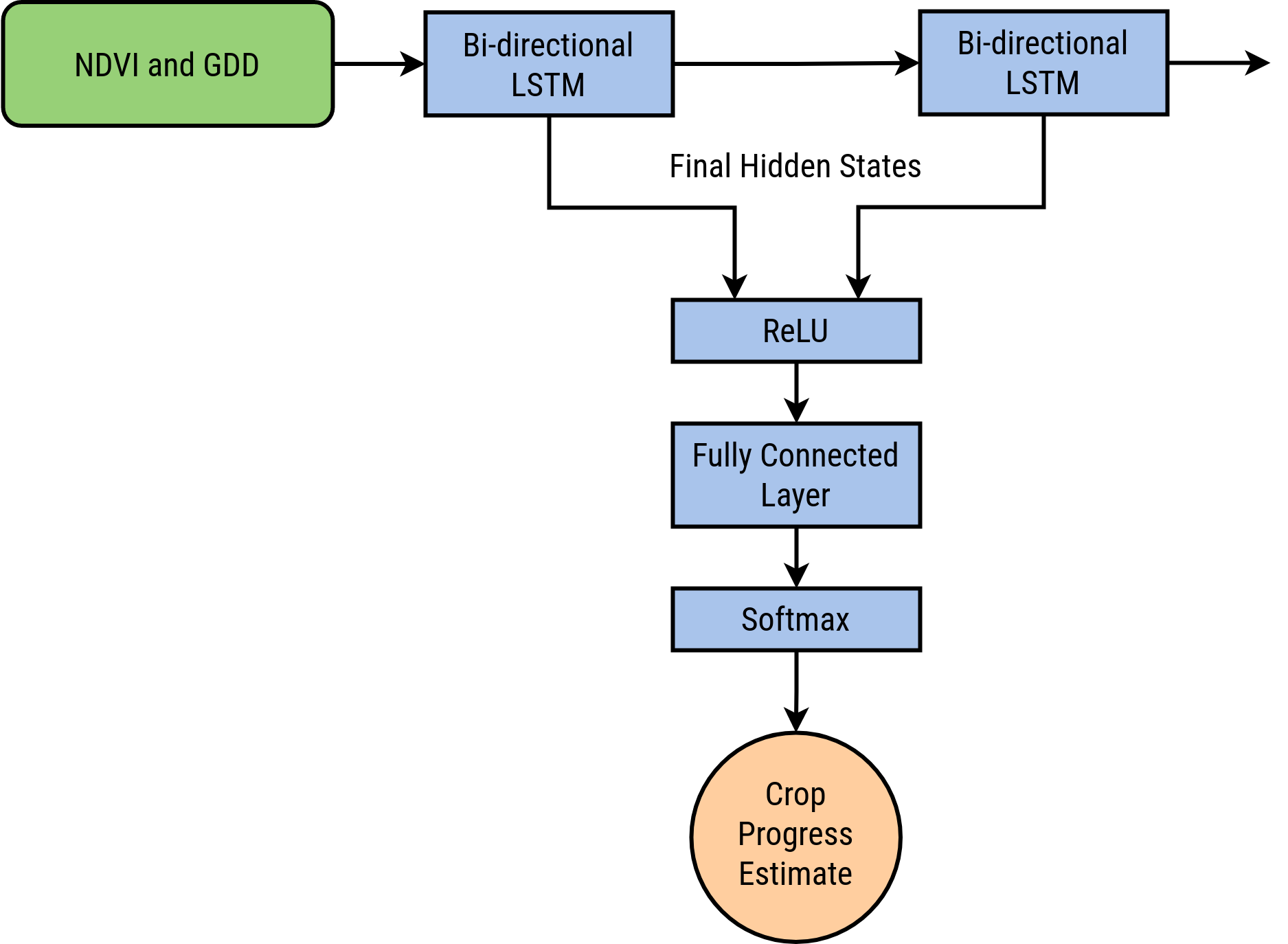}
		\caption{Structure of neural network used for in-season CPE in this study.}
		\label{fig:network_structure}
	\end{figure}

	The NN structure used in this study was based upon Long Short-Term Memory (LSTM), a network architecture which has been used in agricultural RS for yield prediction \citep{schwalbert_satellite-based_2020}, crop mapping \citep{kerner_phenological_2022}, and in-season CPE \citep{worrall_domain-guided_2021}. In this study, we used a simple, two-stacked bi-directional LSTM structure, with the final hidden states of each LSTM layer fed into a single fully connected layer to produce crop progress estimates, as shown in Figure \ref{fig:network_structure}. LSTM layers each had 128 memory cells and the fully connected layer had 6 nodes.
	
	The network used district/zone-level histograms of time series NDVI and GDD to estimate the crop progress distribution for each week, \textit{w}, during the season. Histograms were masked beyond week \textit{w} to simulate in-season estimation and prevent look ahead bias. Kulbeck-Leibler divergence was used for network loss, defined as:
	\begin{equation}
		D_{KL}(P||Q) = \sum_{x \in X} P(x) \log \left( \frac{P(x)}{Q(x)}\right) 
	\end{equation}
	where here $P(x)$ and $Q(x)$ are estimated and actual crop progress on week \textit{w}.  Each week's crop progress was treated as a separate target, and a softmax output layer was used to ensure estimates across all stages for a given week summed to 100\%.

	\subsection{Synthetic Data Generation}
	
	Three models were linked to generate synthetic crop progress data on a 10 km grid within each of the 11 corn-growing zones in Argentina. These were a spatial version of the Richardson-type stochastic weather generator (SWG), as proposed by \cite{wilks_multisite_1998, wilks_simultaneous_1999}, the CSM-IXIM corn model from the Decision Support System for Agro-technology Transfer (DSSAT) crop model suite \citep{lizaso_csmixim_2011}, and the PROSAIL-style RTM Soil-Plant-Atmosphere Radiative Transfer (SPART) model \citep{yang_spart_2020}. 
	
	\subsubsection{Spatial SWG}
	\label{sec:SWG}
	The spatial SWG is a generative weather model that produces spatially correlated precipitation, maximum and minimum temperature, and solar radiation data at daily time steps. The SWG typically uses historical weather data from weather stations in a target region to produce synthetic weather data which preserves both monthly mean, variance, and inter-annual variability of weather variables for individual stations as well as historical correlations between stations. 
	
	In the SWG, precipitation occurrence is modeled as a first order Markov chain and precipitation amount is modeled as a dual-exponential distribution of the form,
	\[
	f[r(k)] = \frac{\alpha}{\beta_1(k)} \exp\left[\frac{-r(k)}{\beta_1(k)}\right] + \frac{1 - \alpha}{\beta_2(k)} \exp\left[\frac{-r(k)}{\beta_2(k)}\right]
	\]
	
	where $\alpha$ is the mixing parameter and $\beta_1$, and $\beta_2$ are distribution parameters for weather station $k$. Daily solar radiation and maximum and minimum temperatures are generated using a first-order multi-variate autoregressive model
	\begin{equation}
		z_t = [\Phi] z_{t-1} + [B] e_t
	\end{equation}
	where $z_t$ is a vector of the residuals of solar radiation and maximum and minimum temperatures, $\Phi$ is a $ 3 \times 3$ matrix of autoregressive parameters, and $B$ is a $ 3 \times 3$ matrix of coefficients which multiply the standard Gaussian random numbers in $e_t$ that drive the model. Autoregressive parameters and correlations between variables are calculated at each station for each month of the year, with different parameters fit for wet and dry days. The SWG generates spatially correlated weather data by using historical correlations between weather at different stations to condition a stream of random numbers which drive weather generation at individual station models. Inputs to the SWG are shown in Table \ref{tab:simulation_inputs} and the SWG implementation used in this study is available on GitHub \citep{worrall_spatial_2022}. 
	
\begin{figure}[htbp]
		\centering
		\includegraphics[width=0.8\textwidth]{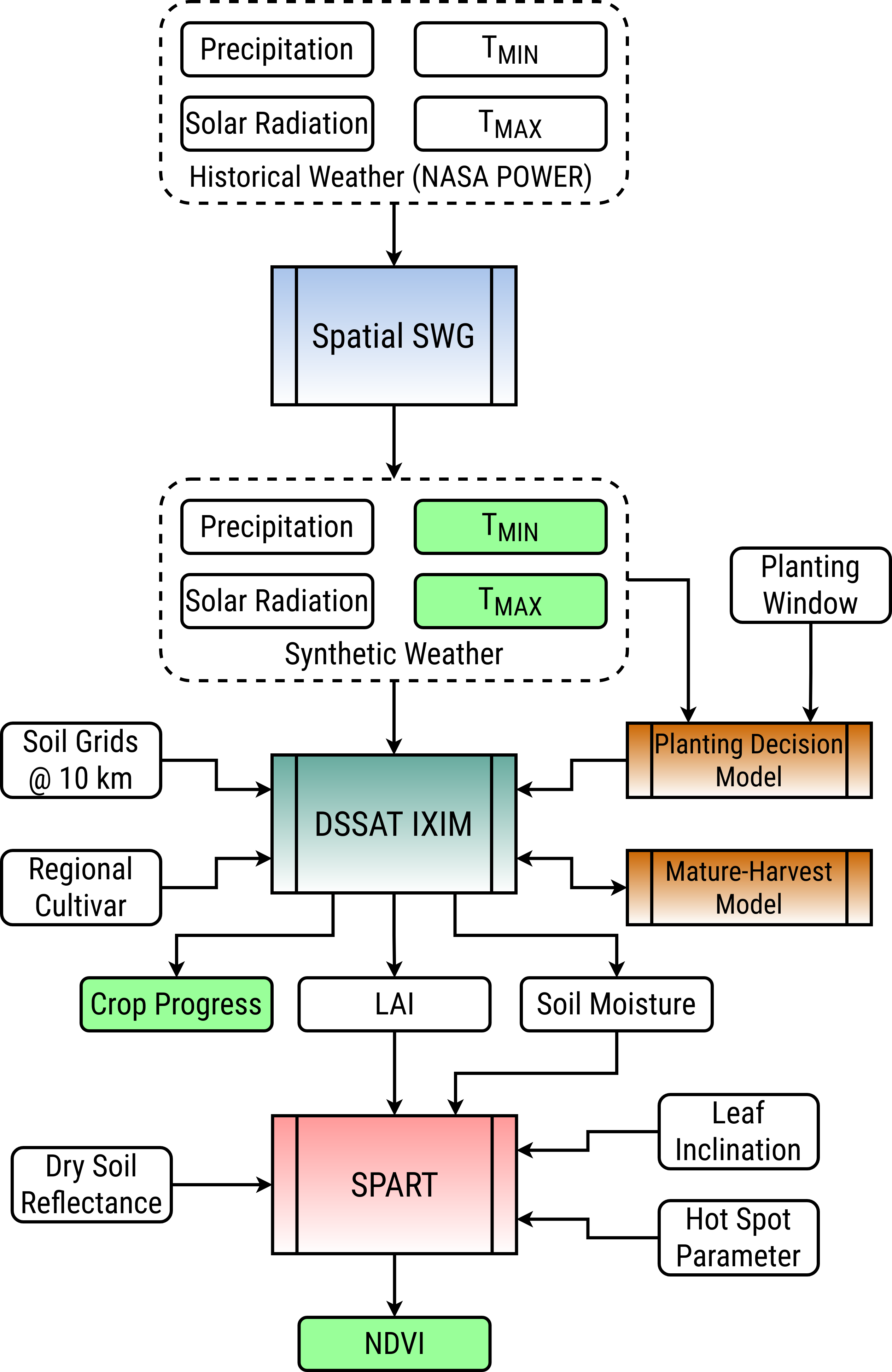}
		\caption{The linked spatial stochastic weather generator (SWG), crop model (DSSAT IXIM), and optical radiative transfer model (SPART) used for synthetic crop progress data generation in this study. The synthetic dataset includes the variables highlighted in green.}
		\label{fig:simulation_chain}
	\end{figure}

	In order to produce synthetic crop progress data which retains the spatial correlations present in real crop progress, synthetic weather used to drive the linked crop model-RTM (see Figure \ref{fig:simulation_chain}) in a zone must be spatially correlated. In this study, zone borders acted as synthetic data generation boundaries, and the SWG was calibrated on historical weather data from within each zone. NASA POWER 0.5 deg gridded weather pixels were used as weather stations, with the weather archive from each of these pixels (1984--2022) used as SWG calibration data \citep{nasa_langley_research_center_nasa_nodate}. Individual station parameters for precipitation, $\alpha$, $\beta_1$, and $\beta_2$,  were derived for each NASA POWER weather pixel using maximum likelihood estimation, and spatial correlations between pixels within a zone were identified following the method from \cite{wilks_simultaneous_1999}. The SWG was used to generate 100 years of synthetic gridded weather data for each zone for 99 full growing seasons. GDDs for each generated synthetic weather pixel were calculated using simulated maximum and minimum temperatures, then aggregated to zone level following the method outlined in Section \ref{sec:rs_and_weather_data}. Synthetic precipitation, solar radiation, maximum and minimum temperature data from each 0.5 deg weather pixel with a zone were used to drive DSSAT IXIM crop model simulations at 10 km, as shown in Figure \ref{fig:simulation_chain}.

	\subsubsection{CSM-IXIM}
	\label{sec:IXIM}
	Crop model simulations were conducted at 10 km resolution within each zone using the CSM-IXIM maize crop model (hereafter, IXIM) within DSSAT, a crop modeling suite that simulates crop growth, phenological progression, and yield of different crops \citep{jones_dssat_2003}. IXIM is a corn model that was developed by \cite{lizaso_csmixim_2011} and is based upon the CERES-Maize model \citep{jones_ceres_1983}. The model inputs include soil (from Section \ref{sec:soil_data}) and weather conditions (from Section \ref{sec:SWG}), as shown in Table \ref{tab:simulation_inputs}, along with eight genetic and phenotypic parameters describing corn cultivars. The fundamental driver of canopy growth and phenological stage progression within IXIM is GDD, and phenological progress of corn is split into six stages: emerged, late juvenile, tasseling, silking, grainfill, mature, with an additional delay in late juvenile to tasseling progression related to cultivar day length sensitivity.
	
	To generate synthetic crop progress data in Argentina, a single corn cultivar, DeKalb DK 752, was used in all IXIM simulations. At the time of the study, this cultivar accounted for 80\% of corn seed sold within Argentina \citep{andres_ferreyra_linked-modeling_2001}. While it would have been possible to calibrate IXIM on BdC survey data for each district, a cultivar based upon the literature was used to preserve the `unsurveyed' nature of this study. Parameters for this cultivar were taken from a field study conducted in Argentina by \cite{andres_ferreyra_linked-modeling_2001}.
	
	An early and a late planting corn simulation was run for each 10 km grid with a district which has historically contained corn. Early and late planted simulations were then weighted by the four-year average of the distribution of early and late planting corn in each zone to preserve the proportion of early and late planted corn present in the zone-level aggregated data (Table \ref{tab:early_late_split}).	
	
	The historical planting window for each zone was defined as the time between the earliest and latest reported planting dates for each zone in BdC corn crop progress reports. A simple, weather-dependent planting decision model was developed to generate planting dates for each 10 km grid square to avoid fixed planting distributions, which would be learnable. DSSAT simulations were conducted in baresoil mode using the synthetic weather data, and days with average soil temperature to 30 cm depth of $\geq$ 7$^\circ$C and moisture between 40\% and 90\% of the soil's drained upper limit and permanent wilting point were considered suitable planting days. A planting date was sampled uniformly from the first ten candidate planting days after the earliest historical planting date within a zone. This random component staggered planting in adjacent grid squares and ensured synthetic planting windows were not unrealistically shortened. 
	
	Harvest dates for corn simulations were modeled as a function of simulated crop maturity date and a random additional $X$ days, where $X \sim\mathcal{N}(30,10)$, to simulate the in-field dry down period between physical maturity and harvest in corn. In-field dry down, which producers use to reduce grain moisture and avoid post-harvest drying costs, is affected by weather. Dry down is not simulated within IXIM and this normal distribution was derived from the historical distribution of median maturity to harvest dates in US Midwest CPRs. Though conditions in the US Midwest differ from those in Argentina, the main objective of incorporating this random component was to avoid fixed harvest distributions. 
	
	Simulated growth stage progression for each grid was aggregated to zone level to form a CPR-style progress percentage for each stage, as defined in Table \ref{tab:stages}. IXIM-defined growth stages `early/late juvenile' and `tasseling' were grouped as `emerged'. The `dough' stage, which is not explicitly demarcated with IXIM, was set as the point at which modeled grain weight reached 33\% of its final weight \citep{rl_nielsen_grain_2021}. Daily LAI and soil moisture values generated by IXIM were used to parameterize SPART canopy reflectance simulations, as shown in Figure \ref{fig:simulation_chain}.
	
	\subsubsection{SPART}
	The SPART model combines four constituent RTMs together to simulate top-of-atmosphere optical reflectances from soil, leaf-canopy, and atmosphere. The atmospheric component was not used in this study because MODIS NDVI products were atmospherically corrected. Inputs to the SPART model include soil and vegetation parameters, shown in Table \ref{tab:simulation_inputs}. The soil RTM within SPART is the Brightness-Shape-Moisture model, which describes all soil spectral reflectances as a combination of three `global soil vectors' derived from \cite{jiang_gsv_2019}. Wet soil reflectance is modeled as a combination of a thin water film on top of a dry soil to account for the wetness component. In this study, the global soil vectors approach was replaced with dry soil reflectance data from the ECOSTRESS soil reflectance library (see Section \ref{sec:soil_data}) assigned to each soil grid \citep{meerdink_ecostress_2019}. 
	
	The leaf-canopy component of SPART is the widely used PROSAIL coupling of the PROSPECT leaf and SAILH canopy RTMs \citep{jacquemoud_prospectsail_2009}. PROSPECT uses leaf water content, chlorophyll and pigment concentration, dry matter, and internal structure information to simulation leaf reflectance and transmittance across the 400 nm to 2500 nm spectrum at 1 nm resolution. In this study, the default PROSPECT parameters were used for all SPART simulations, shown in Table \ref{tab:simulation_inputs}. With LAI, leaf angle distributions, and canopy height information, SPART uses the SAILH model to translate simulated leaf reflectance to top-of-canopy canopy reflectance data by modeling light-canopy-soil interactions.  Information on leaf angle distributions and canopy height by plant age, not simulated by IXIM, were obtained from corn field experiments conducted in the US \citep{judge_impact_2021}.
	
	Daily SPART simulations were conducted for each 10 km grid for every day from IXIM simulation start (pre-planting) until harvest. Simulated canopy reflectances at 620--670 nm ($\rho_{Red}$) and 841--876 nm ($\rho_{NIR}$) matching those of MODIS bands were used to produce synthetic NDVI using:
	
	\begin{equation}
		NDVI = \frac{\rho_{NIR} - \rho_{Red}}{\rho_{NIR} + \rho_{Red}}
		\label{eqn:NDVI}
	\end{equation}
	
	Post-harvest NDVI values for each pixel were generated by sampling from the pre-planting NDVI values for that season, which had a soil moisture-dependent range of < 0.03. Synthetic NDVI data were then aggregated to zone level following the method outlined in Section \ref{sec:rs_and_weather_data}. 
	
	SPART simulations were the most computationally demanding part of the synthetic data generation process. Because of the large number of numerical integration operations within the PROPSECT submodule, canopy reflectance simulations took an average time of 195.1 m-sec. Reusing the default leaf reflectance spectra reduced this to 1.1 m-sec, lowering the computational time required to simulate the $\approx$ 400M NDVI pixels across all zones and seasons from 840 CPU-days to 5 CPU-days.
	
\begin{figure}[htbp]
		\centering
		\includegraphics[width=\textwidth]{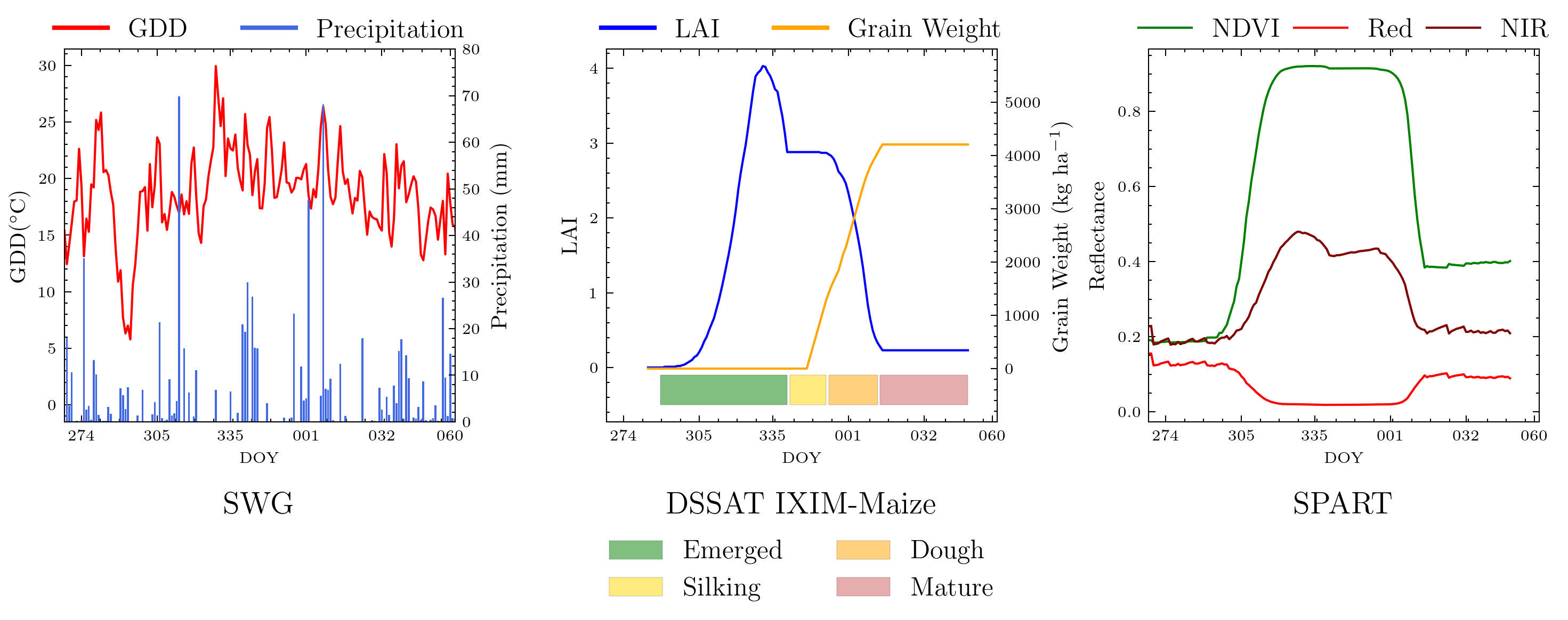}
		\caption{Example outputs produced by a linked SWG-IXIM-SPART simulation for one 10 km grid in Zone V in Argentina.}
		\label{fig:single_simulation_example}
	\end{figure}

\begin{figure}[htbp]
		\centering
		\begin{subfigure}{\textwidth}
			\centering
			\includegraphics[width=\textwidth]{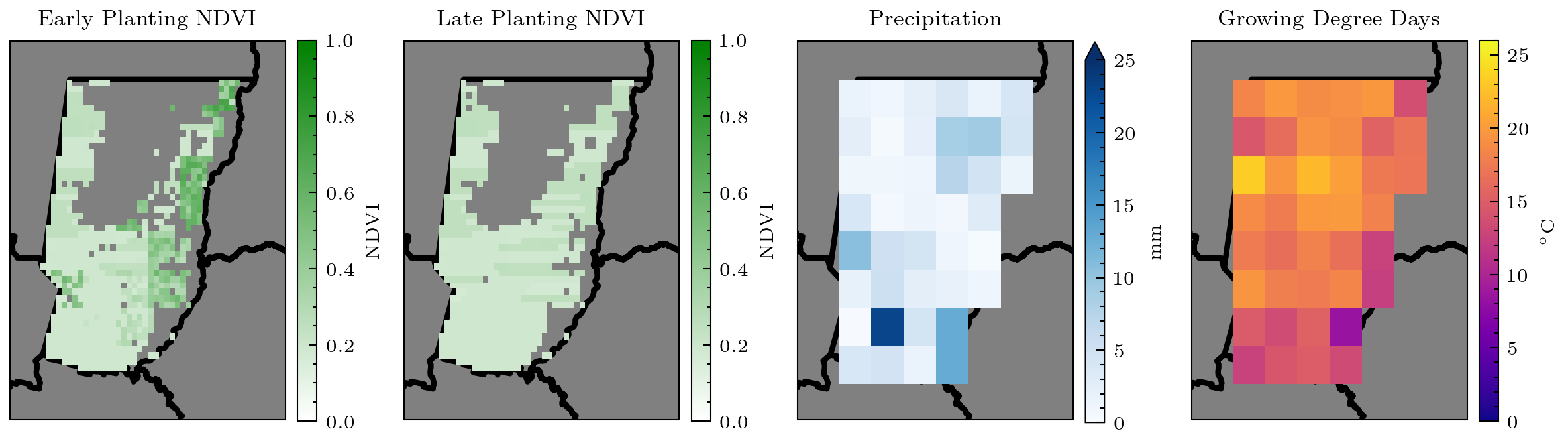}
			\caption{DOY 283}
		\end{subfigure}
		\begin{subfigure}{\textwidth}
			\centering
			\includegraphics[width=\textwidth]{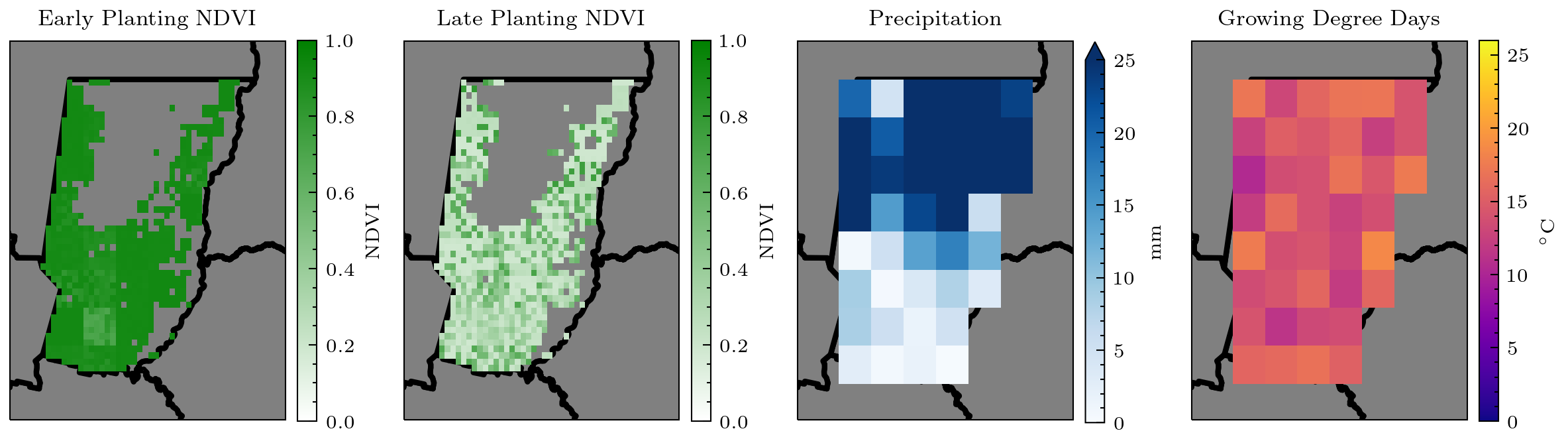}
			\caption{DOY 354}
		\end{subfigure}
		\begin{subfigure}{\textwidth}
			\centering
			\includegraphics[width=\textwidth]{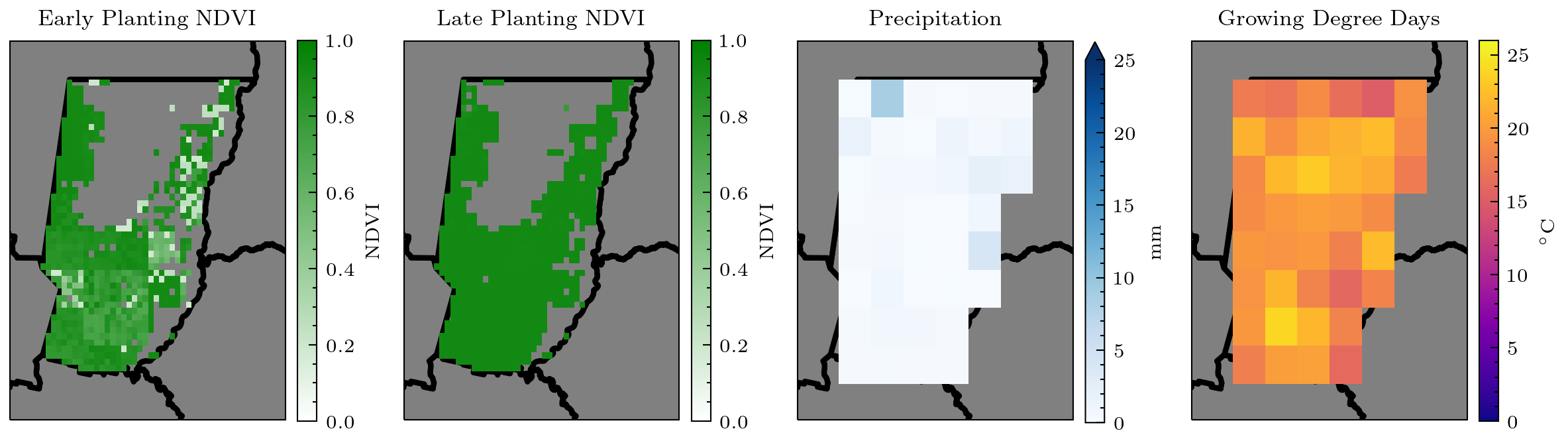}
			\caption{DOY 36}
		\end{subfigure}
		\begin{subfigure}{\textwidth}
			\centering
			\includegraphics[width=\textwidth]{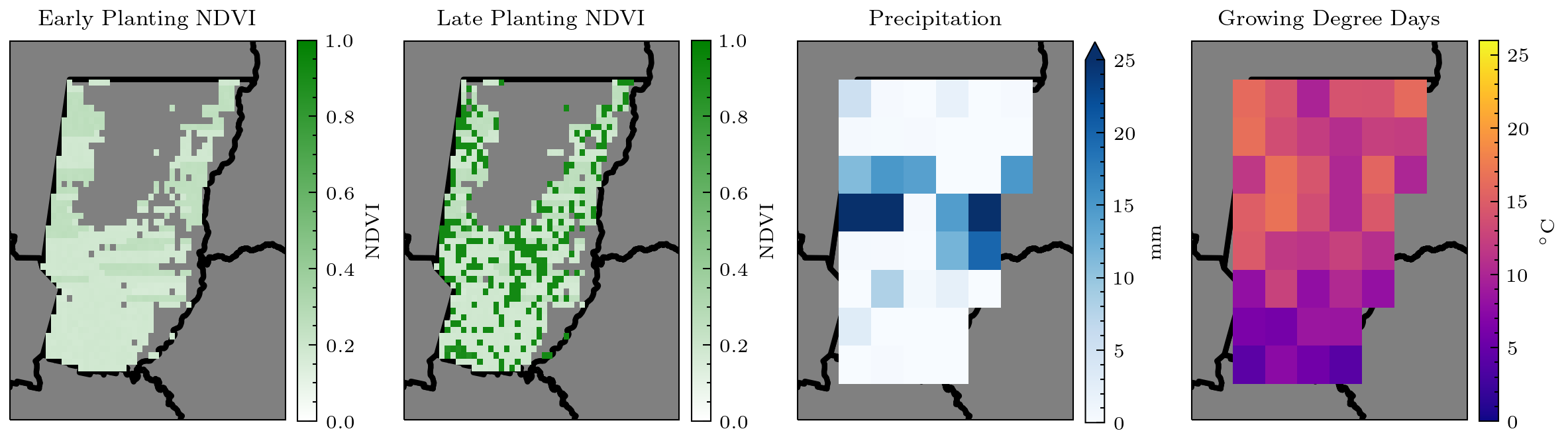}
			\caption{DOY 111}
		\end{subfigure}
		\caption[short]{Examples of NDVI, precipitation, and GDD from four samples days within a full season simulation generated by the linked SWG-IXIM-SPART models in Zone V in Argentina. In (a), emergence has begun in the north eastern part of the zone, while the late planting pixels have yet to be planted. In (b), early planting pixels are at the height of their canopy growth, while emergence begins in late planting pixels across the zone. In (c), some harvest of early planting pixels has begun in the north eastern part of the zone, while late planting pixels are at the height of their growth. In (d), harvest for early planting pixels is complete, while around half of late planting pixels have been harvested. Blank areas in the NDVI plots within the zone did not contain corn in the historical crop maps from \cite{de_abelleyra_first_2020}.}
		\label{fig:zone_V_simulation_example}
	\end{figure}

	Thus 99 seasons of synthetic GDD, crop progress, and NDVI were produced for the 11 corn growing zones in Argentina. Figures \ref{fig:single_simulation_example} and \ref{fig:zone_V_simulation_example} show example synthetic GDD, crop progress, and NDVI generated by the linked SWG-IXIM-SPART models for a single pixel and across a zone, respectively.
	
	\subsection{Network Training}
	\label{sec:data_and_stopping}
	
	\subsubsection{Training Sets}
	We investigated the effect of exposure to surveyed and synthetic crop progress data on NN performance in Argentina by implementing four different training data combinations. The first training dataset, U$_{sur}$, contained data from the US surveyed regions of Iowa and Illinois, over the years 2003-2020 and 2003-2017, respectively. This dataset consisted of 315 district-season combinations of crop progress data and associated NDVI and GDD histograms. This provided a measure for the NN's transferability to an unsurveyed region. The second set, A$_{syn}$, contained synthetic crop progress data produced for the 11 corn-growing zones in Argentina. This dataset contained 1188 zone-season combinations of synthetic crop progress data and associated synthetic NDVI and GDD histograms.
	
	The third set, A$_{syn1}$U$_{sur}$, combined the surveyed region data with synthetic data generated for a single, target zone in Argentina. This dataset produced 11 separate training data combinations, each of which contained the surveyed region data and synthetic data generated for one of the 11 Argentine zones. The fourth set was the union of the surveyed region data from the US and synthetic data for all 11 zones in Argentina, A$_{syn}$U$_{sur}$.
	
	In the first three datasets, all data were combined and training batches were sampled randomly across all available data.
	
	When training on the fourth dataset, A$_{syn}$U$_{sur}$, each batch of synthetic data sampled from A$_{syn}$ during training was paired with a batch of Midwest data sampled with replacement from the U$_{sur}$ data.  The network loss on each of these batches was averaged such that the loss back-propagated through the network for that batch was $loss = \frac{loss_{U_{sur}} + loss_{A_{syn}}}{2}$. This prevented overexposure to synthetic data (80\% of the A$_{syn}$U$_{sur}$ set) during training, which would have encouraged the network to optimize for the underlying IXIM model.
	
	To provide a surveyed region reference (SRR), the network was also trained on the U$_{sur}$ dataset with the 2018--2020 seasons, which contain only Iowa data, excluded. The SRR was then tested on Iowa districts for these seasons to provide a surveyed region performance reference.
	
	\subsubsection{Training Methodology}
	During training, the Adam optimizer was used with a learning rate of $1e^{-3}$, an L2-regularizing weight decay factor of $1e^{-3}$, and a batch size of 256  \citep{kingma_adam_2017}. For this study, all hyperparameters were kept the same across all training sets. When training the network, the best set of weights produced over 100 epochs were selected. Because no unsurveyed region validation data would be available during training, weight selection criteria must be based on performance on surveyed and synthetic validation data. For the datasets which contained only one type of data, U$_{sur}$ and A$_{syn}$, the training data was split into 80/20 training and validation subsets, separated by growing season. The weights produced by the epoch with the lowest validation loss were selected as the final network weights for each set.
	
	For the training sets which contained a mix of surveyed and synthetic data,  A$_{syn1}$U$_{sur}$ and A$_{syn}$U$_{sur}$ , we used the same training/validation split as above. For these sets, we selected the weights produced by the epoch with the lowest weighted divergence in validation loss across both the U$_{sur}$ and A$_{syn}$/A$_{syn1}$ validation subsets, with weighted divergence as: 
	\begin{equation}
		\text{D} = |\text{VL}_{U_{sur}} - \text{VL}_{A}| \times (\text{VL}_{U_{sur}} - \text{VL}_{A}) / 2
	\end{equation}
	
	where VL$_{U_{sur}}$ is the loss for the U$_{sur}$ validation subset and VL$_{A}$ is the validation loss for the A$_{syn}$ or A$_{syn1}$ validation subsets during training.
	
\begin{figure}[htbp]
		\centering
		\includegraphics[width=0.8\textwidth]{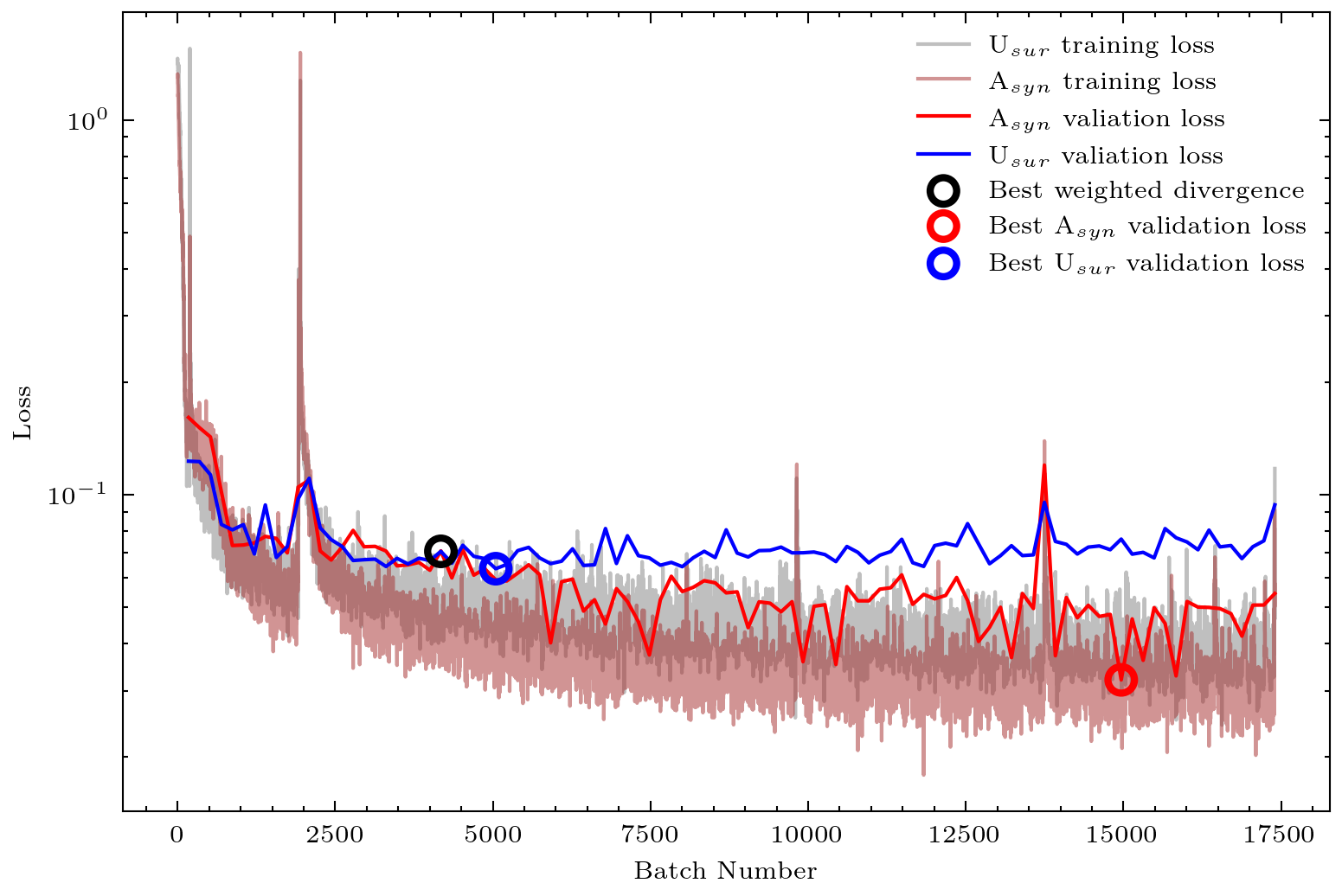}
		\caption{Training and validation loss for the NN-based in-season CPE method trained on the A$_{syn}$U$_{sur}$ dataset. The three circles show the locations of the best weighted divergence measure and validation losses for the A$_{syn}$ and U$_{sur}$ data outlined in Section \ref{sec:data_and_stopping}.}
		\label{fig:stopping_criteria}
	\end{figure}

	Figure \ref{fig:stopping_criteria} shows the location of the lowest weighted divergence for the A$_{syn}$U$_{sur}$ training loss curve, with the lowest surveyed and synthetic validation loss values over 100 epochs also labeled. This selection strategy was implemented as a trade-off between performance on surveyed and synthetic Argentina data. It was based upon the hypothesis that at some point during training (around batch 3500 in Figure \ref{fig:stopping_criteria}), the NN will reach a performance plateau that is achievable from learning generalizable relationships between NDVI, GDD, and crop progress which are valid for both the surveyed and synthetic data. After this point, it will begin to abandon crop progress notions in favor of optimizing for the underlying logic within the crop model used to generate synthetic data, in order to continue to reduce the loss on the combined training set. As the network begins to overfit to IXIM, which is deterministic for set parameters, the surveyed data validation loss will start to increase. This occurs around batch 4000 in Figure \ref{fig:stopping_criteria}, after which the U$_{sur}$ and A$_{syn}$ validation losses begin to diverge. The A$_{syn}$ validation loss continues to reduce in tandem with the training losses for both sets, while the U$_{sur}$ validation loss increases. In the absence of a true validation data from the unsurveyed region, we used the above weighted measure as a stopping criterion to identify the training epoch when this divergence between U$_{sur}$ and A$_{syn}$ validation loss began. 
	

	\subsection{Performance Metrics}
	Many existing studies in crop progress target median transition date into a given stage, and use RMSE of estimated versus observed median transition date as a performance metric, e.g. \cite{diao_remote_2020, gao_within-season_2020}. However, from a risk perspective, knowing the proportion of a region's crop in a given stage of crop progress at any time during the season is more descriptive. In this study, we adapt F$_1$ score, common in ML classification studies, to evaluate CPE method accuracy, penalizing both over and underestimates of crop progress for each stage for every week within a growing season. The adapted F$_1$ score in this study is defined as:
	
	\begin{equation}
		\label{eqn:F1}
		F_1 = \frac{\text{TP}}{\text{TP} + \frac{1}{2}(\text{FP} + \text{FN})}
	\end{equation}

	where TP, FP, and FN are true positive, false positive, and false negative rates, respectively, which are defined in terms of the overlap, overestimate, and underestimate of the NN-derived crop progress and the observed \textit{in situ} values, such that:
	
	\begin{equation}
		\begin{aligned}	
		\text{TP} = \sum^{W} \min \{CP^{est}_w, CP^{obs}_w\} \\
		\text{FP} = \sum^{W} \max \{CP^{est}_w - CP^{obs}_w, 0\} \\
		\text{FN} = \sum^{W} \max \{CP^{obs}_w - CP^{est}_w, 0\}
		\end{aligned}
	\end{equation}

	Where $CP^{est}_w$ is the crop progress estimate for a stage on week $w$ and $CP^{obs}_w$ is the observed crop progress from ground surveys. F$_1$ values can range from 0 to 1. An F$_1$ score of 1 indicates the network perfectly estimated the percentage of a district/zone's crop in that stage for every week in the season, while an F$_1$ score of 0 means there was no overlap at all between the network estimates of crop progress timing and \textit{in situ} crop progress through that stage. We measured overall performance through net F$_1$ score, calculated by summing TP, FP, and FN across stages and/or zones before computing Equation \ref{eqn:F1}.

	After being trained on each of the four training datasets, the NN was assessed on real Argentina crop progress data from the 2019/20 and 2020/21 growing seasons, which remained unseen by the network prior to method evaluation. The performance of the network when trained on these sets was compared with its performance when trained on the SRR and evaluated on Iowa districts during the 2018--2020 seasons.

	\section{Results and Discussion}

	\subsection{Overall Performance} 	
	Table \ref{tab:f1_overall} shows the net F$_1$ scores from both seasons and all zones for each training set and the reference set, SRR. Even though network performance in Argentina was lower for all training sets than performance in Iowa using the SRR, the F1 scores of 0.618--0.736 are significant considering the network did not encounter any real examples of crop progress in Argentina during training. The A$_{syn}$U$_{sur}$ set produced the best net F$_1$ scores for the Pre-emergence, Emerged, Silking, and Mature stages, but, interestingly, the A$_{syn1}$U$_{sur}$ set, which contains synthetic data only from the target zone, produced better scores for the Dough and Harvested stages (see Table \ref{tab:f1_overall}). Synthetic data generation for both of these stages relies on study-defined assumptions, which may explain why including only a single zone of synthetic data produced better results for these stages. For example, the assumptions for the Dough stage include its onset at 33\% grain weight and near-linear grainfill rate in unstressed conditions \citep{lizaso_csmixim_2011}, as shown in Figure \ref{fig:single_simulation_example}. For the Harvested stage, time between maturity and harvest is assumed to be normally distributed and independent of weather (see Section \ref{sec:IXIM}). It is possible that exposure to many examples of these assumptions negatively affected performance on these stages, but the limited examples in the A$_{syn1}$U$_{sur}$ set provided enough zone-specific crop progress information without overwhelming the weather-dependent relationships for these stages present in the surveyed region data. The additional synthetic data in the A$_{syn}$U$_{sur}$ set, which is weighted equally with surveyed region data during training, may encourage the network to adopt these synthetic data assumptions, thereby reducing performance.
	
	When trained on the U$_{sur}$ set, the net F$_1$ score was 8\% lower than when trained on the best overall performing set, as shown in Table \ref{tab:f1_overall}. Performance for U$_{sur}$ set was lower for all stages except the Dough stage, which was marginally higher. The largest performance gap for the network when trained on the U$_{sur}$ set was for the Mature stage, which was 19.5\% lower that the A$_{syn}$U$_{sur}$ set. The network also scored lower for the Mature stage when trained on the U$_{sur}$ set than when trained on the A$_{syn}$ set, which is interesting because this set does not contain any real crop progress data. This is possibly because the GDD to maturity of the synthetic Argentina cultivar is closer to current \textit{in situ} Argentina cultivars than what is grown in the US Midwest. This result shows the limitations in transferability of a network to unsurveyed region CPE when trained only on data from a surveyed region with different crop growth dynamics. 
	
	\subsection{Impact of Including Synthetic Data}	
	The major contribution of synthetic data is to provide examples of the expected cropping regimes from the target, unsurveyed region. In this study, the synthetic data provided examples of crop progress as produced by a dual planting cropping system. Including synthetic data during training, i.e. A$_{syn}$, A$_{syn1}$U$_{sur}$, or A$_{syn}$U$_{sur}$, increased F$_1$ scores for 57 out of 66 zone--stage combinations, and net F$_1$ scores across all stages increased in 9 out of 11 zones for both the 2019/20 and 2020/21 test seasons, shown in Tables \ref{tab:f1_2019} and \ref{tab:f1_2020}, respectively. Figures \ref{fig:F1_comparison_premergence}--\ref{fig:F1_comparison_harvested} show the F$_1$ scores by stages for each zone as a function of early and late planting percentage in those zones.
	
	\subsubsection{Single Planting}	
	The impact of synthetic data in single planting zones was mixed. For zones with $\geqslant$ 85\% of the corn crop planted in a single window, such as VI and VII, performance when trained on U$_{sur}$ data was similar to or better (e.g. Dough, see Figure \ref{fig:F1_comparison_dough} and Tables \ref{tab:f1_2019} and \ref{tab:f1_2020}) than performance when trained on the $A_{syn}U_{sur}$ and $A_{syn1}U_{sur}$ sets. This suggests that in single planting zones, imperfect synthetic data assumptions may sometimes outweigh the limited additional context gained beyond the examples present in the U$_{sur}$ set.
	
\begin{figure}[htbp]
	\centering
	\begin{subfigure}{.5\textwidth}
		\centering
		\includegraphics[width=\textwidth]{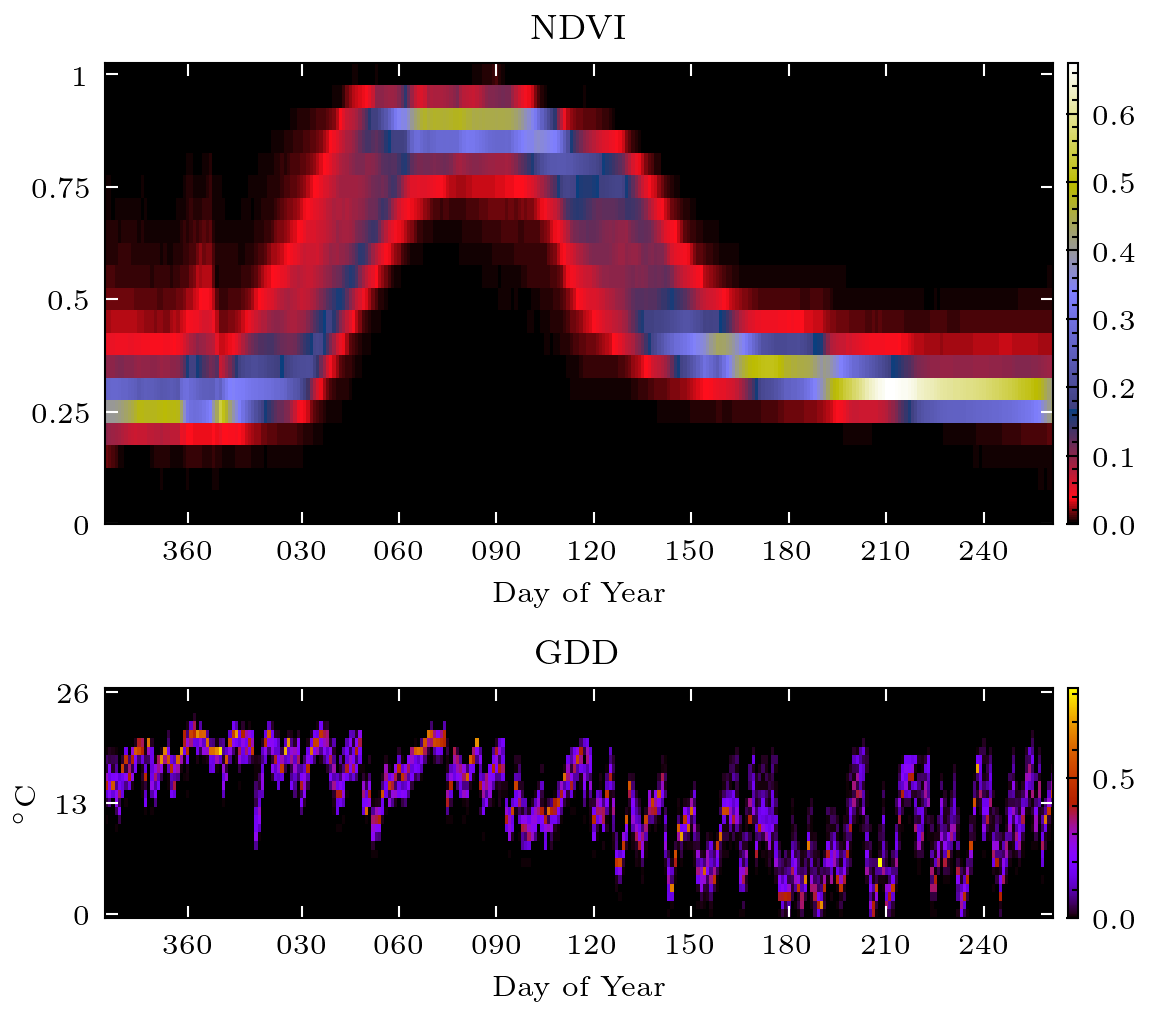}
		\caption{}
		\label{fig:input_I_1920}
	\end{subfigure}%
	\begin{subfigure}{.5\textwidth}
		\centering
		\includegraphics[width=\textwidth]{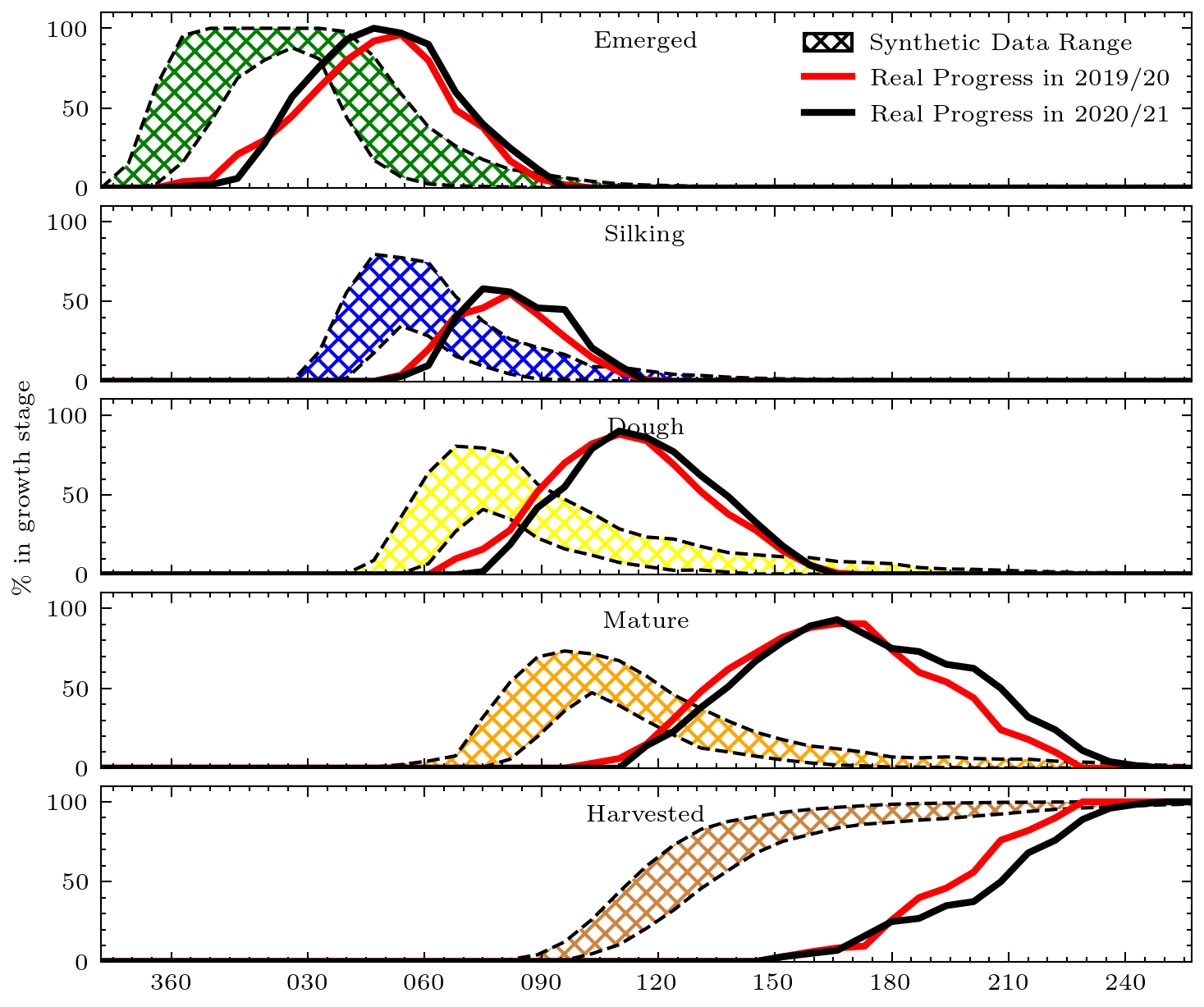}
		\caption{}
		\label{fig:result_I_1920_comp}
	\end{subfigure}
	\begin{subfigure}{.5\textwidth}
		\centering
		\includegraphics[width=\textwidth]{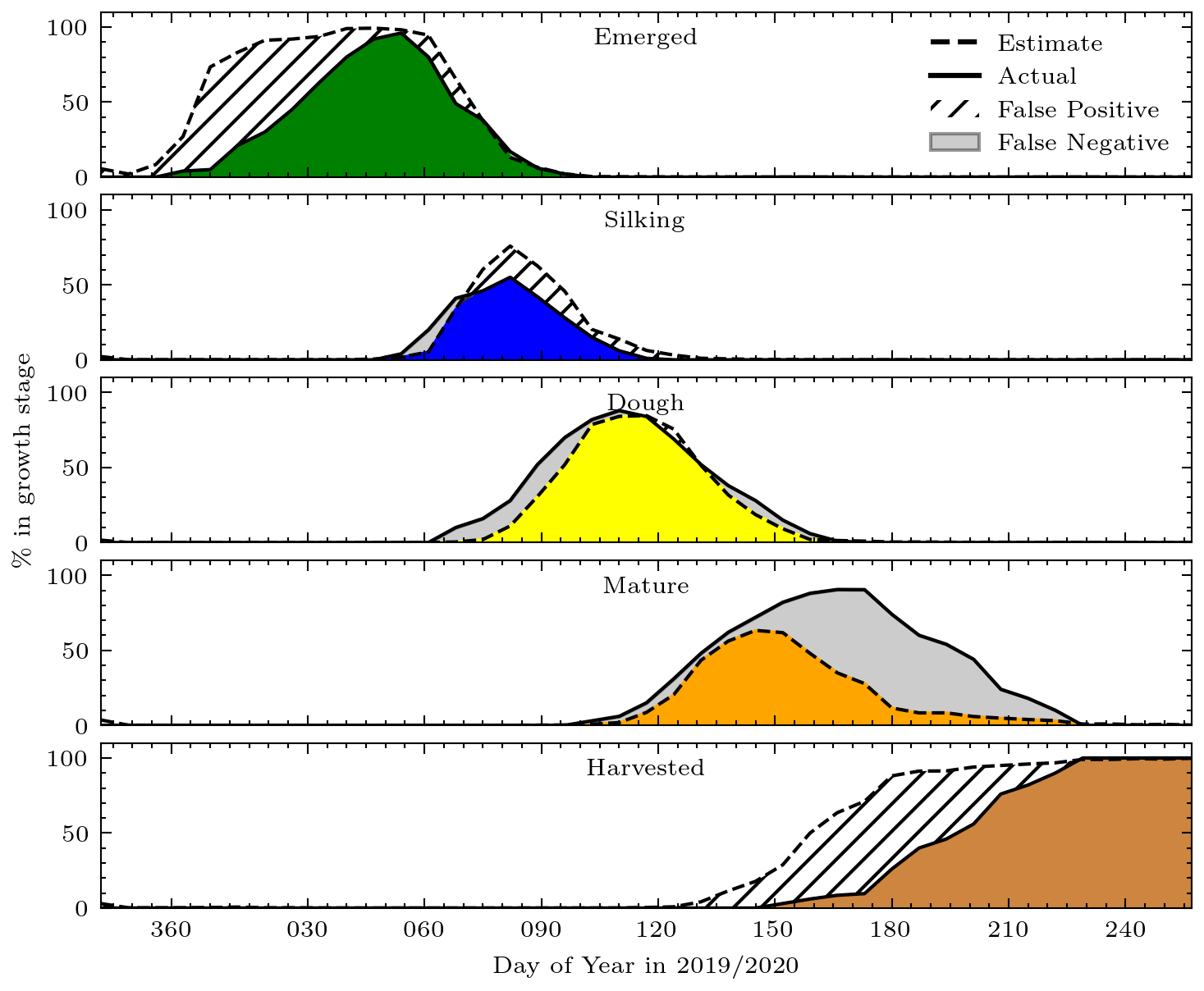}
		\caption{}
		\label{fig:result_I_1920_surveyed}
	\end{subfigure}%
	\begin{subfigure}{.5\textwidth}
		\centering
		\includegraphics[width=\textwidth]{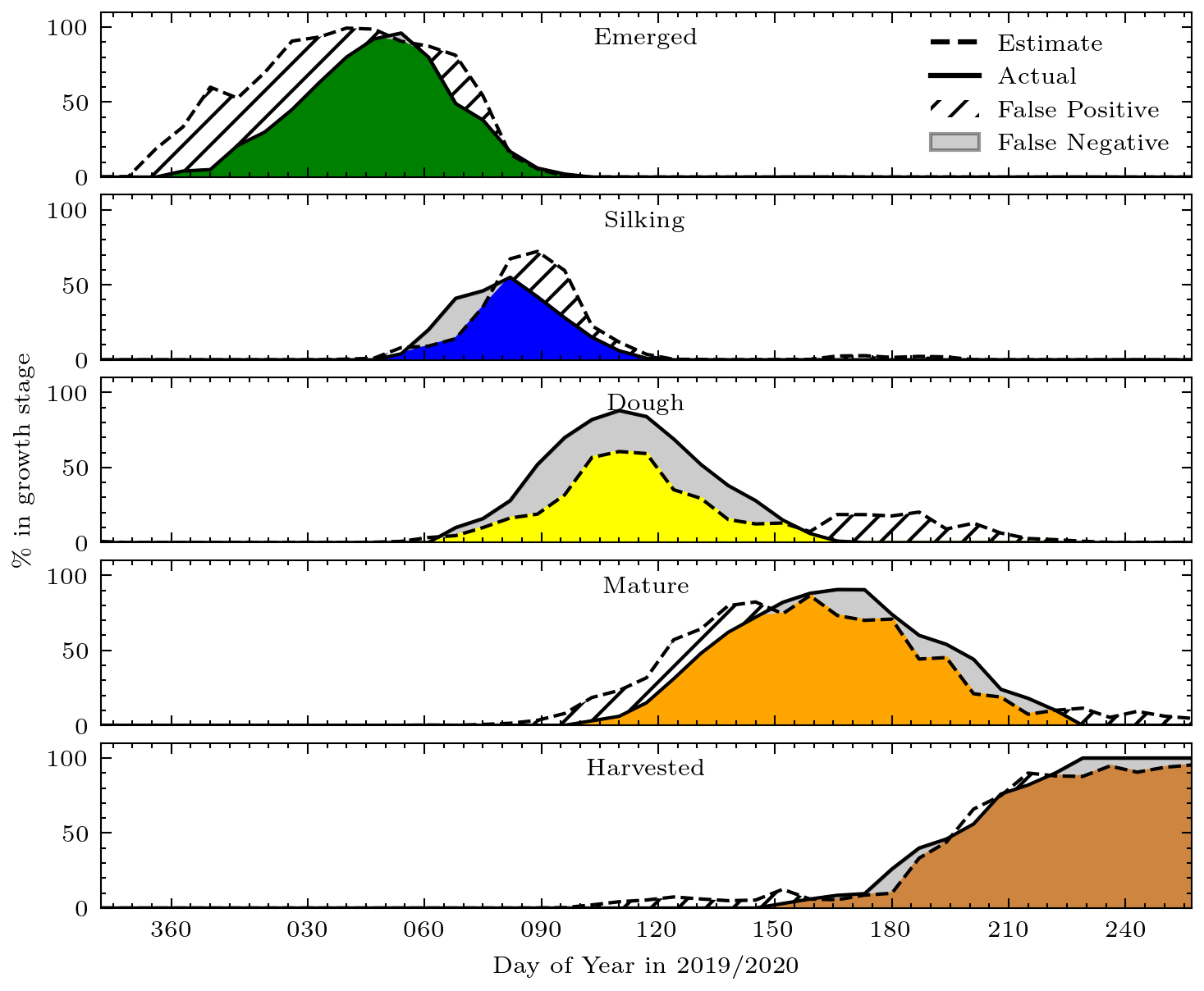}
		\caption{}
		\label{fig:result_I_1920_synth}
	\end{subfigure}
	\begin{subfigure}{.5\textwidth}
		\centering
		\includegraphics[width=\textwidth]{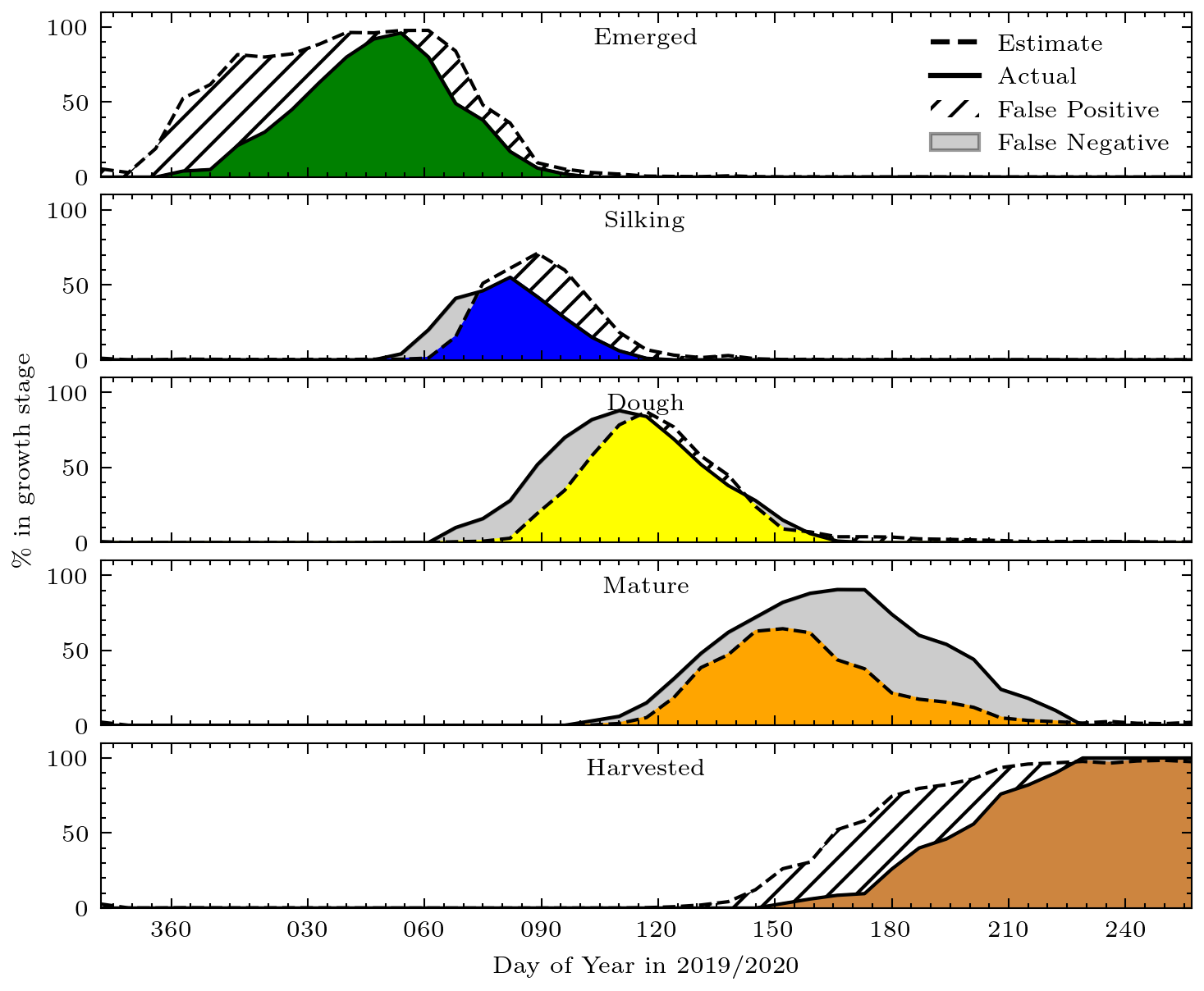}
		\caption{}
		\label{fig:result_I_1920_MSS}
	\end{subfigure}%
	\begin{subfigure}{.5\textwidth}
		\centering
		\includegraphics[width=\textwidth]{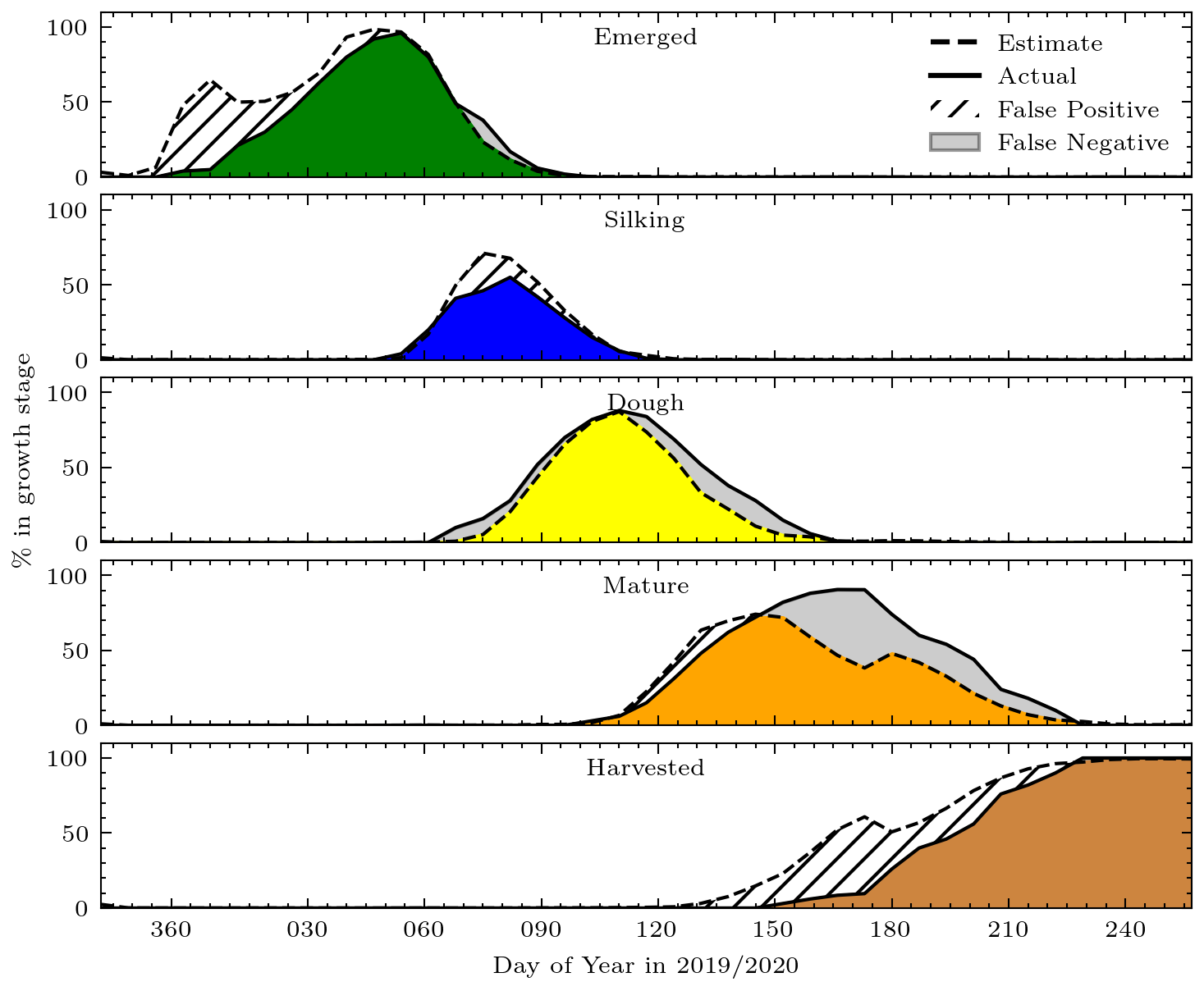}
		\caption{}
		\label{fig:result_I_1920_joint}
	\end{subfigure}
	\caption{Zone I during the 2019/20 season: (a) NDVI and GDD inputs, (b) test seasons and synthetic data range for crop progress, and crop progress estimates when the network was trained on (c) U$_{sur}$, (d) A$_{syn}$, (e) A$_{syn1}$U$_{sur}$, and (f) A$_{syn}$U$_{sur}$.}
	\label{fig:result_I_1920}
\end{figure}

	In Zone I, a 100\% single planting zone, the synthetic data improved network performance. Figures \ref{fig:input_I_1920}--\ref{fig:result_I_1920_joint} show the NDVI and GDD histograms, synthetic and \textit{in situ} crop progress data, and network crop progress estimates for Zone I during the 2019/20 test season, respectively. When trained on the A$_{syn}$U$_{sur}$ set, the network scored higher in Zone I than any other zone across all methods in both test seasons (see Tables \ref{tab:f1_2019} and \ref{tab:f1_2020}). The net F$_1$ score across all stages for this zone during 2019/20 was 0.845 when the network was trained on the A$_{syn}$U$_{sur}$ set, only 5.1\% lower than the net performance in Iowa for the SRR (Table \ref{tab:f1_overall}). For Silking, the most critical stage for corn in terms of yield loss, the F$_1$ score was only 0.7\% lower than the SRR net F$_1$ score for both the 2019/20 and 2020/21 test seasons. When trained on the U$_{sur}$ set, the network scored 9.6\% and 6.5\% lower than the A$_{syn}$U$_{sur}$ set on this zone, though performance on the Dough stage was the highest using U$_{sur}$ for both seasons (Tables \ref{tab:f1_2019} and \ref{tab:f1_2020}). The network also scored highest on the Mature and Harvested stages when trained only on the A$_{syn}$ set (see Table \ref{tab:f1_2019}). These high scores were due to a lower false negative rate on the Mature stage and a lower false positive rate on the Harvested stage, shown in Figure \ref{fig:result_I_1920_synth}.
	
	However, in Zone II, also a 100\% late planting zone, performance improvement from the addition of synthetic data was mixed for the two test seasons. During the 2019/20 season, net performance was highest using the U$_{sur}$ set at 0.723, marginally higher than the A$_{syn1}$U$_{sur}$ set. The addition of synthetic data from all zones, however, reduced performance during this season, particularly for the Silking stage, which was lower for both the A$_{syn}$ and A$_{syn}$U$_{sur}$ sets (Table \ref{tab:f1_2019}). During the 2020/21 season, net performance increased with the addition of synthetic data from all zones, but F$_1$ scores for the Silking and Dough stages were still higher when the network was trained on the U$_{sur}$ and A$_{syn1}$U$_{sur}$ sets.
	
	The mixed impact of including all synthetic data in these single planting zones is possibly due to the presence of dual planting examples in the A$_{syn}$ and A$_{syn}$U$_{sur}$ sets. When tasked with estimating unsurveyed region CPE in a single planting region, these dual planting examples have significantly less benefit and may cause the network to mistake any mid-season NDVI noise as the emergence of a second crop. Additional investigation would be needed to assess these effects, including increasing the proportion of single planting synthetic examples when targeting a single planting zone.
	
	\subsubsection{Dual Planting}
\begin{figure}[htbp]
		\centering
		\begin{subfigure}{.5\textwidth}
			\centering
			\includegraphics[width=\textwidth]{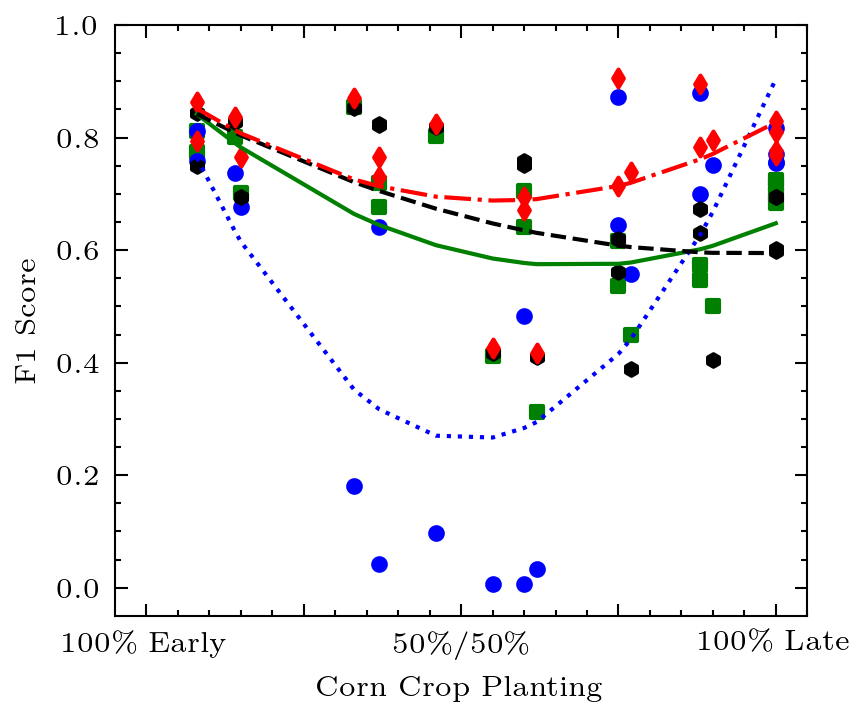}
			\caption{}
			\label{fig:F1_comparison_premergence}
		\end{subfigure}%
		\begin{subfigure}{.5\textwidth}
			\centering
			\includegraphics[width=\textwidth]{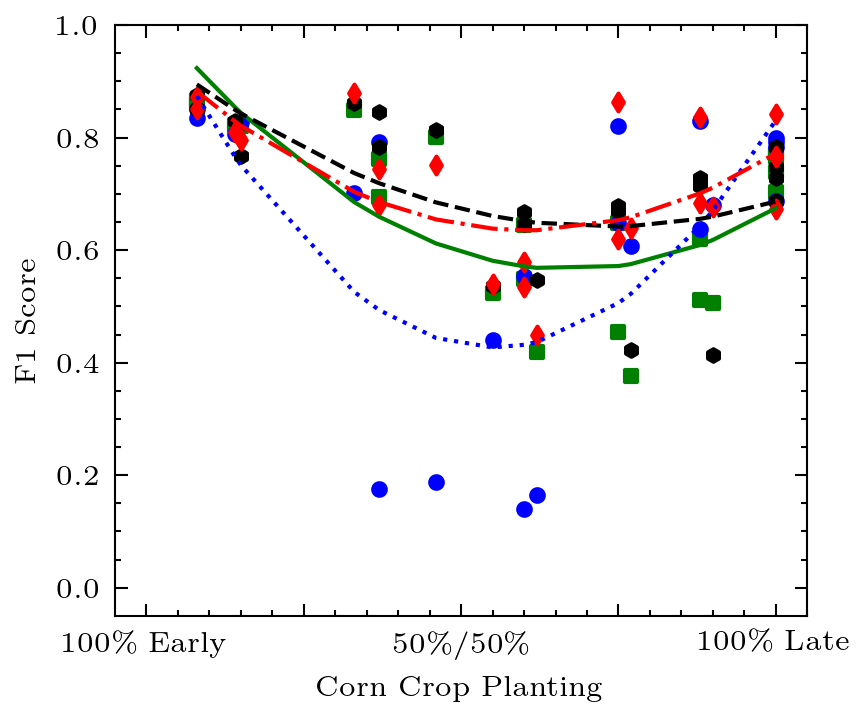}
			\caption{}
			\label{fig:F1_comparison_emerged}
		\end{subfigure}
		\begin{subfigure}{.5\textwidth}
			\centering
			\includegraphics[width=\textwidth]{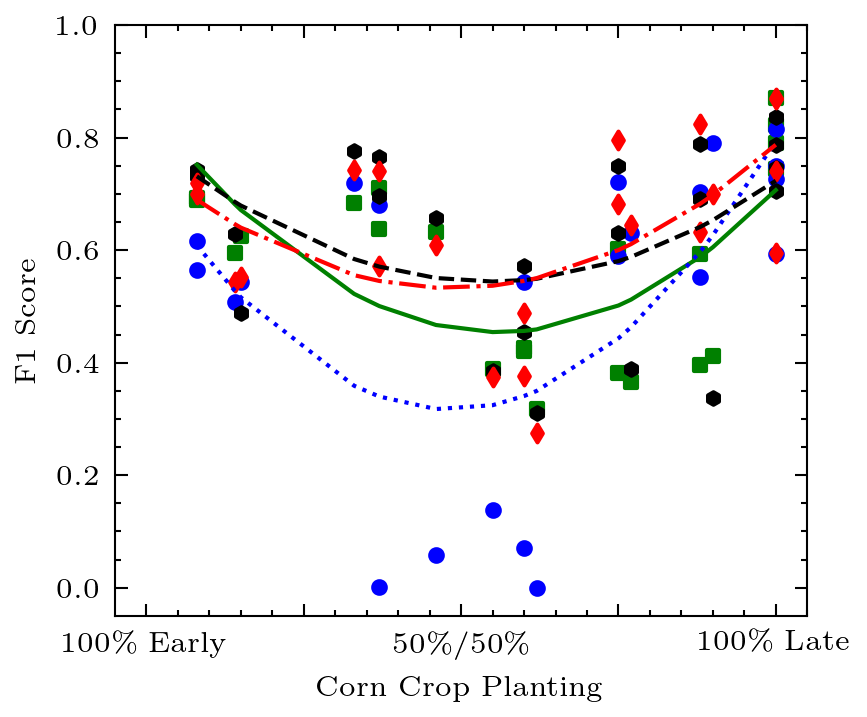}
			\caption{}
			\label{fig:F1_comparison_silking}
		\end{subfigure}%
		\begin{subfigure}{.5\textwidth}
			\centering
			\includegraphics[width=\textwidth]{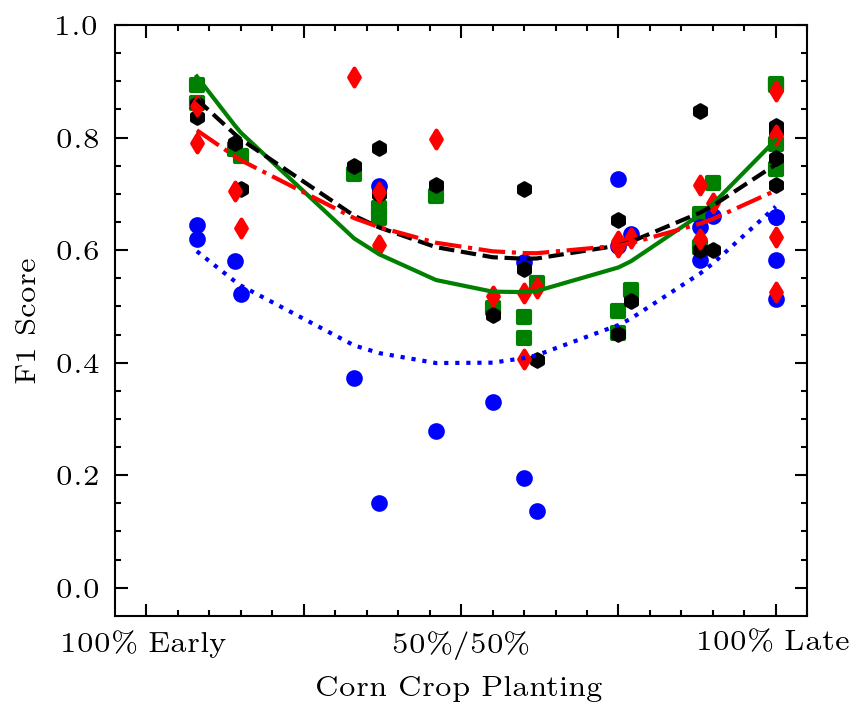}
			\caption{}
			\label{fig:F1_comparison_dough}
		\end{subfigure}
		\begin{subfigure}{.5\textwidth}
			\centering
			\includegraphics[width=\textwidth]{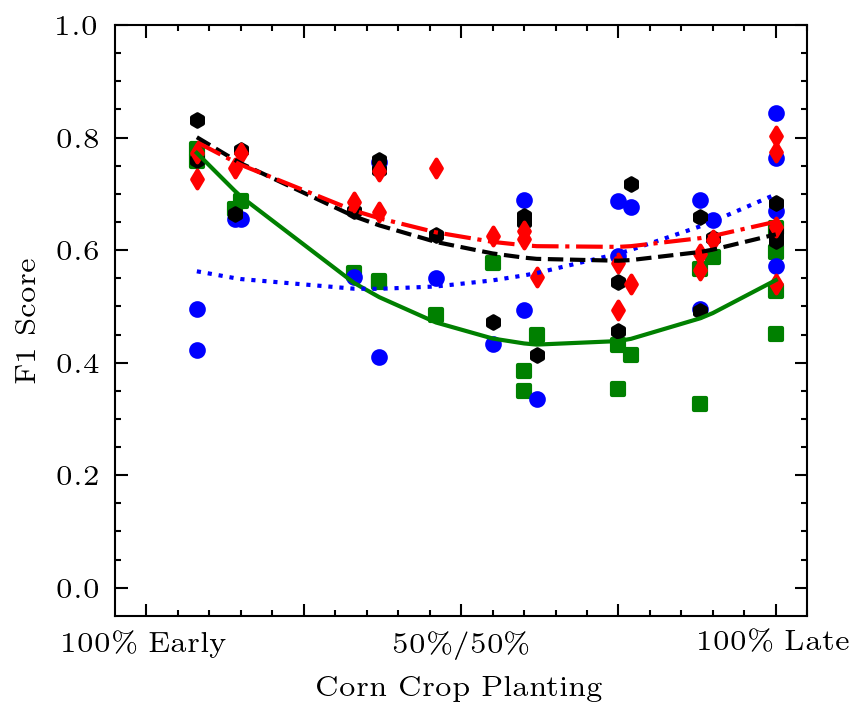}
			\caption{}
			\label{fig:F1_comparison_mature}
		\end{subfigure}%
		\begin{subfigure}{.5\textwidth}
			\centering
			\includegraphics[width=\textwidth]{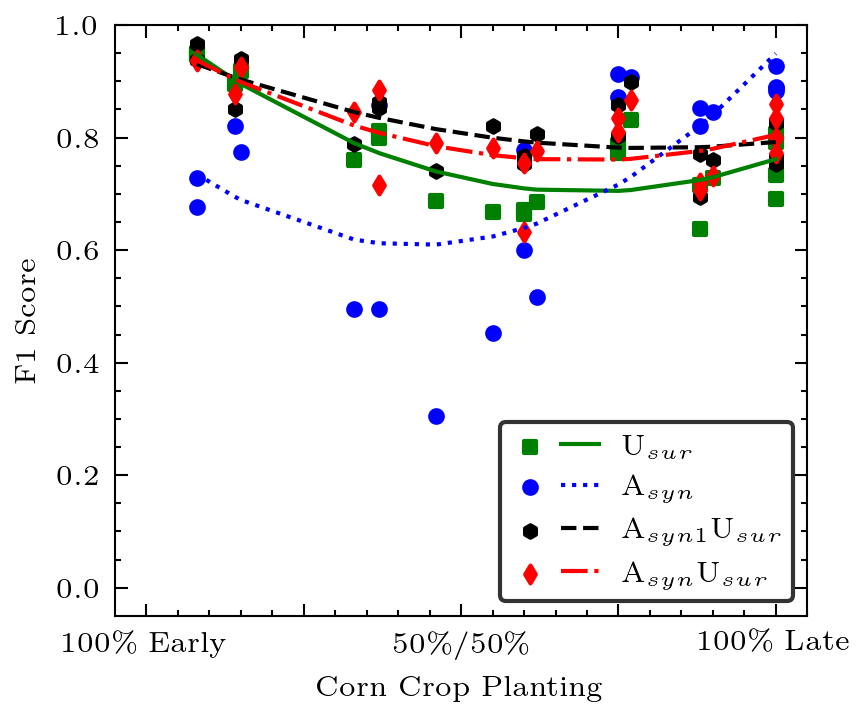}
			\caption{}
			\label{fig:F1_comparison_harvested}
		\end{subfigure}
		\caption[short]{F$_1$ score for the 2019/20 and 2020/21 test seasons in Argentina as a function of the percentage of early and late planted corn for the (a) pre-emergence, (b) emerged, (c) silking, (d) dough, (e) mature, and (f) harvested stages. Lines represent second degree polynomials fit to F$_1$ scores for each training set.}
		\label{fig:F1_vs_planting_split}
	\end{figure}

	In zones where the corn crop was evenly split between early and late planting, such as Zones IX, X, and XII, hereafter mixed zones, F$_1$ scores for U$_{sur}$ set network were significantly lower and the positive impact of including synthetic data was more evident, as shown in Tables \ref{tab:f1_2019} and \ref{tab:f1_2020}. For example, the performance gap between primarily single planting zones and mixed planting zones was significantly reduced when the network was trained on the A$_{syn1}$U$_{sur}$ set, particularly for the Silking (Figure \ref{fig:F1_comparison_silking}) and Mature (Figure \ref{fig:F1_comparison_mature}) stages. This led to an overall 7.1\% performance increase when compared to the U$_{sur}$ set (see Table \ref{tab:f1_overall}). Including synthetic data from all zones during training improved network performance further, with the largest gain for the Pre-emergence, Emerged, and Silking stages, shown in Figures \ref{fig:F1_comparison_premergence}, \ref{fig:F1_comparison_emerged}, and \ref{fig:F1_comparison_silking}.

\begin{figure}[htbp]
		\centering
		\begin{subfigure}{.5\textwidth}
			\centering
			\includegraphics[width=\textwidth]{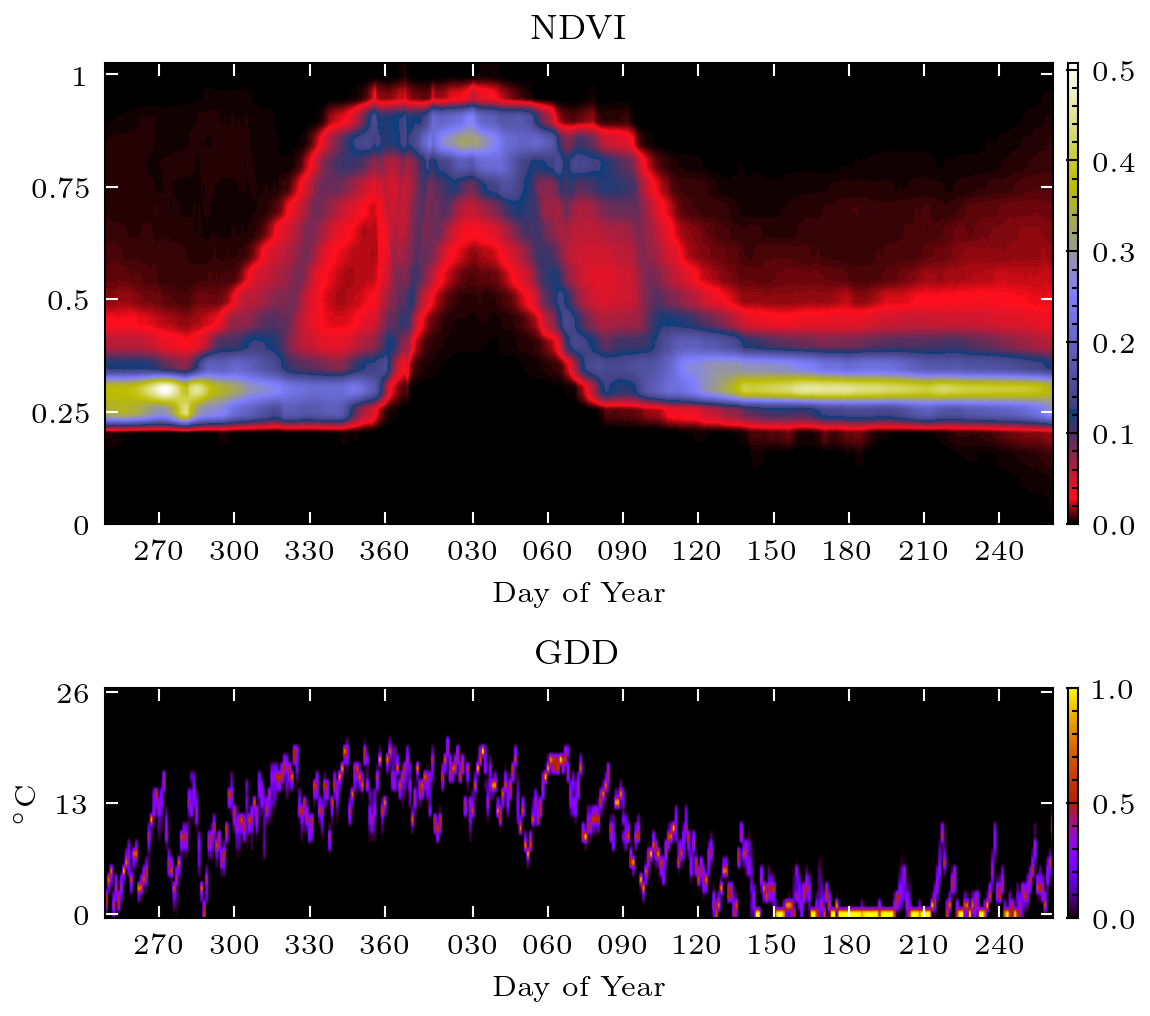}
			\caption{}
			\label{fig:input_IX_1920}
		\end{subfigure}%
		\begin{subfigure}{.5\textwidth}
			\centering
			\includegraphics[width=\textwidth]{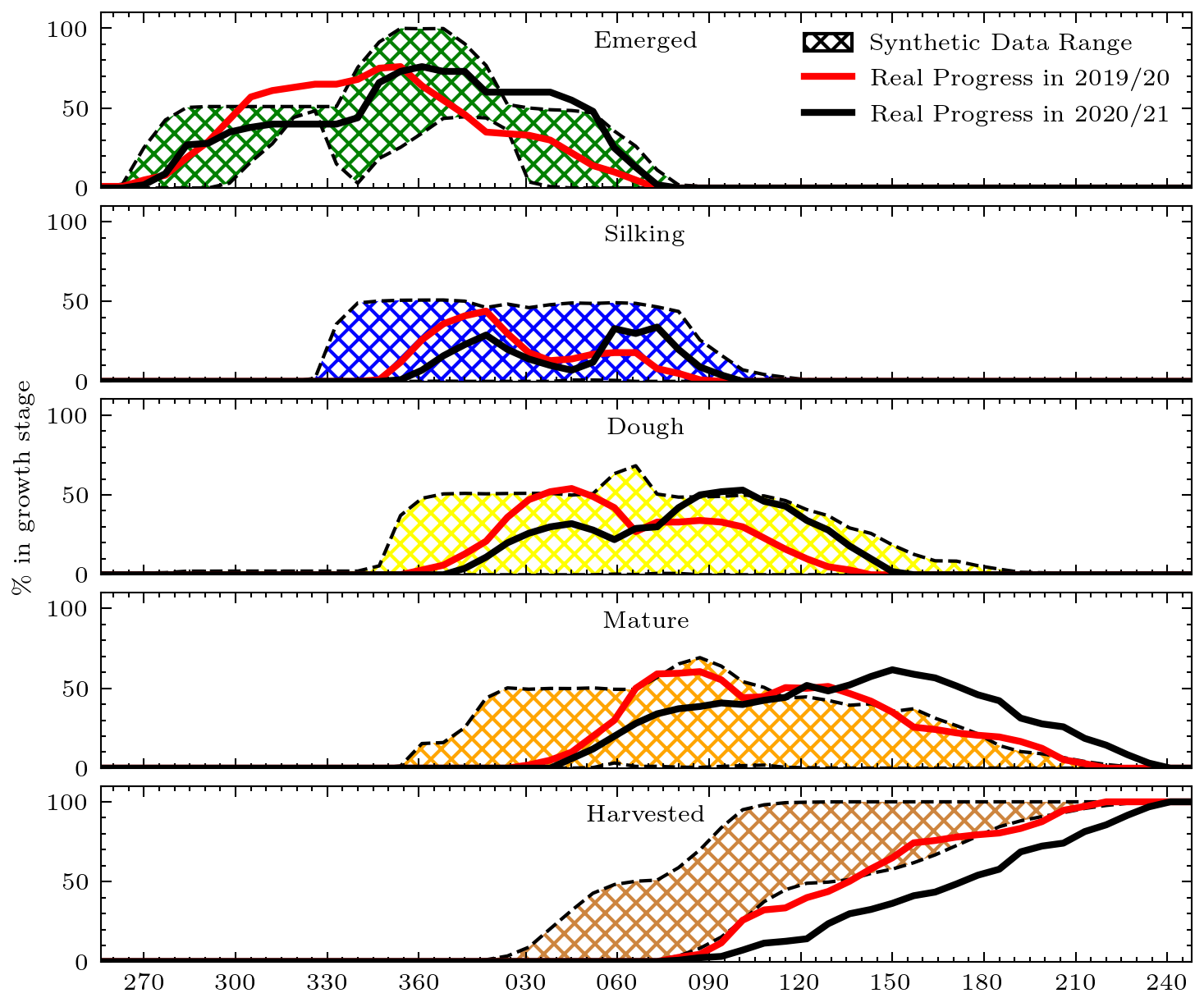}
			\caption{}
			\label{fig:result_IX_1920_comp}
		\end{subfigure}
		\begin{subfigure}{.5\textwidth}
			\centering
			\includegraphics[width=\textwidth]{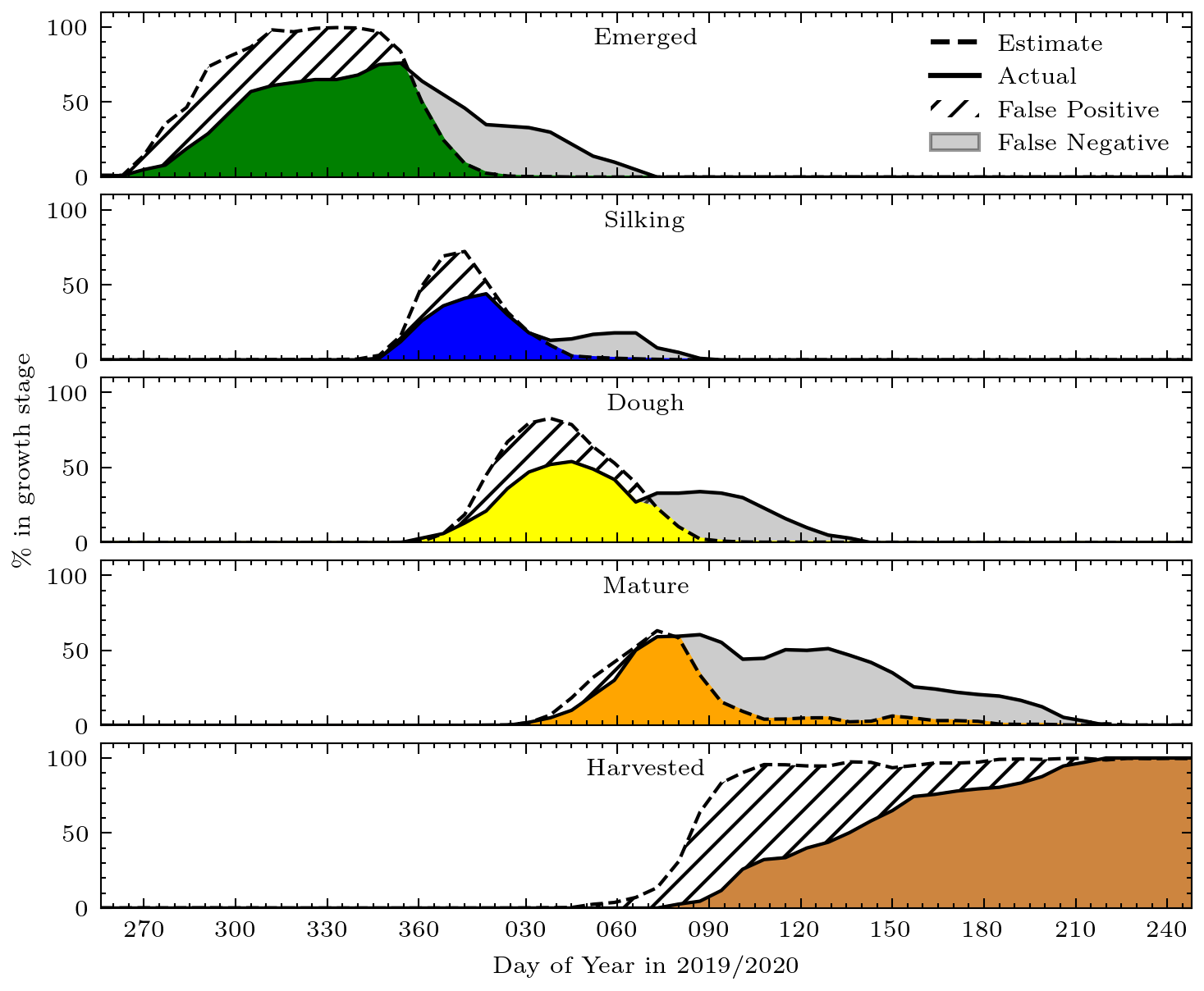}
			\caption{}
			\label{fig:result_IX_1920_surveyed}
		\end{subfigure}%
		\begin{subfigure}{.5\textwidth}
			\centering
			\includegraphics[width=\textwidth]{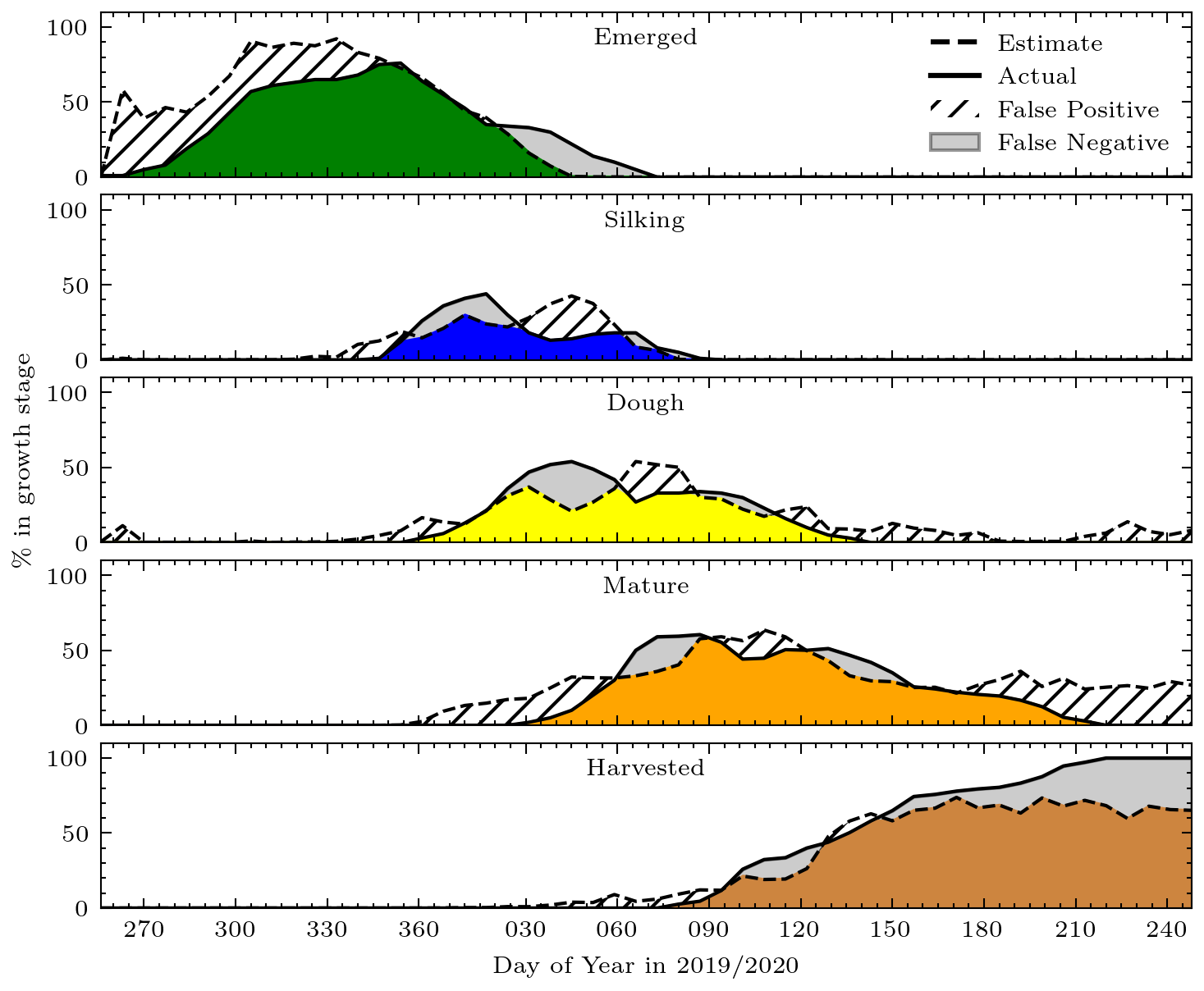}
			\caption{}
			\label{fig:result_IX_1920_synth}
		\end{subfigure}
		\begin{subfigure}{.5\textwidth}
			\centering
			\includegraphics[width=\textwidth]{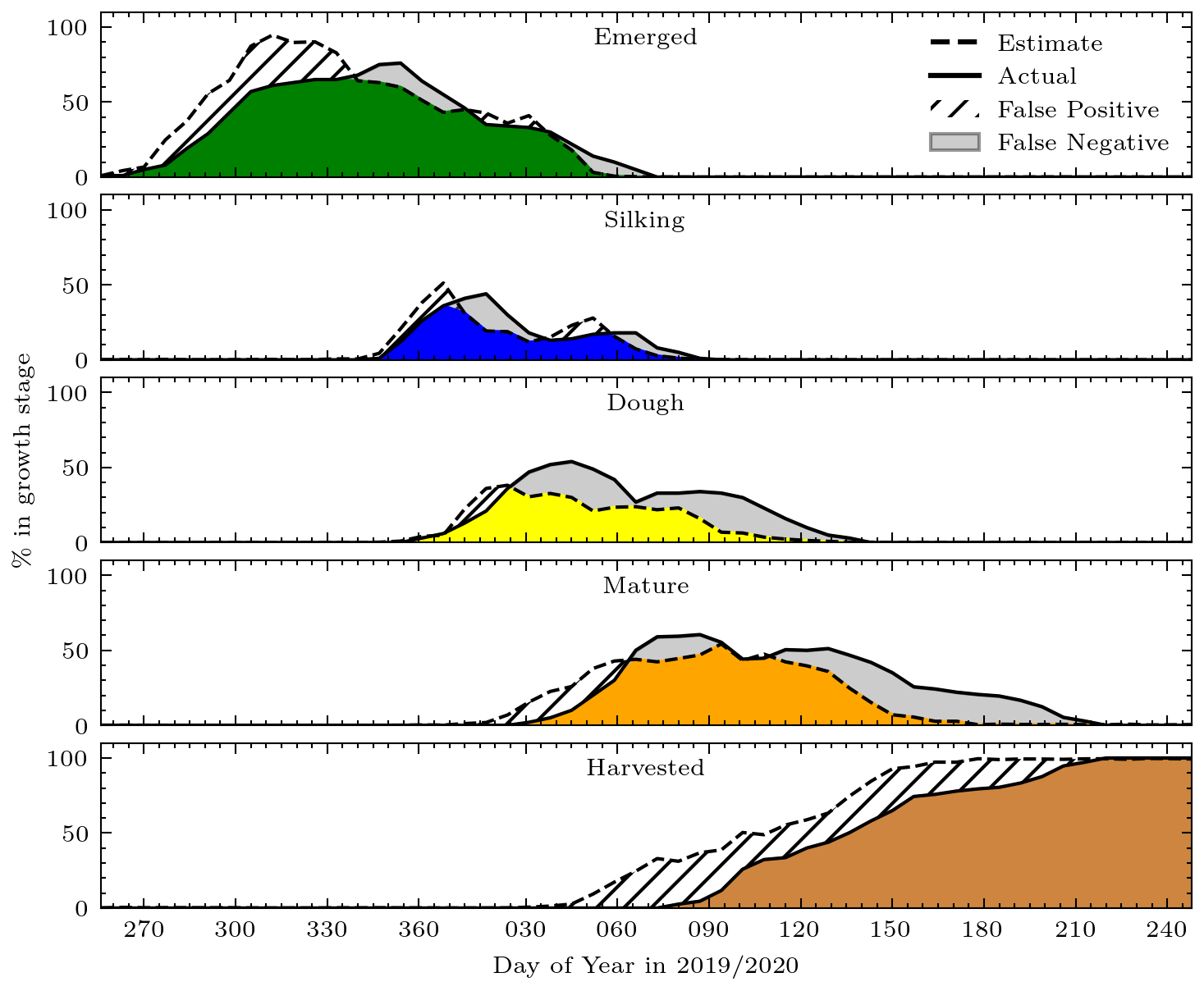}
			\caption{}
			\label{fig:result_IX_1920_MSS}
		\end{subfigure}%
		\begin{subfigure}{.5\textwidth}
			\centering
			\includegraphics[width=\textwidth]{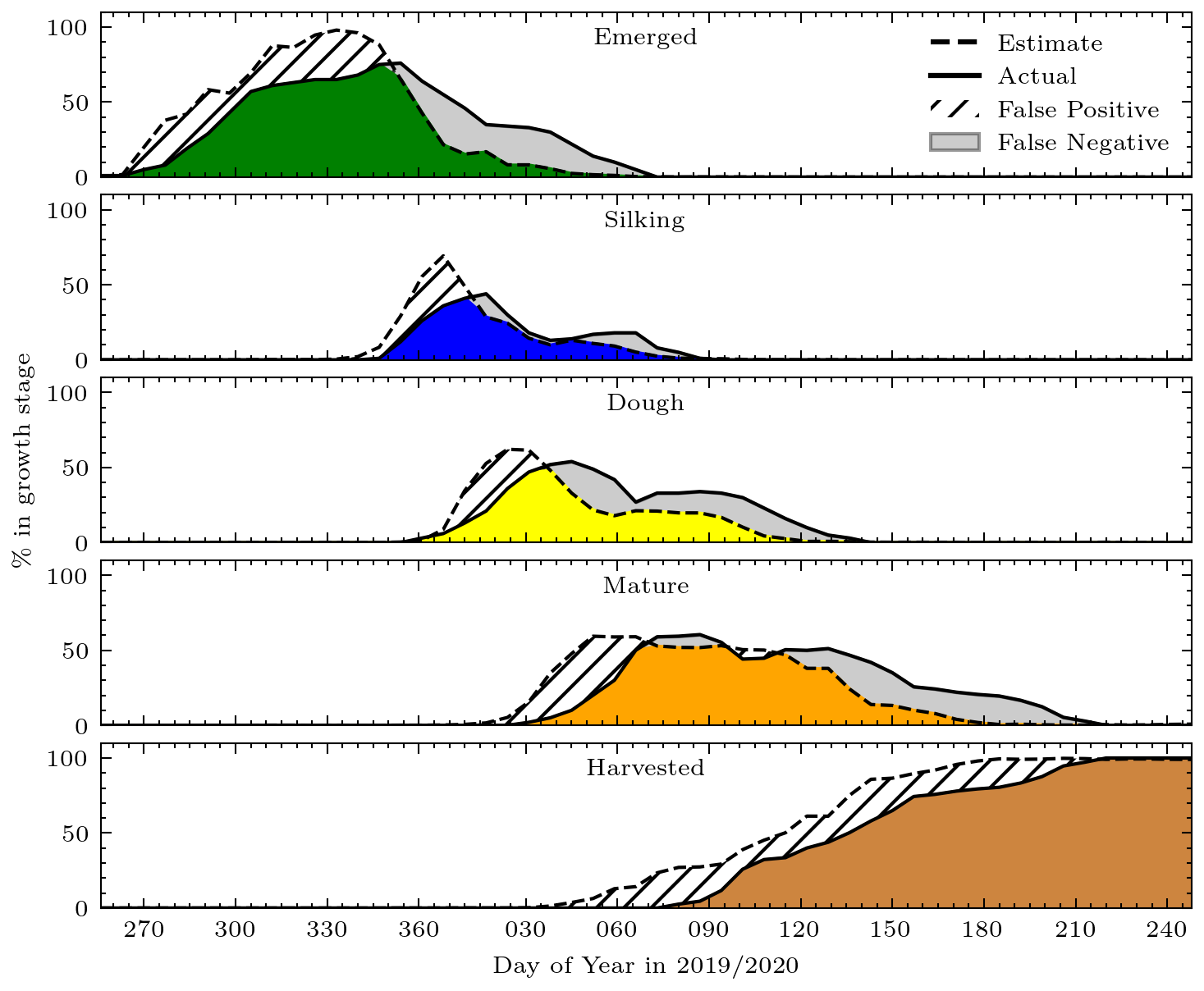}
			\caption{}
			\label{fig:result_IX_1920_joint}
		\end{subfigure}
		\caption{Zone IX during the 2019/20 season: (a) NDVI and GDD inputs, (b) test seasons and synthetic data range for crop progress, and crop progress estimates when the network was trained on (c) U$_{sur}$, (d) A$_{syn}$, (e) A$_{syn1}$U$_{sur}$, and (f) A$_{syn}$U$_{sur}$.}
		\label{fig:result_IX_1920}
	\end{figure}

	
	The benefits of exposure to synthetic examples of dual planting crop progress can be seen in Zone IX, a dual planting zone in which corn progresses bi-modally through the season. Figures \ref{fig:input_IX_1920}--\ref{fig:result_IX_1920_joint} show the NDVI and GDD histograms, synthetic and \textit{in situ} crop progress data, and network crop progress estimates for Zone IX during the 2019/20 test season, respectively. When trained on the U$_{sur}$ data, the network produced uni-modal estimates of crop progress, shown in Figure \ref{fig:result_IX_1920_surveyed}. Without exposure to synthetic examples of dual planting crop progress during training, the network only identified the first green-up curve present in the NDVI histogram, shown in Figure \ref{fig:input_IX_1920}, and estimated near 100\% emergence around DOY 300. This error continued through the season, with crop progress estimates following the early planting crop but missing the late planting crop. This caused high false positive and false negative rates on both the Silking and Dough stages. In contrast, when exposed to synthetic data in the A$_{syn}$ (Figure \ref{fig:result_IX_1920_synth}), A$_{syn1}$U$_{sur}$ (Figure \ref{fig:result_IX_1920_MSS}), and A$_{syn}$U$_{sur}$ (Figure \ref{fig:result_IX_1920_joint}) sets during training, the network could of identify and estimate the bi-modal crop progress in this zone, with higher true positive rates for stages Emerged through Mature and a lower false negative rate on the Harvested stage,  as reflected in the F$_1$ scores in Table \ref{tab:f1_2019}. Agreement between the range of crop progress within the synthetic data for Zone IX and \textit{in situ} crop progress for the two test seasons was high, as shown in Figure \ref{fig:result_IX_1920_comp}, leading to high F$_1$ scores for this zone when the network was trained on the A$_{syn}$ and A$_{syn1}$U$_{sur}$ datasets. 
	
\begin{figure}[htbp]
		\centering
		\begin{subfigure}{.5\textwidth}
			\centering
			\includegraphics[width=\textwidth]{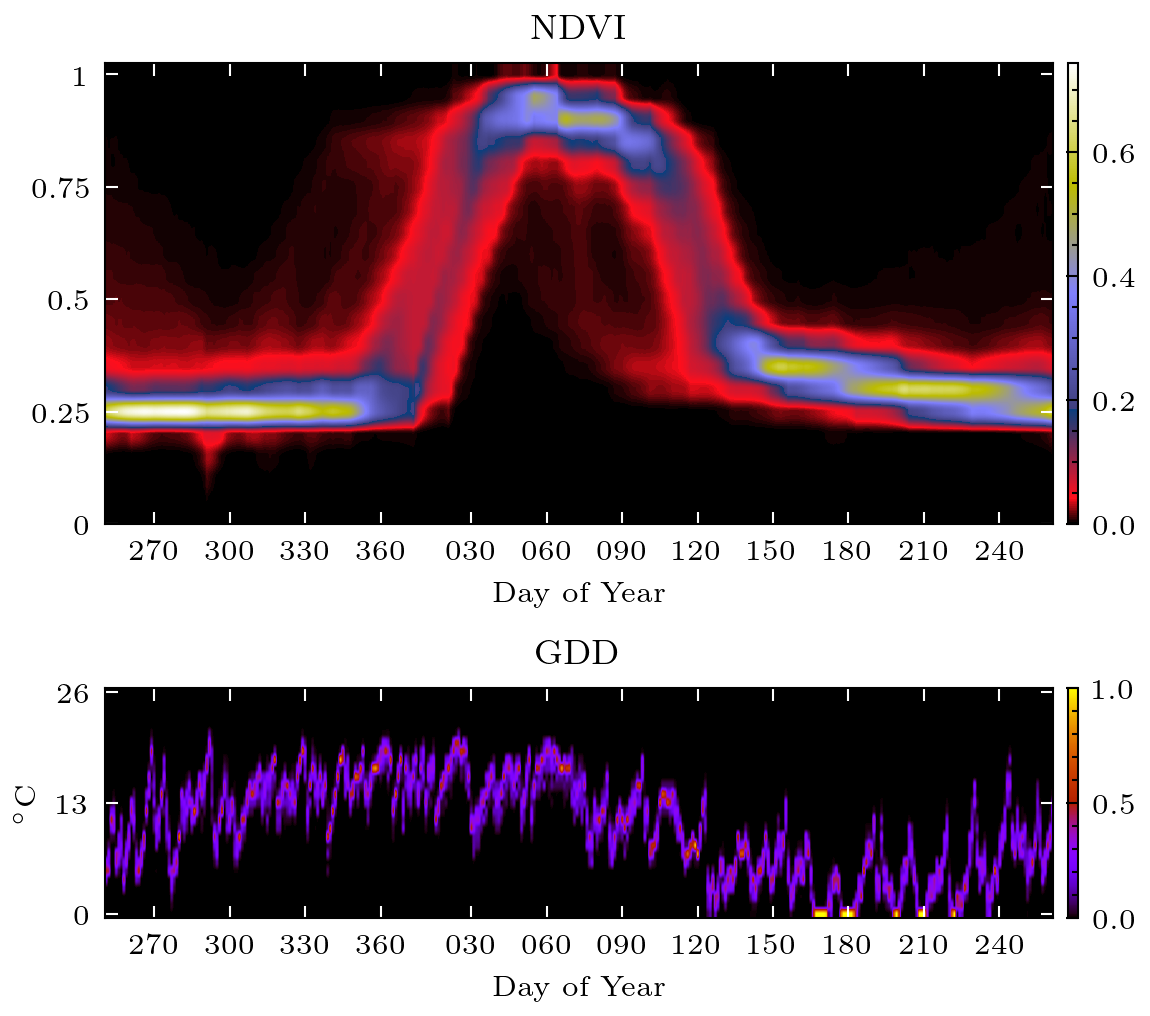}
			\caption{}
			\label{fig:input_III_2021}
		\end{subfigure}%
		\begin{subfigure}{.5\textwidth}
			\centering
			\includegraphics[width=\textwidth]{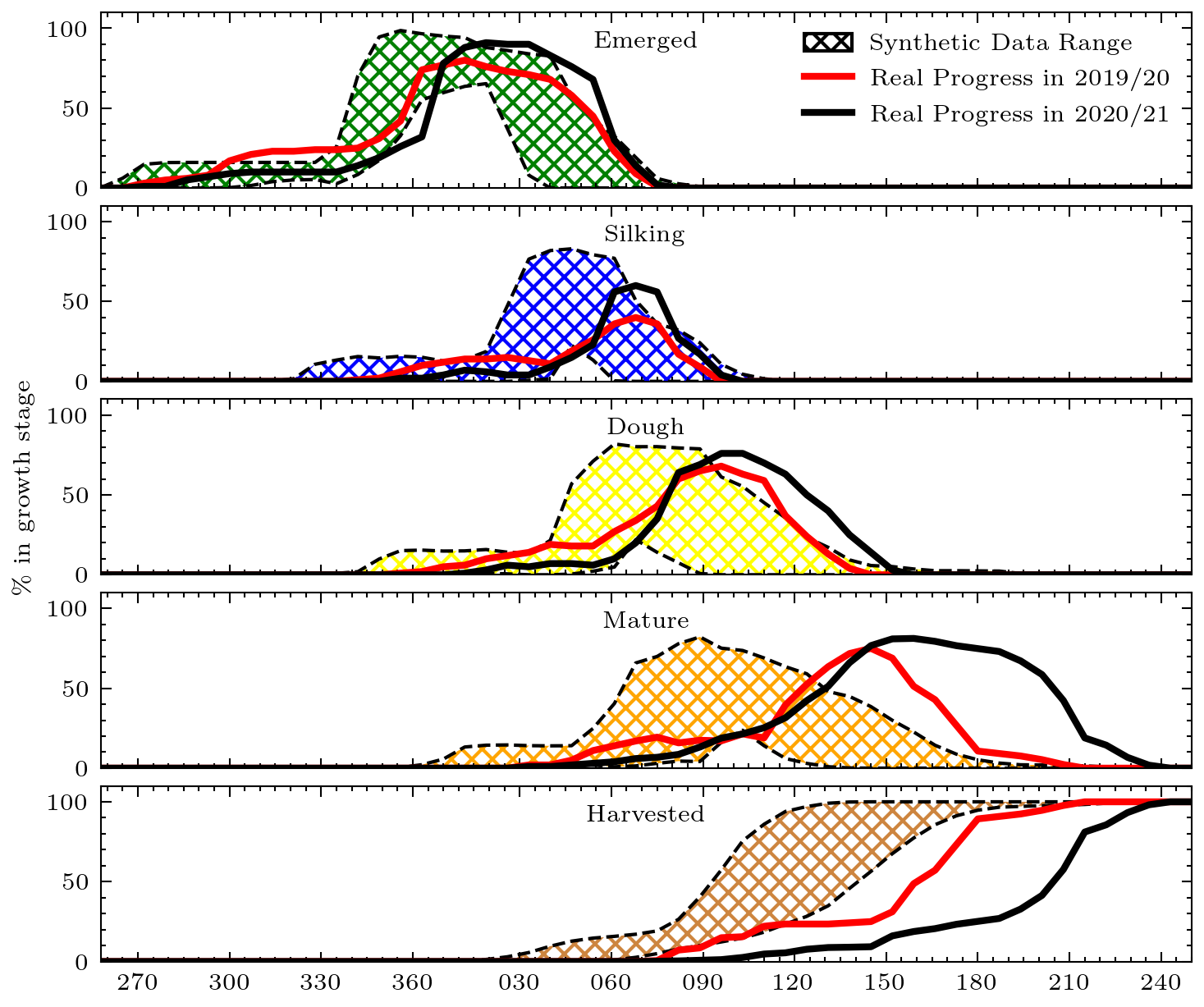}
			\caption{}
			\label{fig:result_III_2021_comp}
		\end{subfigure}
		\begin{subfigure}{.5\textwidth}
			\centering
			\includegraphics[width=\textwidth]{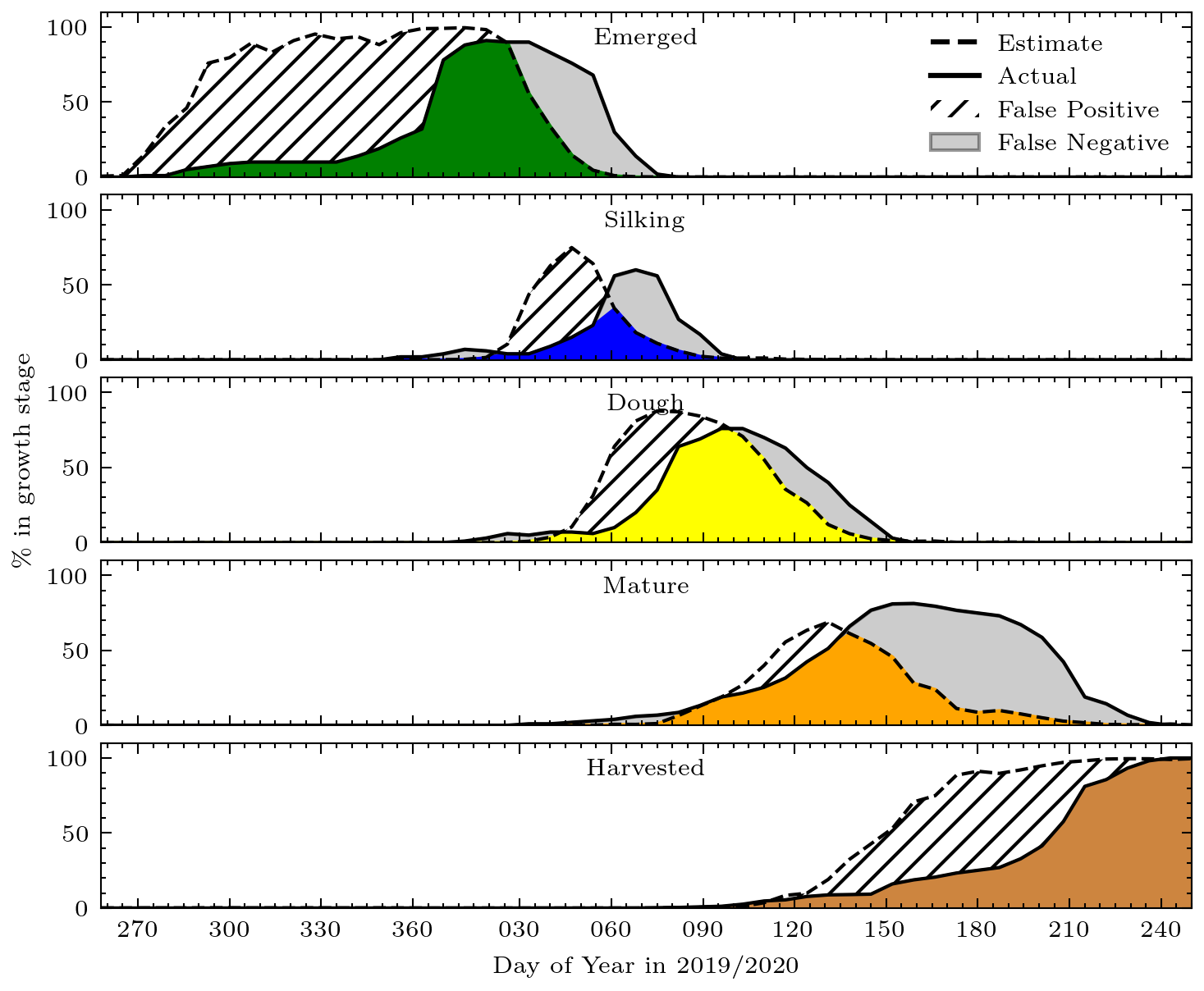}
			\caption{}
			\label{fig:result_III_2021_surveyed}
		\end{subfigure}%
		\begin{subfigure}{.5\textwidth}
			\centering
			\includegraphics[width=\textwidth]{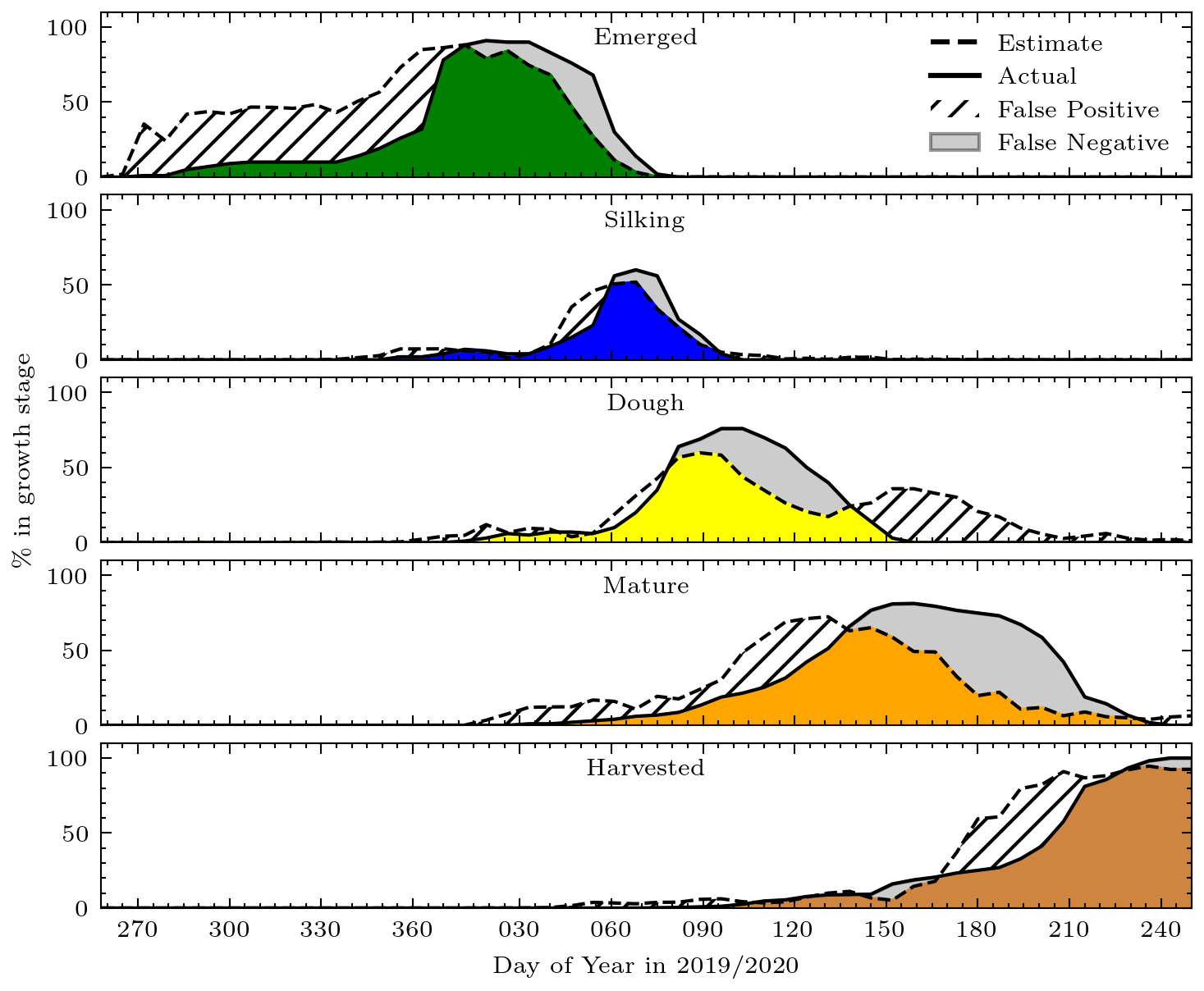}
			\caption{}
			\label{fig:result_III_2021_synth}
		\end{subfigure}
		\begin{subfigure}{.5\textwidth}
			\centering
			\includegraphics[width=\textwidth]{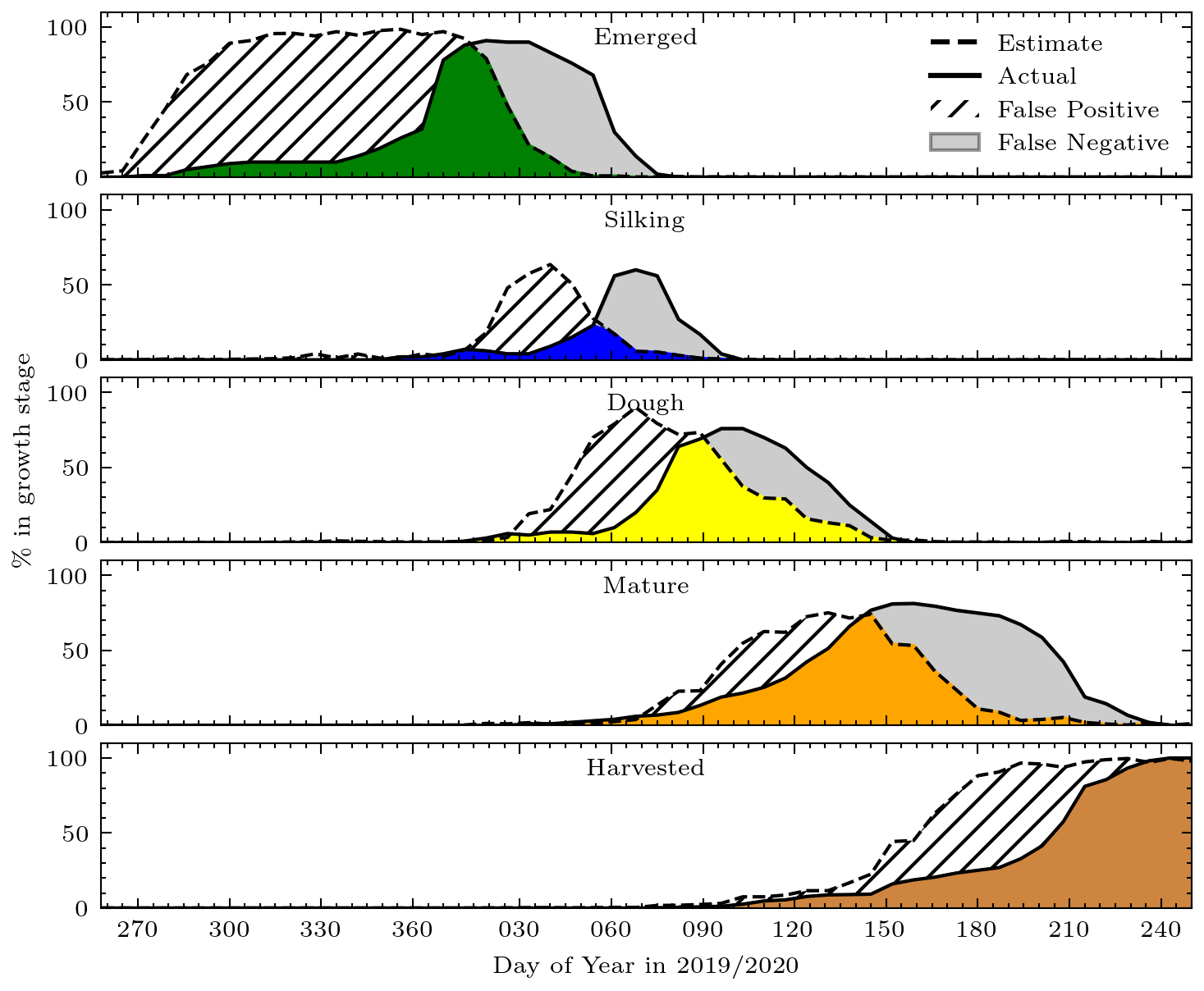}
			\caption{}
			\label{fig:result_III_2021_MSS}
		\end{subfigure}%
		\begin{subfigure}{.5\textwidth}
			\centering
			\includegraphics[width=\textwidth]{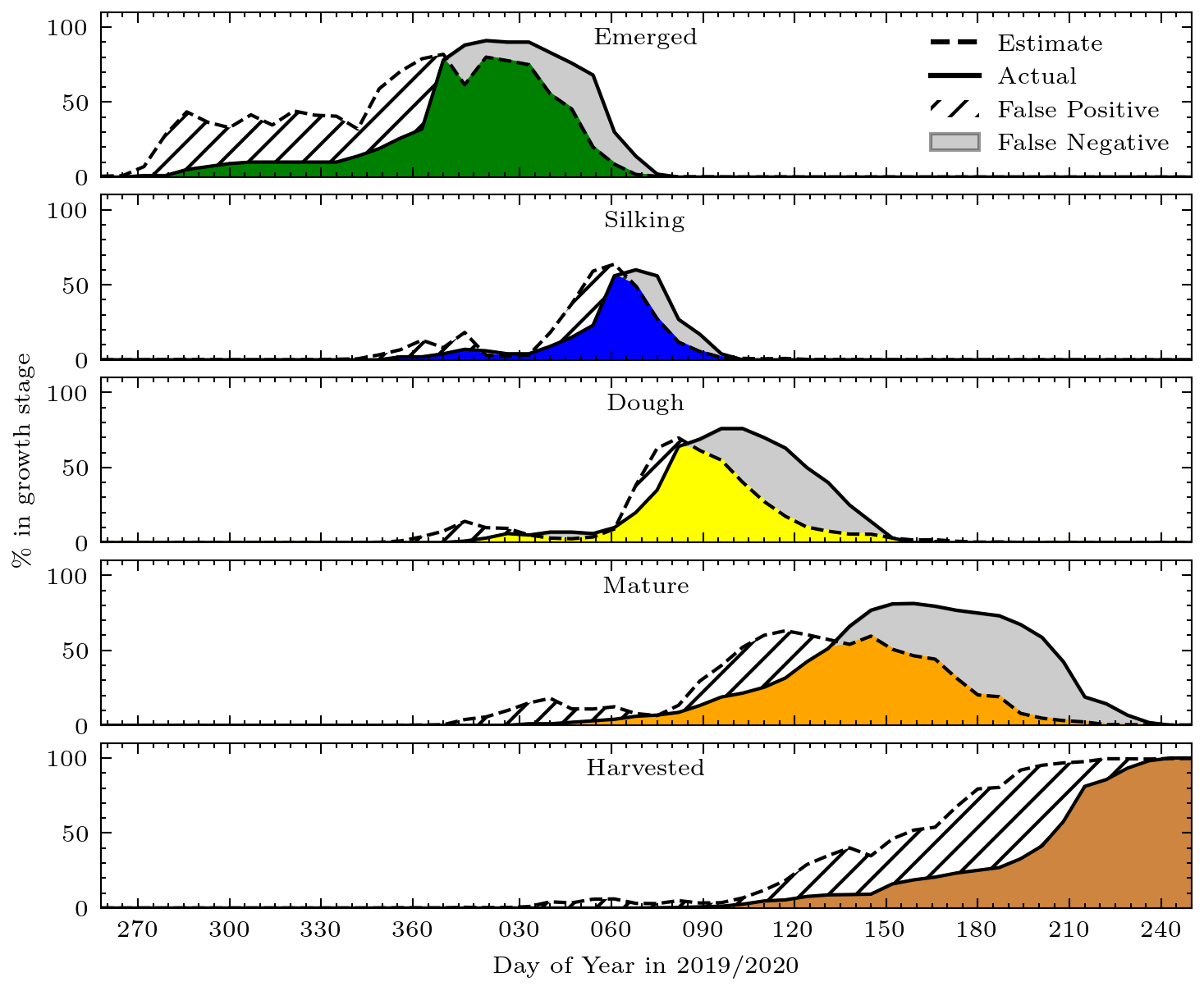}
			\caption{}
			\label{fig:result_III_2021_joint}
		\end{subfigure}
		\caption{Zone III during the 2020/21 season: (a) NDVI and GDD inputs, (b) test seasons and synthetic data range for crop progress, and crop progress estimates when the network was trained on (c) U$_{sur}$, (d) A$_{syn}$, (e) A$_{syn1}$U$_{sur}$, and (f) A$_{syn}$U$_{sur}$.}
		\label{fig:result_III_2021}
	\end{figure}

	While the addition of synthetic data improved overall F$_1$ scores in mixed planting zones, performance was significantly reduced when the network was exposed to synthetic data that were poorly aligned with \textit{in situ} progress. For example, in Zone III during the 2020/21 season, shown in Figure \ref{fig:result_III_2021}, \textit{in situ} crop progress was slower than crop progress in the synthetic data, implying the required GDD to maturity for cultivars grown in Zone III was higher than for the cultivar used in IXIM simulations. When the network was trained on the U$_{sur}$ and A$_{syn1}$U$_{sur}$ sets, network estimates had minimal overlap with \textit{in situ} crop progress for the Silking stage, as shown in Figures \ref{fig:result_III_2021_surveyed} and \ref{fig:result_III_2021_MSS}, respectively. This poor alignment with \textit{in situ} Silking resulted in lower F$_1$ scores for the Silking stage when trained on the A$_{syn1}$U$_{sur}$ set than when trained on the U$_{sur}$ set alone (see Table \ref{tab:f1_2020}).
 	Training on the A$_{syn}$ and A$_{syn}$U$_{sur}$ datasets, however, which contain a broader range of synthetic crop progress examples, mitigated this issue, reducing the network's early false positive rate, shown in Figures \ref{fig:result_III_2021_synth} and \ref{fig:result_III_2021_joint}. This result suggests that using a variety of synthetic crop progress examples from multiple regions during training is preferable when cultivar information from the target region is not known \textit{a priori}.
 	
\begin{figure}[htbp]
	\centering
	\begin{subfigure}{.5\textwidth}
		\centering
		\includegraphics[width=\textwidth]{figs/inputs/Argentina_V_2020_inputs.png}
		\caption{}
		\label{fig:input_V_2021}
	\end{subfigure}%
	\begin{subfigure}{.5\textwidth}
		\centering
		\includegraphics[width=\textwidth]{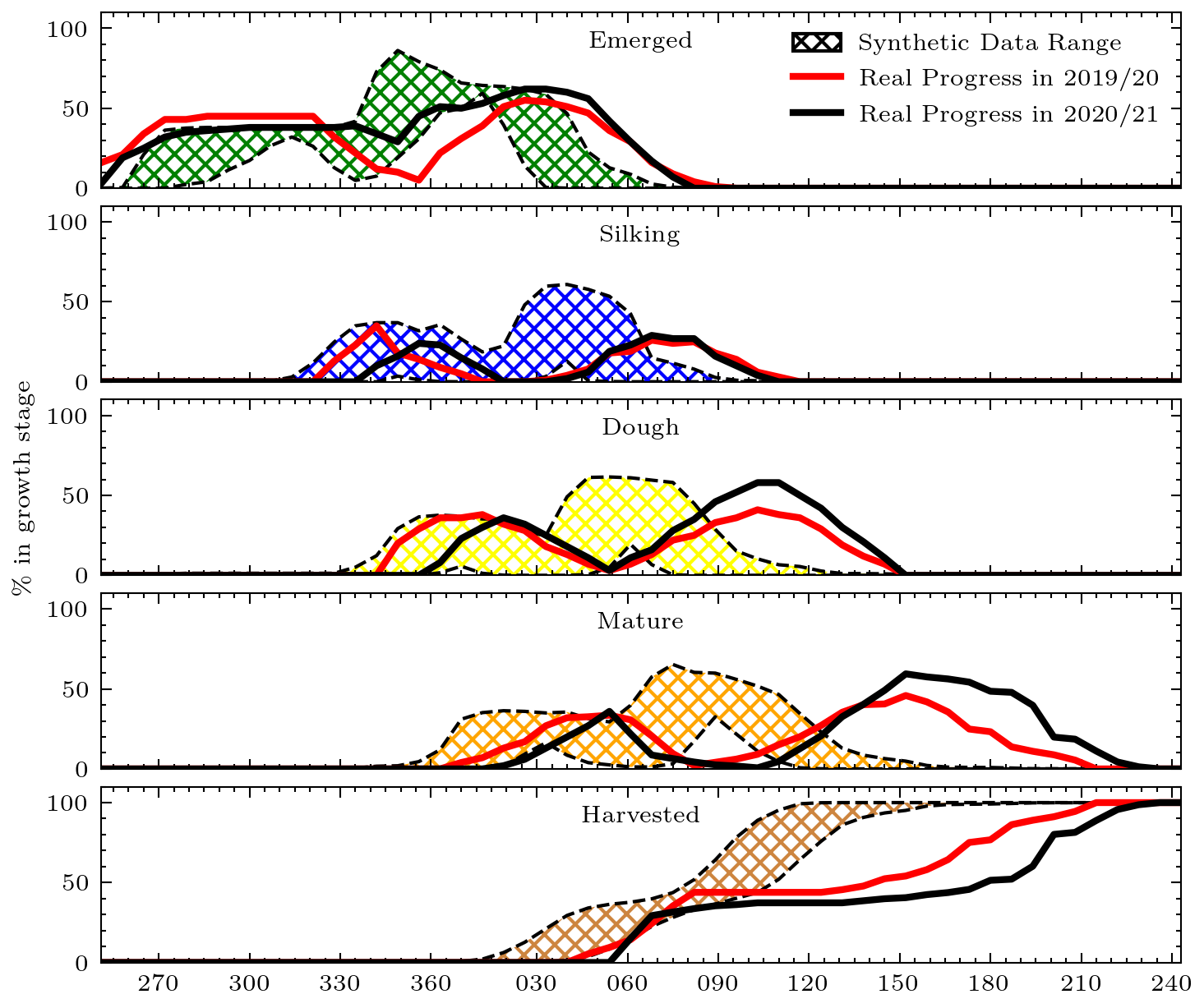}
		\caption{}
		\label{fig:result_V_2021_comp}
	\end{subfigure}
	\begin{subfigure}{.5\textwidth}
		\centering
		\includegraphics[width=\textwidth]{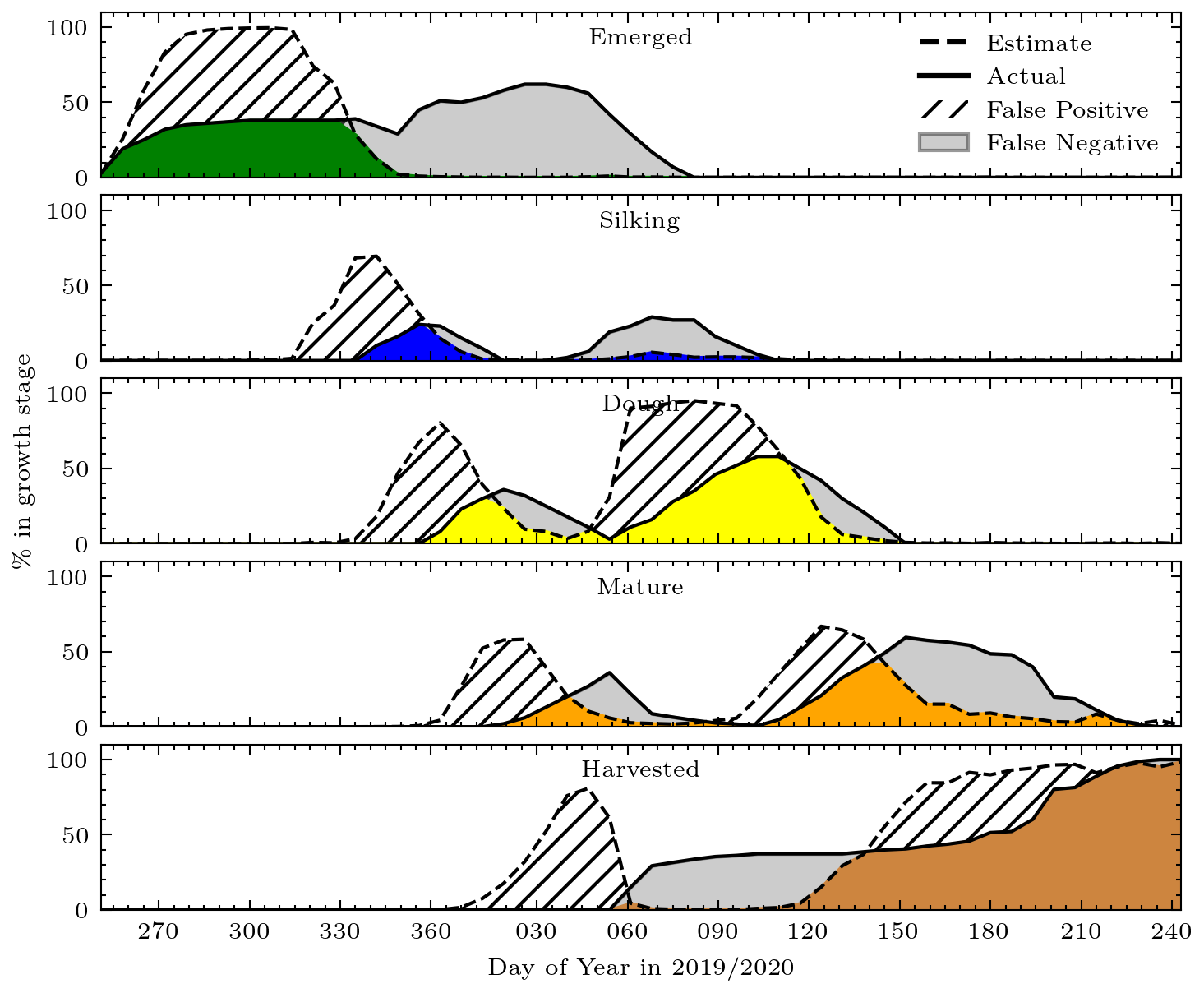}
		\caption{}
		\label{fig:result_V_2021_surveyed}
	\end{subfigure}%
	\begin{subfigure}{.5\textwidth}
		\centering
		\includegraphics[width=\textwidth]{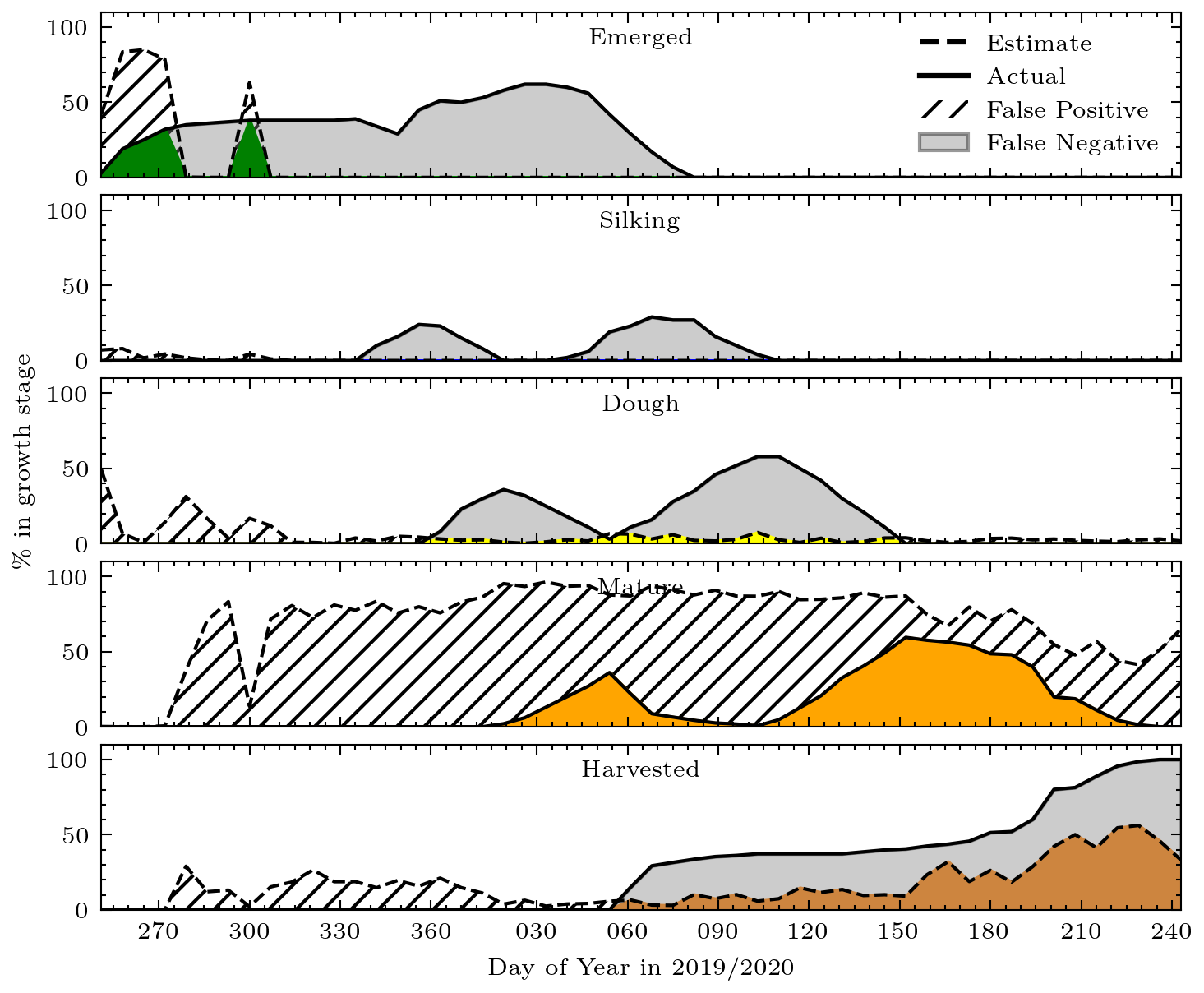}
		\caption{}
		\label{fig:result_V_2021_synth}
	\end{subfigure}
	\begin{subfigure}{.5\textwidth}
		\centering
		\includegraphics[width=\textwidth]{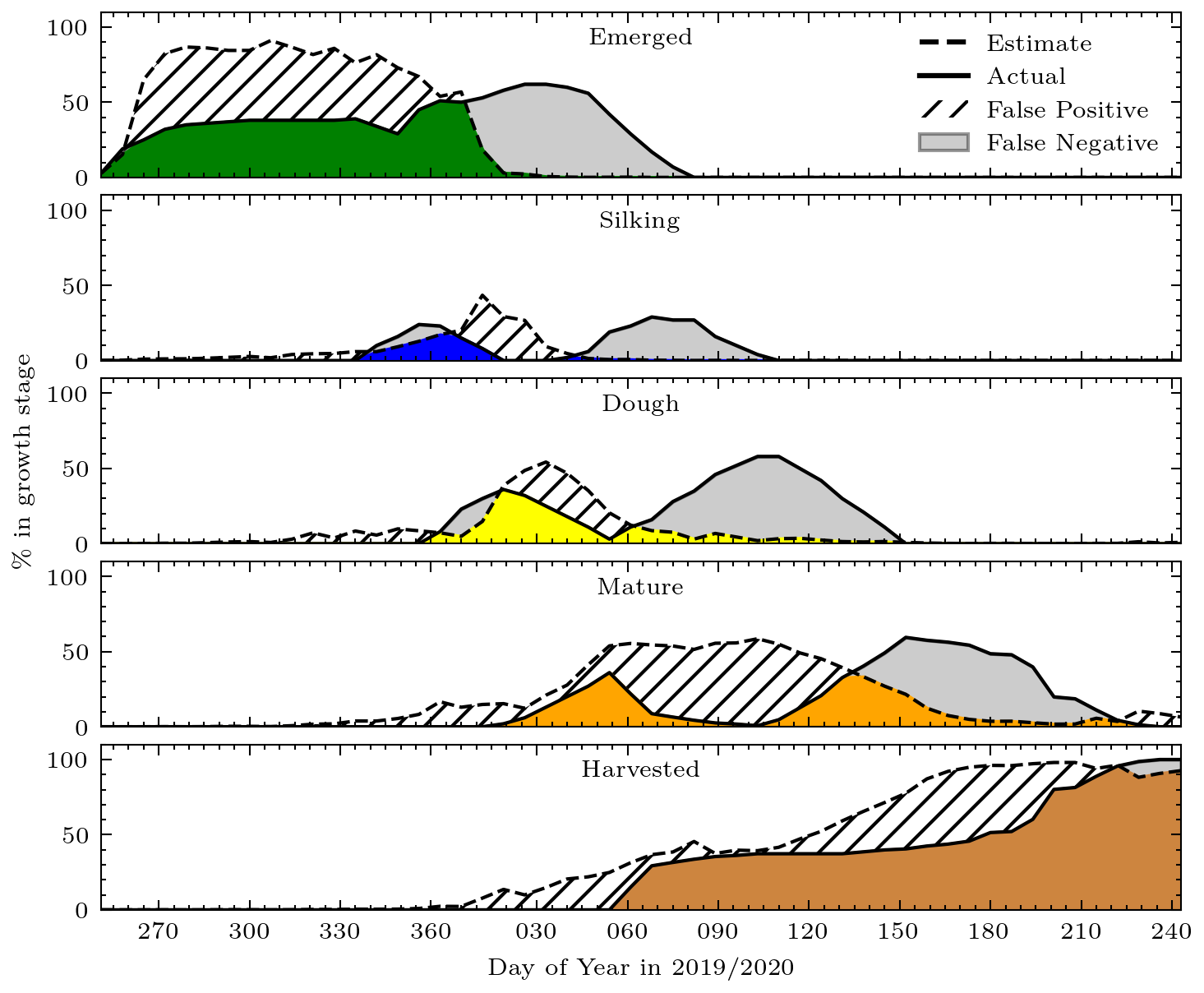}
		\caption{}
		\label{fig:result_V_2021_MSS}
	\end{subfigure}%
	\begin{subfigure}{.5\textwidth}
		\centering
		\includegraphics[width=\textwidth]{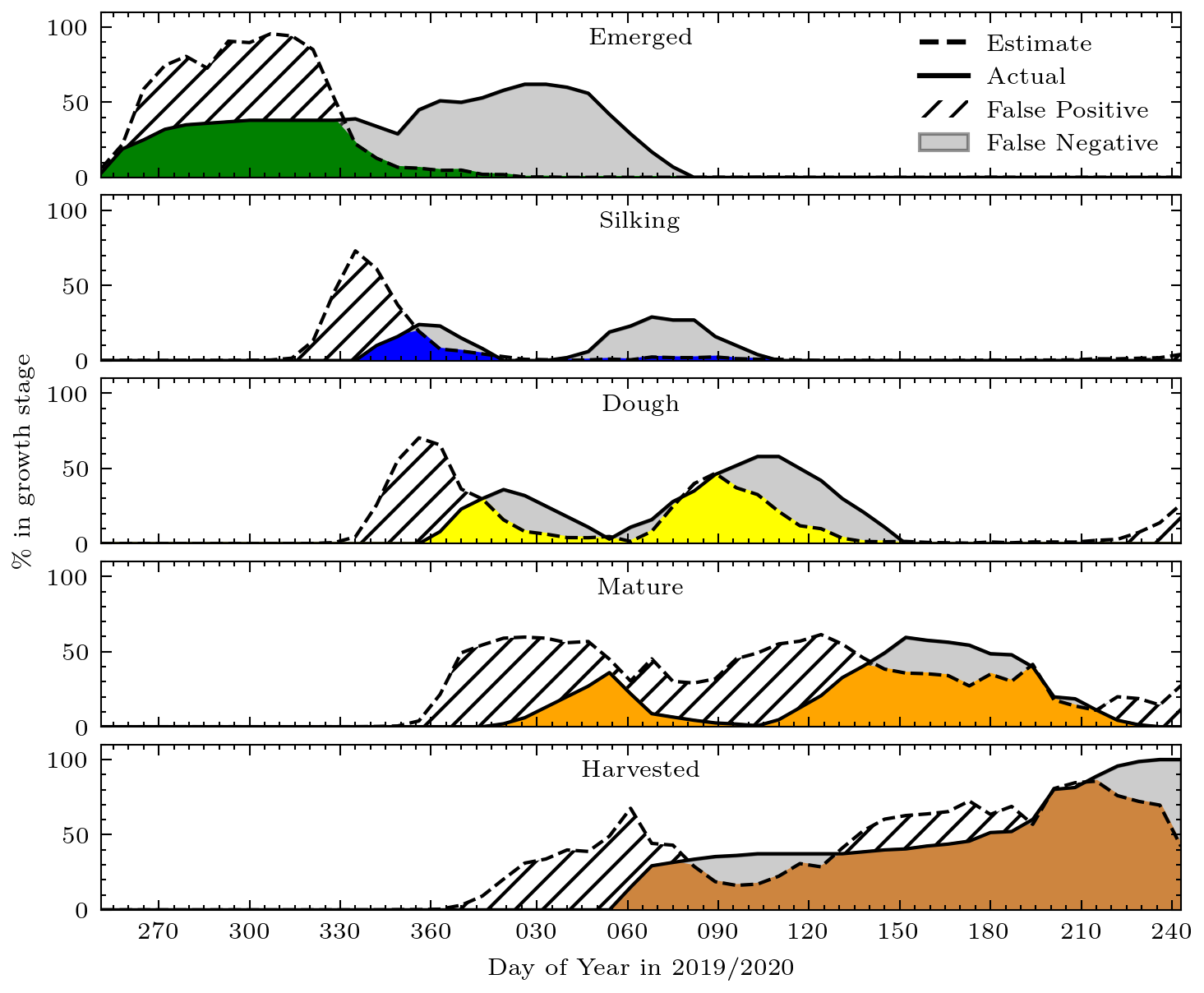}
		\caption{}
		\label{fig:result_V_2021_joint}
	\end{subfigure}
	\caption{Zone V during the 2020/21 season: (a) NDVI and GDD inputs, (b) test seasons and synthetic data range for crop progress, and crop progress estimates when the network was trained on (c) U$_{sur}$, (d) A$_{syn}$, (e) A$_{syn1}$U$_{sur}$, and (f) A$_{syn}$U$_{sur}$.}
	\label{fig:result_V_2021}
\end{figure}

 	Combining real and synthetic data, i.e. A$_{syn1}$U$_{sur}$, A$_{syn}$U$_{sur}$, increased network performance in most dual planting zones, but F$_1$ scores remained very low for Zone V, shown in Figure \ref{fig:result_V_2021}, where the early and late planting windows are the most separated (see Figure \ref{fig:arg_planting_windows}). 	Network performance was low for Zone V because alignment between the range of synthetic crop progress data and \textit{in situ} crop progress, shown in Figure \ref{fig:result_V_2021_comp}, was particularly poor for this zone. While this issue was mitigated when training on the A$_{syn}$ and A$_{syn}$U$_{sur}$ datasets for other zones, Zone V planting windows are so separate that synthetic data from other zones was not sufficient to provide examples which encompass \textit{in situ} crop progress for this zone.
 	
 	Another factor contributing to low performance could be the significant pre-planting and post-harvest noise in the NDVI data from off-season crops, shown in Figure \ref{fig:input_V_2021}. Noise in the NDVI histogram caused crop progress estimates to revert from Harvested to earlier stages as the late planting NDVI curve peaked between DOY 60 and 90 (Figure \ref{fig:result_V_2021_surveyed}) and during post-season NDVI noise after DOY 210 (Figure \ref{fig:result_V_2021_joint}).

	\subsection{Mitigation of NDVI Noise Sensitivity}	
\begin{figure}[htbp]
		\centering
		\begin{subfigure}{.5\textwidth}
			\centering
			\includegraphics[width=\textwidth]{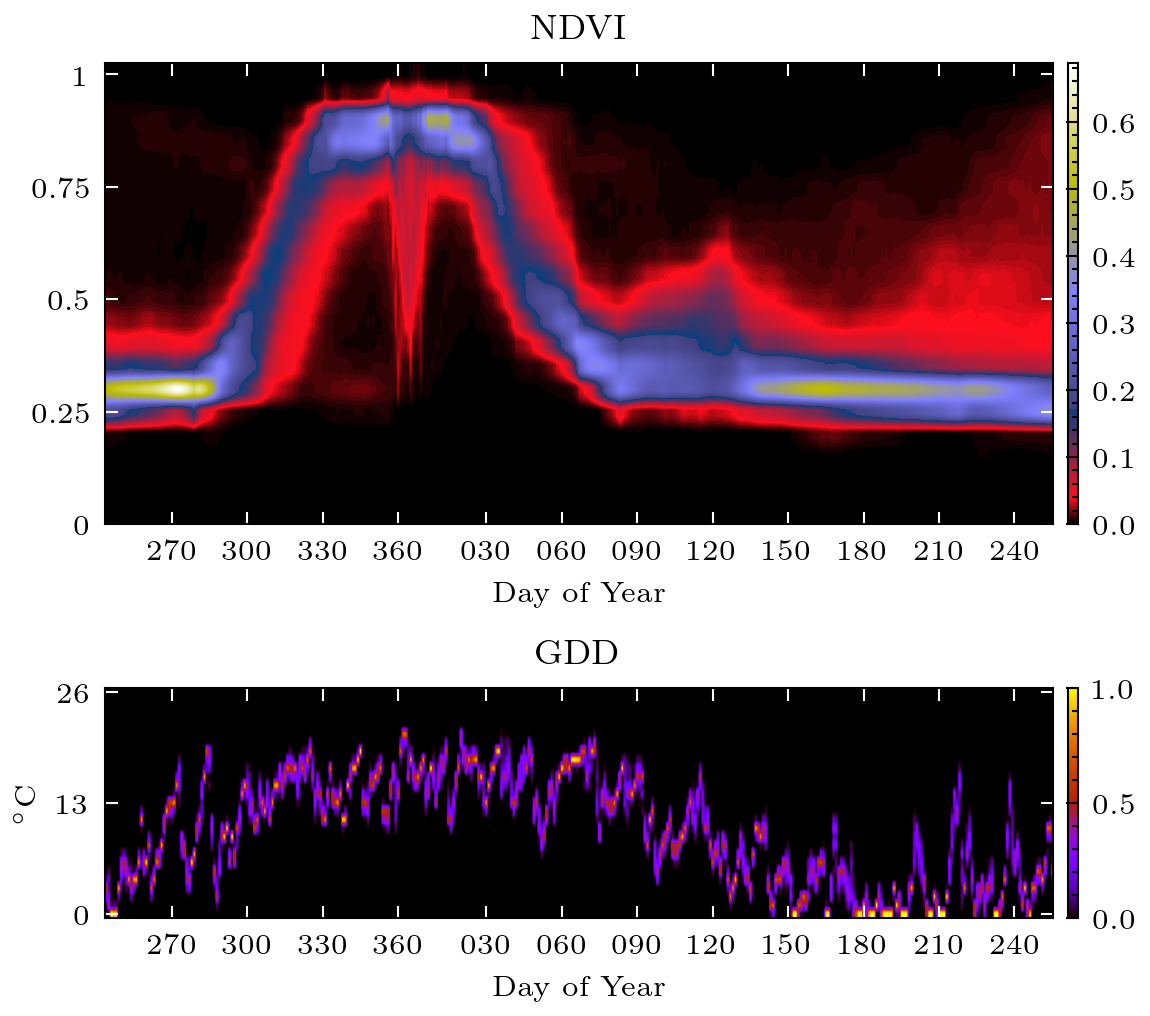}
			\caption{}
			\label{fig:input_VI_1920}
		\end{subfigure}%
		\begin{subfigure}{.5\textwidth}
			\centering
			\includegraphics[width=\textwidth]{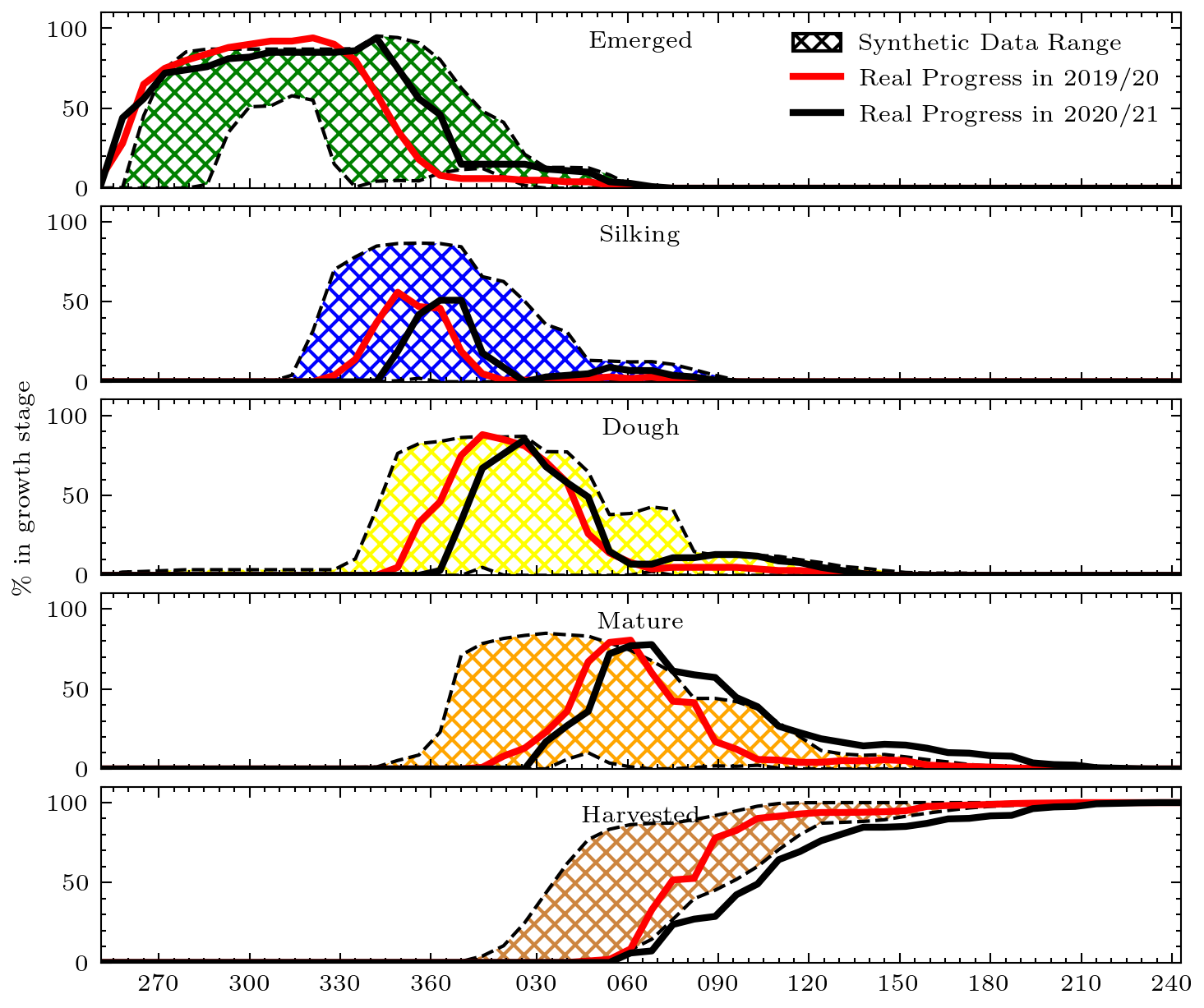}
			\caption{}
			\label{fig:result_VI_1920_comp}
		\end{subfigure}
		\begin{subfigure}{.5\textwidth}
			\centering
			\includegraphics[width=\textwidth]{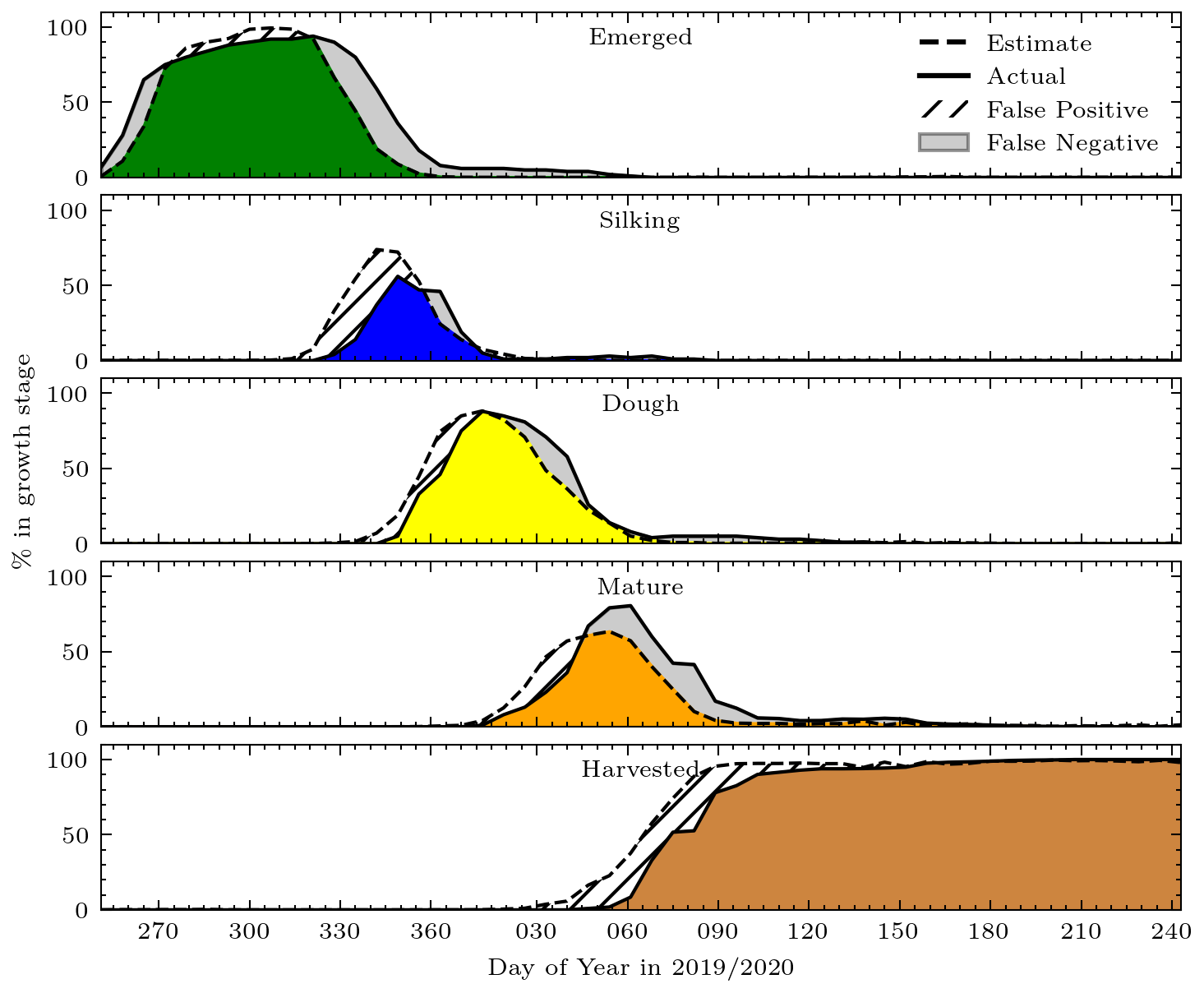}
			\caption{}
			\label{fig:result_VI_1920_surveyed}
		\end{subfigure}%
		\begin{subfigure}{.5\textwidth}
			\centering
			\includegraphics[width=\textwidth]{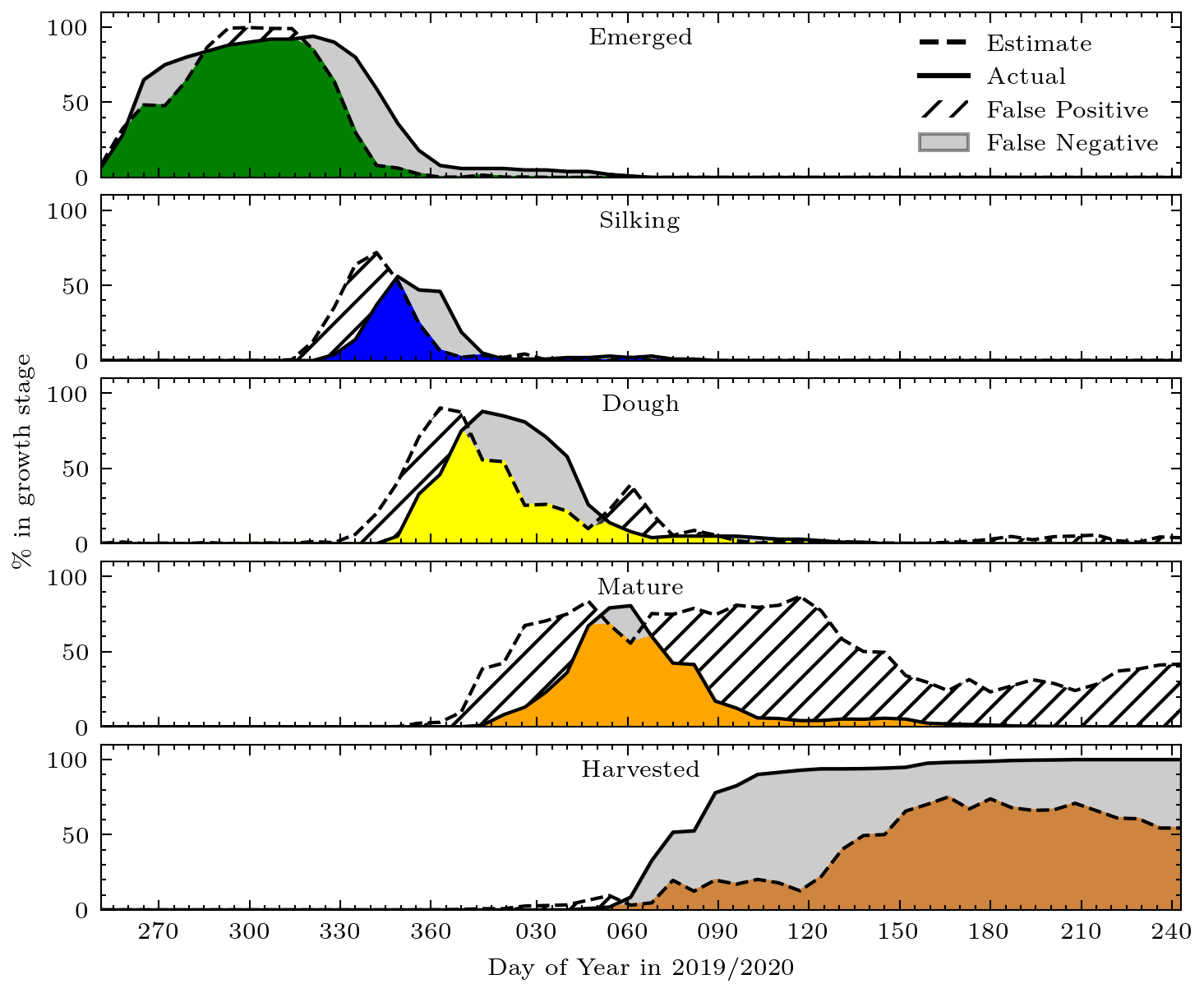}
			\caption{}
			\label{fig:result_VI_1920_synth}
		\end{subfigure}
		\begin{subfigure}{.5\textwidth}
			\centering
			\includegraphics[width=\textwidth]{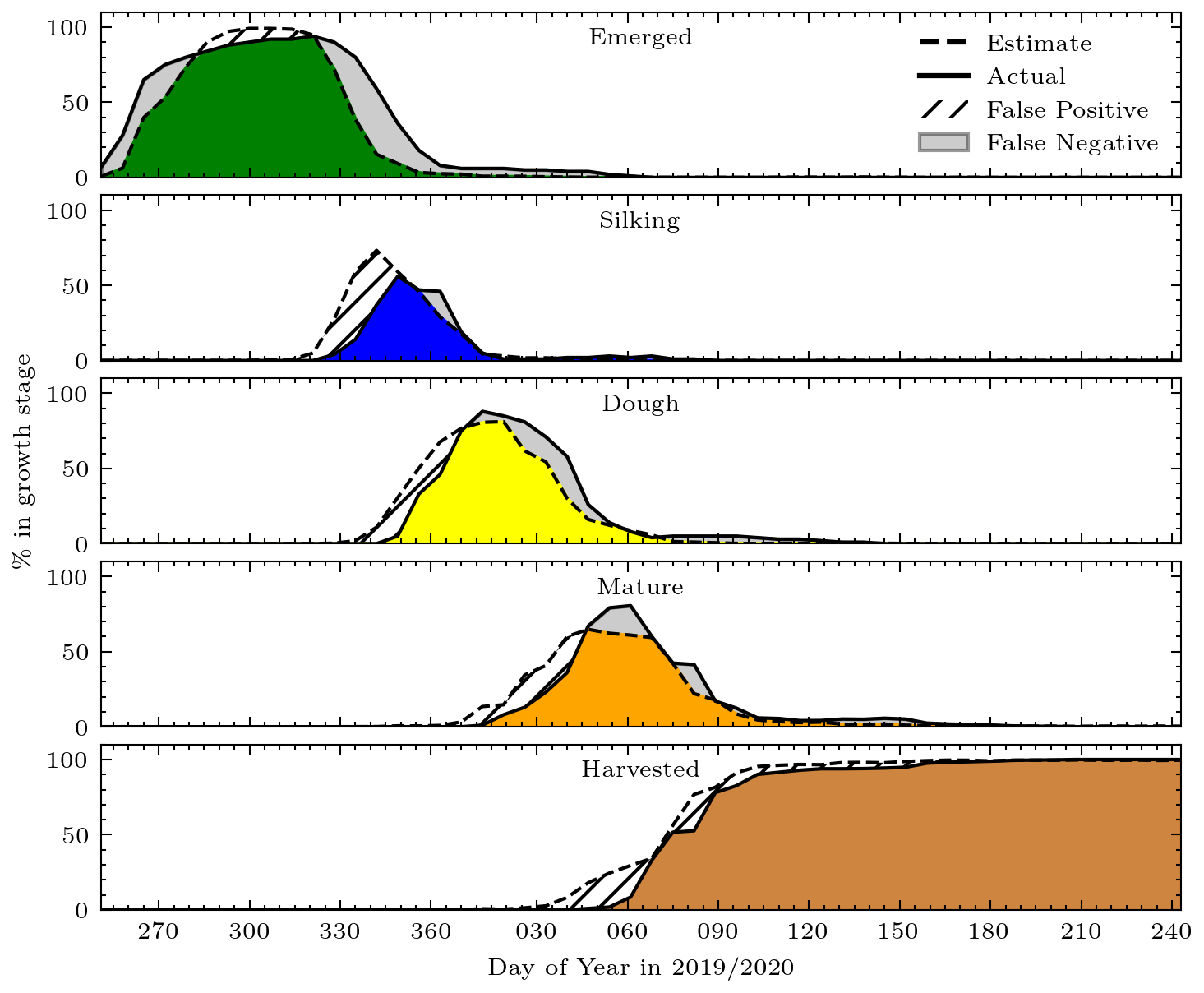}
			\caption{}
			\label{fig:result_VI_1920_MSS}
		\end{subfigure}%
		\begin{subfigure}{.5\textwidth}
			\centering
			\includegraphics[width=\textwidth]{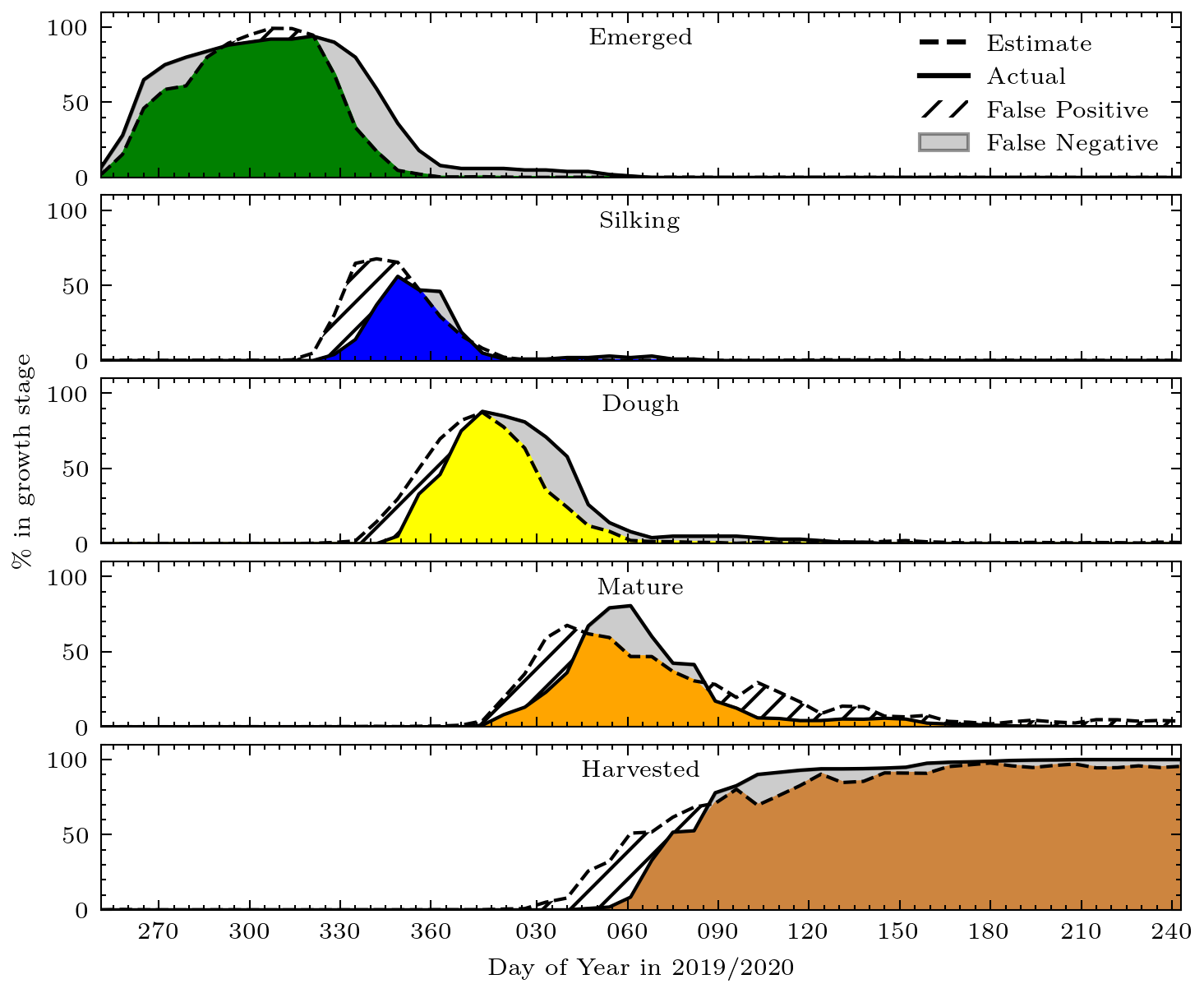}
			\caption{}
			\label{fig:result_VI_1920_joint}
		\end{subfigure}
		\caption{Zone VI during the 2019/20 season: (a) NDVI and GDD inputs, (b) test seasons and synthetic data range for crop progress, and crop progress estimates when the network was trained on (c) U$_{sur}$, (d) A$_{syn}$, (e) A$_{syn1}$U$_{sur}$, and (f) A$_{syn}$U$_{sur}$.}
		\label{fig:result_VI_1920}
	\end{figure}

\begin{figure}[htb]
		\centering
		\begin{subfigure}[b]{\textwidth}
			\centering
			\includegraphics[width=0.475\linewidth]{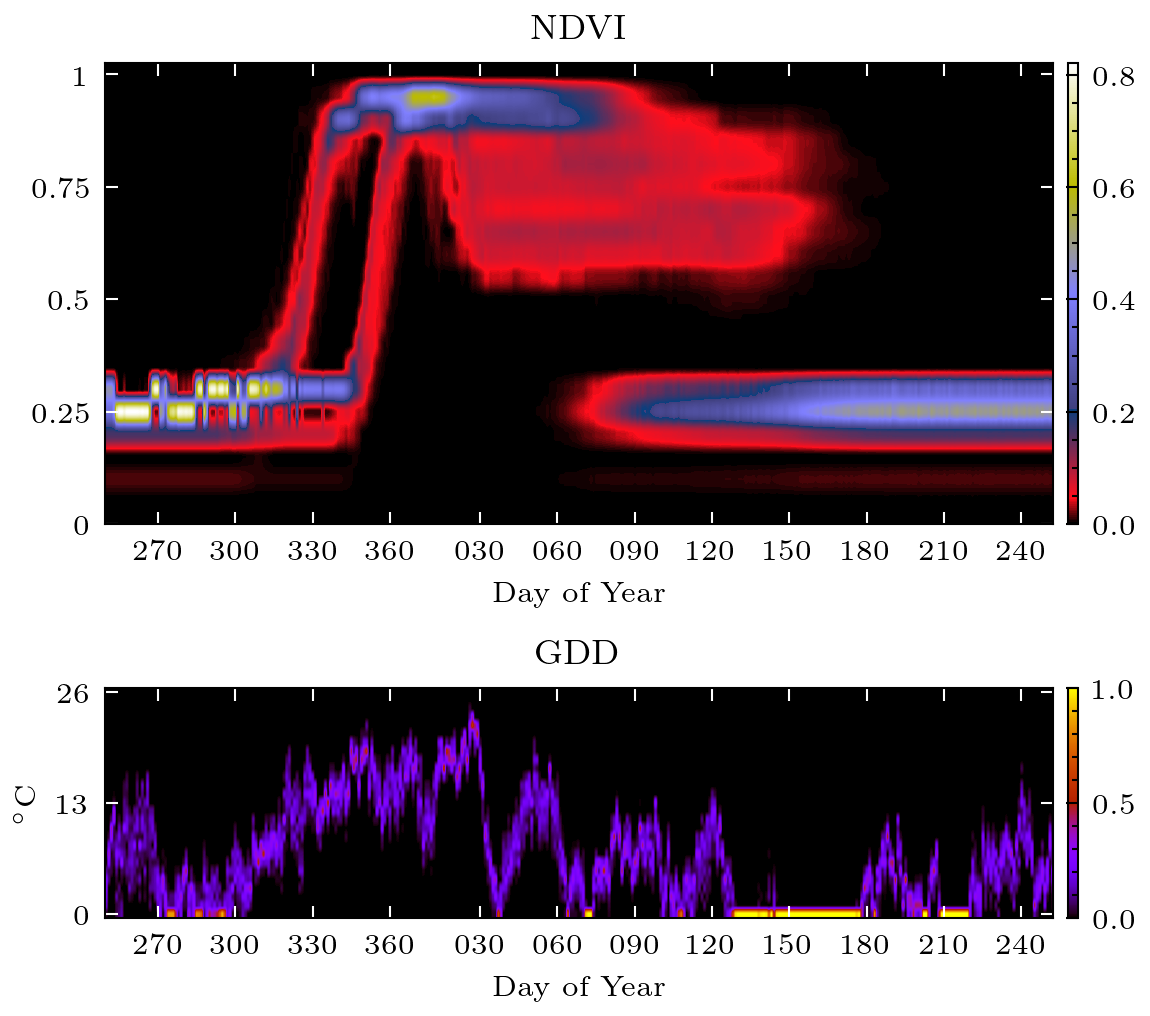}%
			\hfill
			\includegraphics[width=0.475\linewidth]{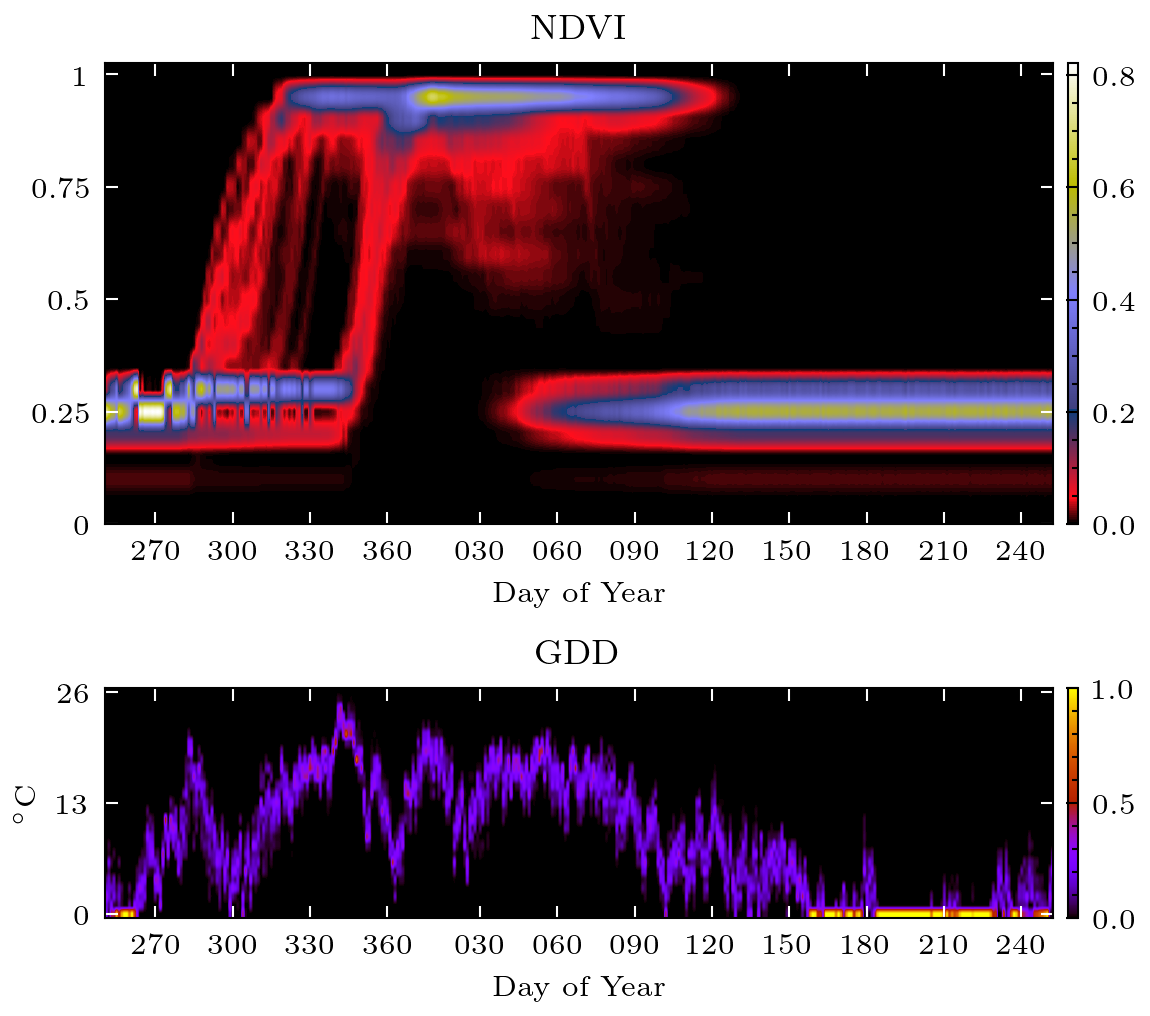}
			\caption{}
		\end{subfigure}
		\vskip\baselineskip
		\begin{subfigure}[b]{\textwidth}
			\centering
			\includegraphics[width=0.475\linewidth]{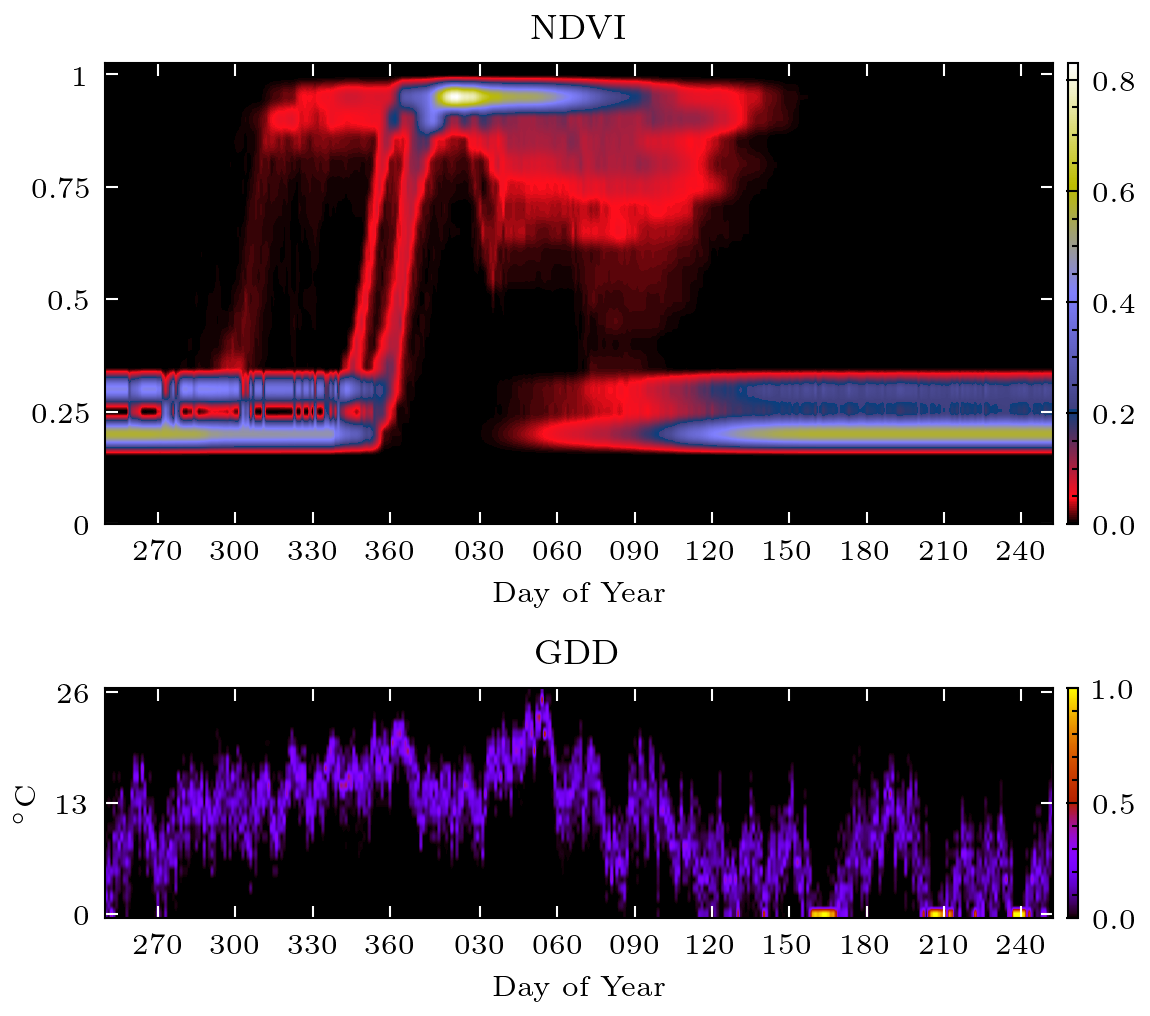}%
			\hfill
			\includegraphics[width=0.475\linewidth]{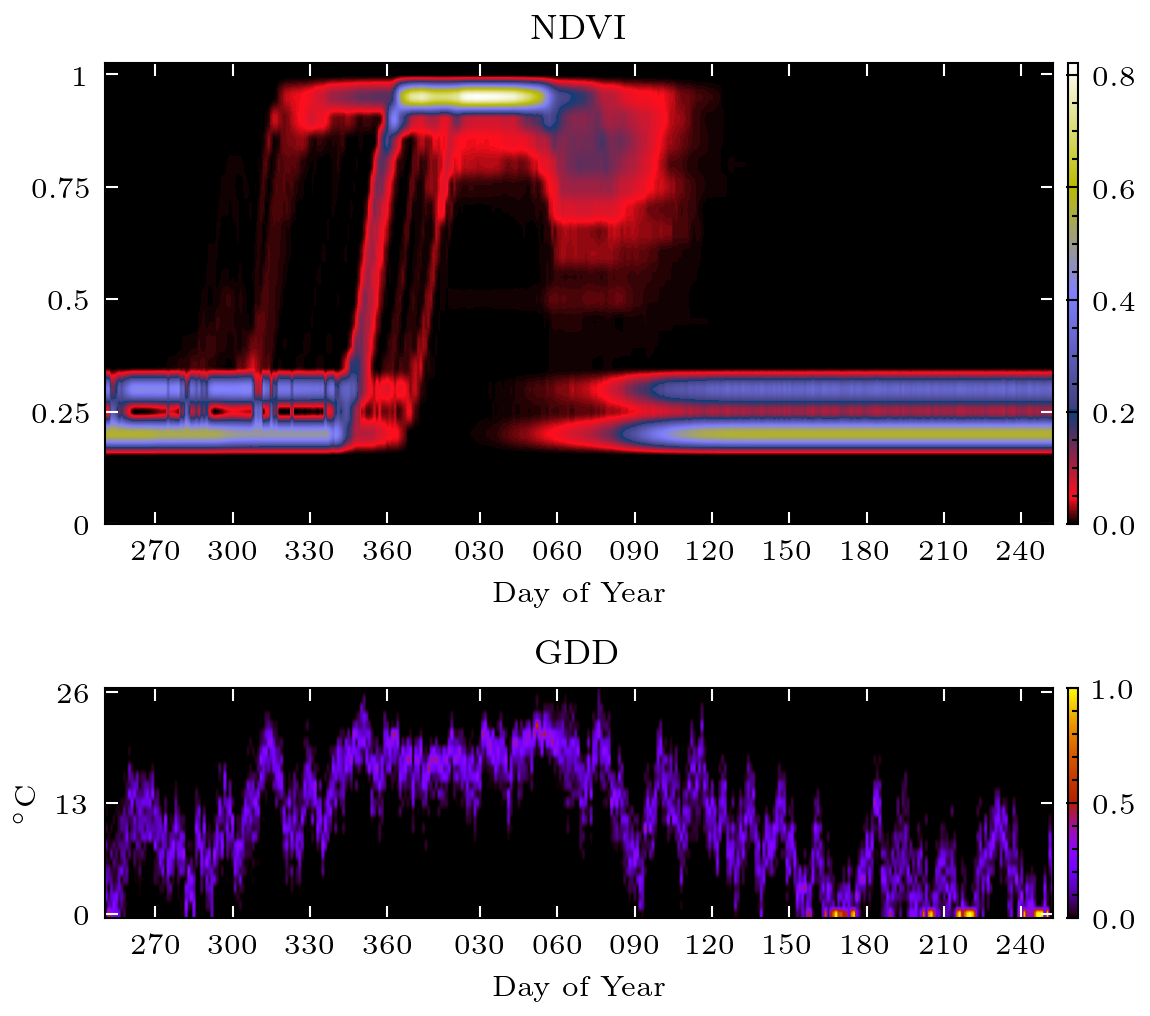}
			\caption{}
		\end{subfigure}
		\vskip\baselineskip
		\begin{subfigure}[b]{\textwidth}
			\centering
			\includegraphics[width=0.475\linewidth]{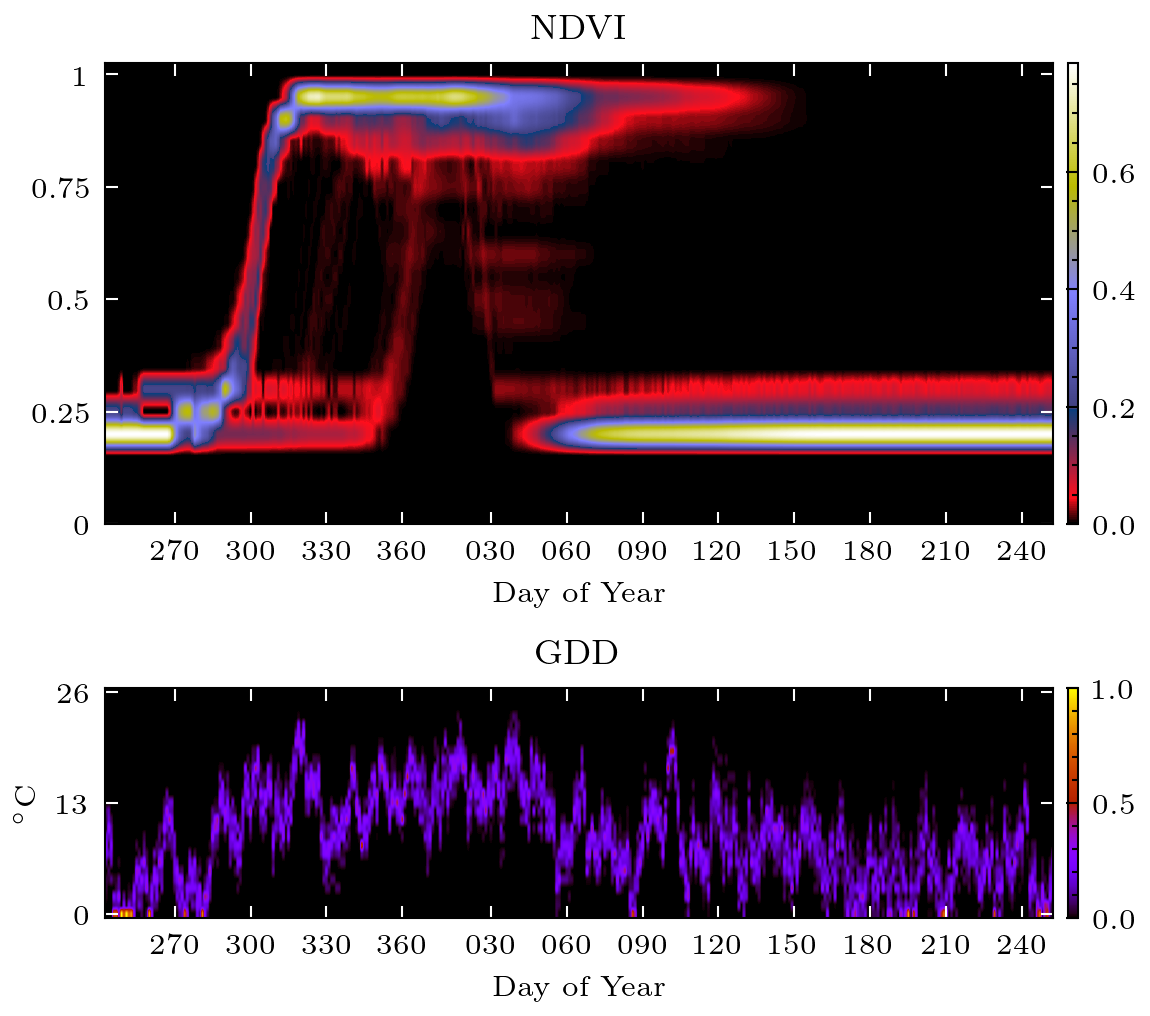}%
			\hfill
			\includegraphics[width=0.475\linewidth]{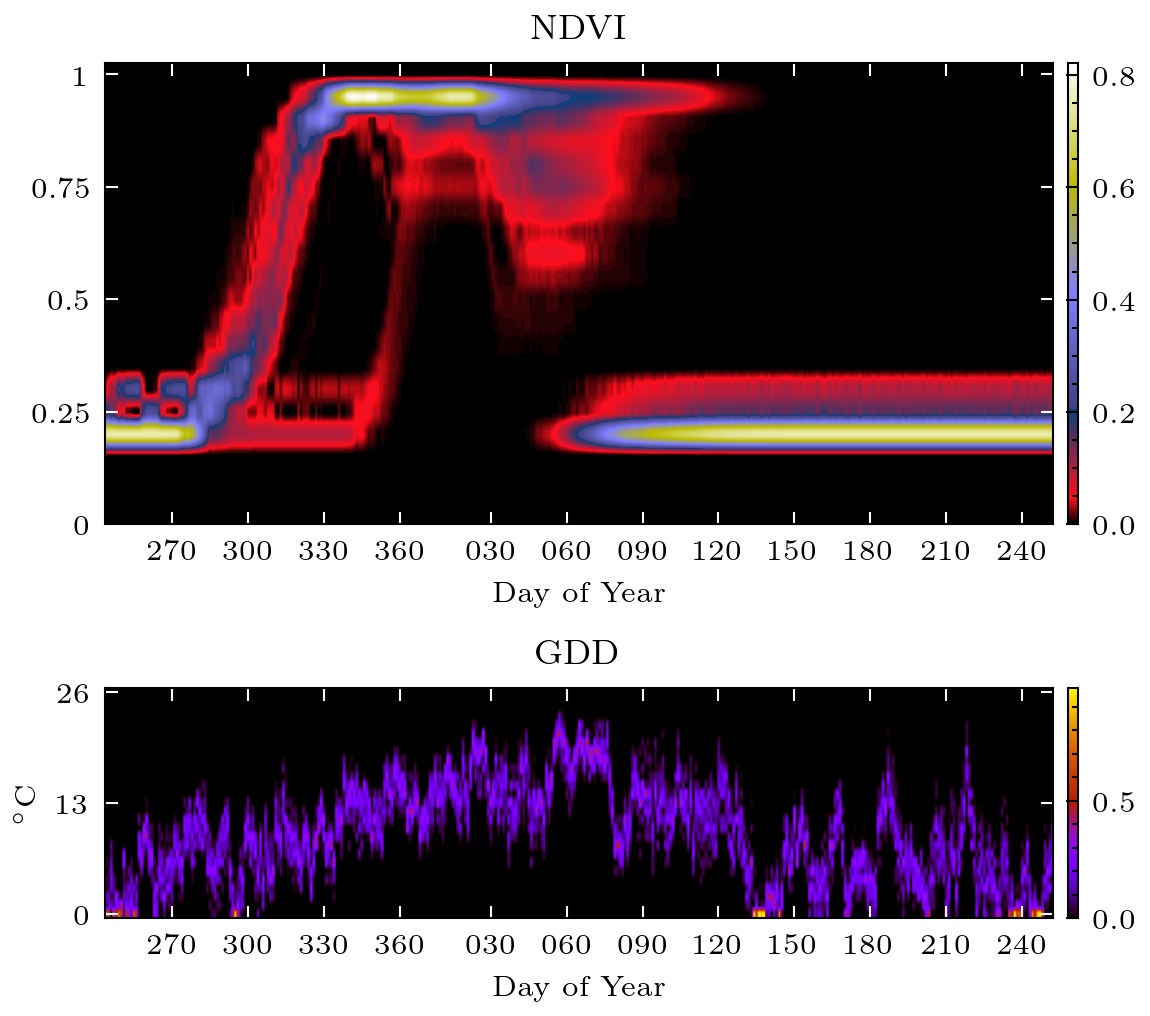}
			\caption{}
		\end{subfigure}
		\caption{Example synthetic NDVI and GDD histograms generated by the linked SWG-IXIM-SPART models for Zones (a) IX, (b), III, and (c) VI.}
		\label{fig:synthetic_input_examples}
	\end{figure}

	Network sensitivity to noise in the NDVI histogram was much greater when the network was trained only on the synthetic data, A$_{syn}$, with no examples from a surveyed region. For example in Zone VI during the 2019/20 season, shown in Figure \ref{fig:result_VI_1920}, which had 92\% early planting, a late increase in NDVI from DOY 60 to 90 (Figure \ref{fig:input_VI_1920}) caused a delayed estimate of the transition from Mature to Harvested and a small reversion from Mature to Dough when the network was trained on the A$_{syn}$ set, shown in Figure \ref{fig:result_VI_1920_synth}. As this NDVI increase subsided, further NDVI noise between DOY 180 and 240, likely from cover crop or winter crop emergence in some fields, caused another reversion from Harvested to Mature. This sensitivity to NDVI can be explained by inspection of example NDVI histograms from the synthetic data, shown in Figure \ref{fig:synthetic_input_examples}, which contain none of the early and late season noise caused by cover crop, double cropping, or cloud cover that is present in real NDVI data. 
	
	When trained on datasets that included surveyed region data, the network was less sensitive to NDVI noise, seen in Figures \ref{fig:result_VI_1920_surveyed}, \ref{fig:result_VI_1920_MSS}, and \ref{fig:result_VI_1920_joint}. Only a small reversion occurred when the network was trained on the A$_{syn}$U$_{sur}$ set, and the network was completely insensitive to this post-season noise when trained on the A$_{syn1}$U$_{sur}$ and U$_{sur}$ sets. This is notable because winter cover crops and dual cropping systems are uncommon in Iowa and Illinois, suggesting the de-sensitizing component in the surveyed region data is likely observational NDVI noise, such as cloud contamination, rather than pre- and post-season cropping. This suggests that surveyed region data contributes both real crop progress examples and realistic noise profiles which improve network performance.

	\subsection{Potential Improvements}
	Based upon the results of this study, we provide insights into three potential improvements to RS-based unsurveyed region CPE.
	
	In-season CPE could be significantly improved by increasing the variety and quantity of synthetic crop progress data. In this study, we used a single cultivar in Argentina, and planting windows from the historical distribution of early and late planting for each zone. Both cultivar and planting distributions could be varied. For example, seasonal GDD ranges for each zone could be calculated, and existing viable cultivars from the IXIM library could be used to increase the variety of GDD-specific silking and maturity times within the synthetic dataset. Additionally, simulations from another crop model, such as APSIM Maize \citep{holzworth_apsim_2014}, could provide more synthetic crop progress data using different crop modeling logic. This data could be combined during training, with validation losses monitored separately to ensure that a network learns crop progress notions which are applicable to both models and as well as surveyed region data.
	
	Secondly, the processing of RS data captured for real crop progress seasons in unsurveyed regions could be improved. The results from this study show that NDVI noise from pre- and post-season vegetation can affect network performance. This noise could be reduced by pixel-level masking of pre- and post-season NDVI signals. In this study, we used linearly interpolated pixel-level NDVI values, but higher complexity filters such as the Savitsky-Golay method \citep{chen_simple_2004} could be used to smooth NDVI pixel values before aggregation. Existing trend-based techniques such as those proposed by \cite{gao_within-season_2020, gao_detecting_2020} could then be used to estimate SOS and EOS dates on pixel-level NDVI curves. While these methods often have a delayed confirmatory signal, they may be sufficient to mask post-season pixel-level NDVI noise which affected network performance in this study, particularly in zones with a large separation between early and late planted crops, such as Zone V. Additionally, if early and late planting pixels can be differentiated in-season using crop mapping methods, it may also be possible to treat each planting window as a standalone season and remove the additional complexity required of a network to learn the dynamics of dual planting systems. 
	
	Finally, RS-based in-season CPE could be further improved using complementary information from multiple remote sensing systems. For example, this study used optical data from MODIS, but observations at other wavelengths, such as microwaves from the Sentinel 1 and upcoming NISAR missions, could provide additional guidance for canopy properties such as soil and vegetation water content and biomass. This could improve differentiation between crop progress stages, and their synthetic equivalents could be modeled using radiative transfer models linked to the synthetic data generation process.

	\section{Conclusions}
	Existing crop progress estimation methods have been limited to surveyed regions due to the requirement of \textit{in situ}, ground survey data for method training/calibration. In this study, we linked existing weather, crop, and optical RTM models to produce synthetic examples of crop progress for an unsurveyed region. Using Argentina as a test `unsurveyed' region surrogate, we combined surveyed crop progress data from the US Midwest and synthetic data produced for zones in Argentina to train an LSTM-based NN for in-season CPE. Adapting F$_1$ score to measure CPE accuracy, results showed that introducing synthetic data into the training set improved network CPE performance, particularly in regions with two separate planting windows. Net F$_1$ scores across all six target stages of corn crop progress in Argentina increased by 8.7\% when exposed to a combination of surveyed region and synthetic data during training, with performance most improved in dual planting zones. Results also showed that surveyed region data helps mitigate network sensitivity to NDVI noise. Overall, unsurveyed region performance was only 21\% lower than the best case scenario, when the same network was trained and evaluated in surveyed US Midwest states. This result is significant given that the network had never seen real crop progress examples from Argentina during training.
	
	Building upon the results of this study, we outlined areas that could further improve the accuracy of CPE in unsurveyed regions, including seasonal masking of NDVI and increasing the quantity and variety of synthetic data. The aim of this study was in-season CPE in unsurveyed regions, but the method for generating synthetic data for network training presented here could also be applied to other areas of agricultural RS, such as crop mapping and yield estimation in both surveyed and unsurveyed regions.
	
	\section*{Declaration of Interest}
	The authors declare that they have no known competing financial interests or personal relationships that could have appeared to influence the work reported in this paper.
	
	\section*{Acknowledgments}
	We gratefully acknowledge the USDA-NASS Upper Midwest and Heartland Regional Offices for providing the tabulated data from USDA Crop Progress and Condition reports. This study was funded in part by the University of Florida Informatics Institute, the University of Florida Center for Remote Sensing, and USDA-NRCS CIG Grant AGR00020960.

	\bibliographystyle{elsarticle-harv}
	\bibliography{refs/refs.bib}
	
	\clearpage
	\section{Figures}

	\newpage
	\newpage
	\newpage
	\newpage

	\newpage
	\newpage
	\newpage
	\newpage
	\newpage
	\clearpage

\clearpage
	\clearpage
\clearpage

	\clearpage
	\clearpage
	\clearpage

	\section{Tables}
	\begin{table}[htbp]
		\caption{Distribution of corn between early and late planting windows within zones in Argentina.}
		\label{tab:early_late_split}
		\resizebox{\textwidth}{!}{
			\begin{tabular}{@{}lllllll@{}}
				\toprule
				Zone & \multicolumn{2}{c}{2019/20}    & \multicolumn{2}{c}{2020/21}    & \multicolumn{2}{c}{Four Year Average} \\ \midrule
				& Early Planting & Late Planting & Early Planting & Late Planting & Early Planting     & Late Planting    \\ \midrule
				I    & -              & 100\%         & -              & 100\%         & -  & 100\%            \\
				II   & -              & 100\%         & -              & 100\%         & -  & 100\%            \\
				III  & 23\%           & 77\%          & 10\%           & 90\%          & 16\%               & 84\%             \\
				IV   & 25\%           & 75\%          & 12\%           & 88\%          & 19\%               & 81\%             \\
				V    & 45\%           & 55\%          & 38\%           & 62\%          & 38\%               & 62\%             \\
				VI   & 92\%           & 8\%           & 85\%           & 15\%          & 87\%               & 13\%             \\
				VII  & 92\%           & 8\%           & 86\%           & 14\%          & 86\%               & 14\%             \\
				IX   & 63\%           & 37\%          & 40\%           & 60\%          & 51\%               & 49\%             \\
				X    & 63\%           & 37\%          & 40\%           & 60\%          & 47\%               & 53\%             \\
				XII  & 67\%           & 33\%          & 54\%           & 46\%          & 59\%               & 41\%             \\
				XIII & 25\%           & 75\%          & 12\%           & 88\%          & 19\%               & 81\%             \\ \bottomrule
			\end{tabular}
		}
	\end{table}
	\newpage
	\begin{sidewaystable}[htbp]
		\caption{Sources, resolutions, and pre-processing of inputs to NN-based in-season CPE method}
		\label{tab:inputs}
		\centering
		\resizebox{\textwidth}{!}{
			\begin{tabular}{@{}llllll@{}}
				\toprule
				Input               & Source             & Spat. Resolution & Temp. Resolution & Pre-processing & Reference \\ \midrule
				NDVI                & MODIS Aqua         & 250 m              & 7 days              & linear interpolation - binned histogram   &  \citep{jenkerson_emodis_2010}        \\
				Growing Degree Days & NASA POWER         & 0.5$^\circ \times$ 0.625$^\circ$         & Daily          & binned histogram   &  \citep{nasa_langley_research_center_nasa_nodate}          \\ \bottomrule
		\end{tabular}}
	\end{sidewaystable}
	\newpage
	\begin{table}[htbp]
		\caption{Synthetic data generation inputs for the SWG, IXIM, and SPART models}
		\label{tab:simulation_inputs}
		\centering
		\resizebox{\textwidth}{!}{
			\begin{tabular}{@{}llll@{}}
				\toprule
				Parameter & Name   & Unit             & Value / Source     \\ \midrule
				\multicolumn{4}{c}{SWG}           \\ \midrule
				TMAX      & Daily Maximum Temperature              & $^\circ$C       & \cite{nasa_langley_research_center_nasa_nodate} \\
				TMIN      & Daily Minimum Temperature              & $^\circ$C       & \cite{nasa_langley_research_center_nasa_nodate}      \\
				SRAD      & Daily Total Solar Radiation            & W/m$^2$/day      & \cite{nasa_langley_research_center_nasa_nodate}      \\
				PPT       & Daily precipitiation   & mm/day           & \cite{nasa_langley_research_center_nasa_nodate}      \\ \midrule
				\multicolumn{4}{c}{DSSAT IXIM}       \\ \midrule
				TMAX      & Daily Maximum Temperature              & $^\circ$C       & Stochastic Weather Generator       \\
				TMIN      & Daily Minimum Temperature              & $^\circ$C       & Stochastic Weather Generator       \\
				SRAD      & Daily Total Solar Radiation            & W/m$^2$/day      & Stochastic Weather Generator       \\
				PPT       & Daily precipitiation   & mm/day           & Stochastic Weather Generator       \\
				Cultivar  & Corn cultivar parameters               & -                & \cite{andres_ferreyra_linked-modeling_2001}      \\
				Soil      & Soil properties        & -                & \cite{han_development_2019}        \\ \midrule
				\multicolumn{4}{c}{SPART}  \\ \midrule
				-         & Dry soil reflectance   & -                & \cite{meerdink_ecostress_2019} \\
				SM$_p$    & Soil moisture volume (\%)              & -                & DSSAT simulation   \\
				C$_{ab}$  & Chlorophyll a + b content              & $\mu$g cm$^{-2}$ & 40 \\
				C$_{dm}$  & Leaf mass per unit area                & g cm$^{-2}$      & 0.01               \\
				C$_{w}$   & Equivalent water thickness             & cm               & 0.02               \\
				C$_{s}$   & Brown pigments         & -                & 0  \\
				C$_{ant}$ & Anthocyanin content    & $\mu$g cm$^{-2}$ & 10 \\
				N         & Leaf structure parameter               & -                & 1.5                \\
				LAI       & Leaf area index        & m$^2$ m$^{-2}$   & DSSAT simulation   \\
				LIDF      & Leaf inclination distribution function & -                & \cite{judge_impact_2021}       \\
				q         & hot spot parameter     & -                & \cite{judge_impact_2021}        \\ \bottomrule
			\end{tabular}
		}
	\end{table}
	\newpage
	\begin{table}[htbp]
		\caption{Comparison of USDA, BdC, and study-defined growth stages}
		\label{tab:stages}
		\centering
		\begin{tabular}{@{}lll@{}}
			\toprule
			Target Stages              & BdC Stages     & USDA Stages              \\ \midrule
			\multirow[t]{2}{*}{Emerged}   & Expansión foliar                & \multirow[t]{2}{*}{Emerged} \\ 
			& Panojamiento   &          \\ \hline
			Silking    & Floración Femenina             & Silking  \\ \midrule
			\multirow[t]{2}{*}{Dough} & \multirow[t]{2}{*}{Grano Pastoso} & Dough    \\
			&                & Dent     \\ \hline 
			Mature     & Madurez Fisiológica            & Mature   \\ \midrule
			Harvested  & Cosecha        & Harvested                \\ \bottomrule
		\end{tabular}
	\end{table}
	\newpage

%
%

	\begin{table}[htbp]
			\centering
			\caption{Net F$_1$ scores for each training dataset in all zones in Argentina across the 2019/2020 and 2020/2021 test seasons. Bold values indicate the highest net F$_1$ score for each stage. Values in italics represent the scores for the SRR training set tested in Iowa during the 2018, 2019, and 2020 growing seasons.}
			\label{tab:f1_overall}
			\begin{tabular}{@{}ccccccc|c@{}}
				\toprule
				Training Set        & Preemergence   & Emerged        & Silking        & Dough          & Mature         & Harvested      & Overall        \\ \midrule
				U$_{sur}$           & 0.632          & 0.667          & 0.582          & 0.667          & 0.529          & 0.786          & 0.677          \\
				A$_{syn}$           & 0.586          & 0.661          & 0.574          & 0.480          & 0.567          & 0.755          & 0.618          \\
				A$_{syn1}$U$_{sur}$ & 0.656          & 0.706          & 0.633          & \textbf{0.690} & 0.639          & \textbf{0.834} & 0.723          \\
				A$_{syn}$U$_{sur}$  & \textbf{0.764} & \textbf{0.721} & \textbf{0.641} & 0.668          & \textbf{0.657} & 0.822          & \textbf{0.736} \\ \midrule
				SRR (Iowa)          & \textit{0.900} & \textit{0.913} & \textit{0.875} & \textit{0.909} & \textit{0.846} & \textit{0.917} & \textit{0.898} \\ \bottomrule
			\end{tabular}
		\end{table}
	\begin{sidewaystable}[]
			\centering
			\caption{F$_1$ score of each training method for all zones in Argentina in 2019/20. Bold values indicate the highest F$_1$ score for each stage within each zone. Net$_{S}$ and Net$_{Z}$ represent net F$_1$ scores by stage and zone, respectively.}
			\label{tab:f1_2019}
			\resizebox{\textwidth}{!}{
\begin{tabular}{@{}lcccccc|c|cccccc|c|@{}}
\toprule
Zone     & Pre-emergence  & Emerged        & Silking        & Dough          & Mature         & Harvested      & Net$_{S}$            & Pre-emergence  & Emerged        & Silking        & Dough          & Mature         & Harvested      & Net$_{S}$        \\ \midrule
\multicolumn{8}{c}{U$_{sur}$}                                                                                                   & \multicolumn{7}{c}{A$_{syn1}$U$_{sur}$}                                           \\ \midrule
I        & 0.684          & 0.771          & 0.791          & \textbf{0.894} & 0.640          & 0.793          & 0.764          & 0.694          & 0.748          & 0.706          & 0.821          & 0.684          & 0.828          & 0.759 \\
II       & 0.718          & \textbf{0.704} & {0.744}        & \textbf{0.744} & 0.527          & 0.806          & \textbf{0.723} & 0.602          & 0.687          & \textbf{0.746} & {0.716}        & \textbf{0.617} & {0.826}        & {0.717}    \\
III      & 0.450          & 0.377          & 0.366          & 0.530          & 0.413          & 0.832          & 0.563          & 0.389          & 0.422          & 0.389          & 0.510          & \textbf{0.718} & {0.899}        & 0.605          \\
IV       & 0.536          & 0.454          & 0.381          & 0.453          & 0.431          & 0.787          & 0.572          & 0.562          & \textbf{0.679} & \textbf{0.749} & \textbf{0.654} & 0.544          & {0.858}        & {0.700}    \\
V        & 0.412          & 0.523          & \textbf{0.389} & {0.498}        & {0.578}        & 0.667          & 0.552          & {0.417}        & {0.536}        & {0.386}        & 0.485          & 0.473          & \textbf{0.821} & {0.606}    \\
VI       & {0.775}        & \textbf{0.860} & 0.690          & \textbf{0.861} & {0.758}        & {0.951}        & {0.879}        & 0.750          & {0.850}        & \textbf{0.743} & 0.837          & \textbf{0.831} & \textbf{0.967} & \textbf{0.889} \\
VII      & 0.811          & 0.867          & 0.692          & \textbf{0.894} & \textbf{0.780} & \textbf{0.947} & \textbf{0.882} & {0.844}        & \textbf{0.876} & \textbf{0.731} & 0.860          & 0.760          & {0.938}        & 0.876 \\
IX       & 0.677          & 0.694          & 0.710          & 0.675          & 0.543          & 0.811          & 0.716          & \textbf{0.822} & \textbf{0.845} & \textbf{0.765} & 0.699          & {0.743}        & {0.863}        & \textbf{0.815} \\
X        & 0.720          & {0.762}        & 0.638          & {0.657}        & 0.546          & {0.800}        & {0.724}        & \textbf{0.825} & \textbf{0.783} & {0.696}        & \textbf{0.781} & \textbf{0.760} & \textbf{0.853} & \textbf{0.803} \\
XII      & {0.854}        & 0.849          & 0.683          & 0.735          & 0.559          & 0.761          & 0.759          & 0.852          & {0.862}        & \textbf{0.776} & {0.749}        & \textbf{0.670} & {0.789}        & 0.792    \\
XIII     & 0.616          & 0.649          & 0.602          & 0.491          & 0.354          & 0.772          & 0.637          & 0.620          & 0.670          & 0.630          & 0.451          & 0.457          & 0.803          & 0.662          \\ \midrule
Net$_{Z}$      & 0.641          & 0.684          & 0.607          & 0.668          & 0.557          & 0.822          & 0.666          & {0.661}        & {0.722}        & {0.664}        & {0.692}        & \textbf{0.662} & \textbf{0.867} & {0.711}    \\ \midrule
\multicolumn{8}{c}{A$_{syn}$}                                                                                                   & \multicolumn{7}{c}{A$_{syn}$U$_{sur}$}                                                        \\ \midrule
I        & {0.754}        & 0.783          & 0.727          & 0.659          & \textbf{0.844} & \textbf{0.927} & 0.803          & \textbf{0.830} & \textbf{0.842} & \textbf{0.870} & {0.883}        & {0.802}        & {0.859}        & \textbf{0.845} \\
II       & {0.770}        & {0.688}        & 0.593          & 0.513          & {0.572}        & \textbf{0.883} & 0.704          & \textbf{0.777} & 0.674          & 0.595          & 0.525          & 0.539          & 0.799          & 0.683          \\
III      & {0.557}        & {0.607}        & {0.631}        & \textbf{0.629} & {0.676}        & \textbf{0.907} & 0.696          & \textbf{0.739} & \textbf{0.640} & \textbf{0.645} & {0.622}        & 0.540          & 0.866          & \textbf{0.716} \\
IV       & {0.644}        & {0.649}        & 0.590          & 0.607          & \textbf{0.687} & \textbf{0.913} & \textbf{0.717} & \textbf{0.714} & 0.619          & {0.682}        & {0.616}        & {0.578}        & 0.834          & 0.702    \\
V        & 0.007          & 0.440          & 0.139          & 0.330          & 0.433          & 0.452          & 0.359          & \textbf{0.426} & \textbf{0.540} & 0.375          & \textbf{0.518} & \textbf{0.625} & {0.782}        & \textbf{0.614} \\
VI       & 0.758          & 0.835          & 0.565          & 0.620          & 0.423          & 0.676          & 0.641          & \textbf{0.794} & {0.850}        & {0.719}        & 0.791          & 0.726          & 0.937          & 0.857          \\
VII      & 0.812          & 0.851          & 0.616          & 0.645          & 0.495          & 0.728          & 0.689          & \textbf{0.864} & {0.873}        & {0.699}        & {0.857}        & {0.773}        & {0.938}        & {0.876}    \\
IX       & 0.641          & {0.792}        & 0.680          & \textbf{0.714} & \textbf{0.756} & 0.858          & 0.769          & {0.766}        & 0.745          & {0.740}        & {0.703}        & 0.741          & \textbf{0.885} & {0.791}    \\
X        & 0.043          & 0.176          & 0.001          & 0.151          & 0.410          & 0.495          & 0.325          & {0.730}        & 0.681          & 0.572          & 0.609          & {0.667}        & 0.715          & 0.679          \\
XII      & 0.181          & 0.701          & 0.720          & 0.373          & 0.552          & 0.496          & 0.488          & \textbf{0.871} & \textbf{0.879} & {0.742}        & \textbf{0.907} & \textbf{0.686} & \textbf{0.845} & \textbf{0.831} \\
XIII     & {0.872}        & {0.821}        & {0.721}        & \textbf{0.727} & \textbf{0.589} & \textbf{0.873} & 0.800          & \textbf{0.906} & \textbf{0.864} & \textbf{0.796} & {0.605}        & {0.494}        & {0.810}        & {0.783}    \\ \midrule
Net$_{Z}$      & 0.579          & 0.693          & 0.585          & 0.502          & 0.556          & 0.759          & 0.606          & \textbf{0.768} & \textbf{0.749} & \textbf{0.675} & \textbf{0.699} & {0.657}        & {0.851}        & \textbf{0.729} \\ \bottomrule
	\end{tabular}
				}
		\end{sidewaystable}
	\newpage
\begin{sidewaystable}[]
	\centering
	\caption{F$_1$ score of each training method for all zones in Argentina in 2020/21. Bold values indicate the highest F$_1$ score for each stage within each zone. Net$_{S}$ and Net$_{Z}$ represent net F$_1$ scores by stage and zone, respectively.}
	\label{tab:f1_2020}
	\resizebox{\textwidth}{!}{
\begin{tabular}{@{}lcccccc|c|cccccc|c|@{}}
	\toprule
	Zone     & Pre-emergence  & Emerged        & Silking        & Dough          & Mature         & Harvested      & Net$_{S}$            & Pre-emergence  & Emerged        & Silking        & Dough          & Mature         & Harvested      & Net$_{S}$        \\ \midrule
	\multicolumn{8}{c}{U$_{sur}$}                                                                                                   & \multicolumn{7}{c}{A$_{syn1}$U$_{sur}$}                                           \\ \midrule
	I        & 0.700          & 0.765          & \textbf{0.870} & \textbf{0.896} & 0.597          & 0.733          & 0.746          & 0.695          & {0.783}        & 0.786          & {0.812}        & 0.627          & 0.753          & 0.740          \\
	II       & 0.724          & 0.739          & {0.822}        & \textbf{0.789} & 0.451          & 0.691          & 0.691          & 0.598          & 0.728          & \textbf{0.836} & {0.764}        & 0.632          & 0.763          & 0.715          \\
	III      & 0.501          & 0.506          & 0.412          & \textbf{0.720} & 0.588          & 0.729          & 0.590          & 0.404          & 0.414          & 0.337          & 0.601          & {0.621}        & {0.760}        & 0.544          \\
	IV       & 0.547          & 0.511          & 0.396          & {0.664}        & 0.566          & 0.716          & 0.587          & 0.630          & \textbf{0.714} & \textbf{0.788} & \textbf{0.848} & {0.659}        & {0.770}        & \textbf{0.722} \\
	V        & 0.313          & 0.419          & \textbf{0.318} & \textbf{0.542} & {0.450}        & 0.685          & 0.508          & {0.410}        & \textbf{0.547} & {0.311}        & 0.405          & 0.414          & \textbf{0.807} & \textbf{0.568} \\
	VI       & {0.701}        & {0.821}        & \textbf{0.626} & \textbf{0.768} & 0.688          & 0.919          & \textbf{0.821} & 0.695          & 0.768          & 0.489          & 0.709          & \textbf{0.778} & \textbf{0.940} & 0.811          \\
	VII      & 0.801          & {0.822}        & {0.595}        & {0.779}        & {0.674}        & \textbf{0.896} & \textbf{0.814} & {0.828}        & \textbf{0.829} & \textbf{0.629} & \textbf{0.790} & 0.664          & 0.851          & 0.797 \\
	IX       & 0.642          & 0.548          & 0.426          & 0.443          & 0.350          & 0.671          & 0.556          & \textbf{0.751} & \textbf{0.667} & \textbf{0.572} & {0.566}        & {0.661}        & 0.753          & \textbf{0.689} \\
	X        & {0.706}        & {0.645}        & {0.420}        & {0.482}        & 0.385          & {0.664}        & 0.594          & \textbf{0.758} & \textbf{0.647} & \textbf{0.455} & \textbf{0.708} & \textbf{0.650} & \textbf{0.767} & \textbf{0.695} \\
	XII      & 0.803          & {0.801}        & {0.632}        & 0.697          & 0.484          & 0.687          & 0.700          & {0.818}        & \textbf{0.814} & \textbf{0.658} & {0.715}        & {0.627}        & {0.741}        & {0.742}    \\
	XIII     & 0.574          & 0.620          & 0.594          & 0.605          & 0.326          & 0.638          & 0.572          & 0.674          & 0.728          & 0.691          & 0.601          & 0.492          & 0.694          & 0.659          \\ \midrule
	Net$_{Z}$      & 0.623          & 0.651          & 0.557          & {0.666}        & 0.504          & 0.744          & 0.626          & {0.652}        & {0.691}        & {0.600}        & \textbf{0.688} & {0.619}        & \textbf{0.794} & {0.673}    \\ \midrule
	\multicolumn{8}{c}{A$_{syn}$}                                                                                                   & \multicolumn{7}{c}{A$_{syn}$U$_{sur}$}                                                        \\ \midrule
	I        & {0.756}        & \textbf{0.799} & 0.816          & 0.659          & {0.763}        & \textbf{0.890} & 0.784          & \textbf{0.769} & 0.768          & {0.869}        & 0.805          & \textbf{0.775} & {0.835}        & \textbf{0.798} \\
	II       & \textbf{0.817} & \textbf{0.793} & 0.750          & 0.583          & \textbf{0.670} & \textbf{0.889} & \textbf{0.752} & {0.810}        & {0.765}        & 0.740          & 0.623          & {0.642}        & {0.773}        & {0.725}    \\
	III      & {0.751}        & \textbf{0.681} & \textbf{0.790} & {0.661}        & \textbf{0.653} & \textbf{0.846} & \textbf{0.725} & \textbf{0.795} & {0.676}        & {0.700}        & 0.683          & 0.618          & 0.731          & {0.707}    \\
	IV       & {0.699}        & 0.638          & 0.552          & 0.582          & \textbf{0.690} & \textbf{0.852} & 0.685          & \textbf{0.783} & {0.683}        & {0.632}        & 0.620          & 0.594          & 0.707          & {0.682}    \\
	V        & 0.033          & 0.165          & 0.000          & 0.137          & 0.335          & 0.517          & 0.294          & \textbf{0.417} & {0.449}        & 0.275          & {0.533}        & \textbf{0.553} & {0.777}        & 0.565 \\
	VI       & 0.677          & \textbf{0.827} & 0.544          & 0.522          & 0.656          & 0.775          & 0.711          & \textbf{0.766} & 0.795          & {0.553}        & {0.640}        & {0.773}        & {0.926}        & {0.811}    \\
	VII      & 0.737          & 0.806          & 0.507          & 0.581          & 0.656          & 0.820          & 0.731          & \textbf{0.837} & 0.810          & 0.544          & 0.706          & \textbf{0.746} & {0.878}        & 0.801          \\
	IX       & 0.483          & 0.554          & {0.544}        & \textbf{0.579} & \textbf{0.689} & \textbf{0.778} & 0.625          & {0.694}        & {0.579}        & 0.488          & 0.523          & 0.620          & {0.754}        & {0.644}    \\
	X        & 0.007          & 0.140          & 0.070          & 0.195          & 0.494          & 0.600          & 0.345          & 0.671          & 0.535          & 0.376          & 0.407          & {0.634}        & 0.633          & 0.575          \\
	XII      & 0.098          & 0.188          & 0.058          & 0.279          & 0.551          & 0.306          & 0.318          & \textbf{0.825} & 0.751          & 0.610          & \textbf{0.797} & \textbf{0.746} & \textbf{0.791} & \textbf{0.769} \\
	XIII     & {0.879}        & {0.830}        & {0.704}        & {0.641}        & {0.496}        & \textbf{0.821} & 0.750          & \textbf{0.896} & \textbf{0.836} & \textbf{0.824} & \textbf{0.716} & \textbf{0.564} & {0.719}        & \textbf{0.762} \\ \midrule
	Net$_{Z}$      & 0.592          & 0.629          & 0.561          & 0.460          & 0.567          & 0.750          & 0.573          & \textbf{0.760} & \textbf{0.694} & \textbf{0.605} & 0.639          & \textbf{0.657} & {0.787}        & \textbf{0.687} \\ \bottomrule
\end{tabular}
	}
\end{sidewaystable}
\end{document}